\documentclass[11pt,a4paper]{article}
\usepackage[hyperref]{arr}
\usepackage{amsmath}
\usepackage{xcolor}
\usepackage{times}
\usepackage{ragged2e}
\usepackage{bm}
\usepackage{CJKutf8}
\usepackage{latexsym}
\usepackage{subfigure}
\usepackage{graphicx}
\usepackage{float}
\usepackage{longtable}
\usepackage{booktabs} 
\usepackage{multirow}
\usepackage{amssymb}
\usepackage[ruled,vlined]{algorithm2e}
\usepackage{color, soul}

\usepackage{fancyhdr}
\newcommand{\vpara}[1]{\vspace{0.07in}\noindent\textbf{#1}\xspace}
\usepackage{microtype}

\aclfinalcopy 

\usepackage[T1]{fontenc}

\usepackage[utf8]{inputenc}

\title{Pre-Trained Models: Past, Present and Future}

\author{Xu Han$^{1}\thanks{\quad The first six authors contribute equally to organize this paper. The order is determined by dice rolling.}$\hspace{0.5em}, Zhengyan Zhang$^{1*}$, Ning Ding$^{1*}$, Yuxian Gu$^{1*}$,  \textbf{Xiao Liu}$^{1*}$, \textbf{Yuqi Huo}$^{2*}$, \\ 
\textbf{Jiezhong Qiu}$^{1}$, \textbf{Yuan Yao}$^{1}$, \textbf{Ao Zhang}$^{1}$, \textbf{Liang Zhang}$^{2}$, \textbf{Wentao Han}$^{1}$\thanks{\quad All faculty authors are alphabetically sorted.}, \textbf{Minlie Huang}$^{1\dagger}$, \\
\textbf{Qin Jin}$^{2\dagger}$, 
\textbf{Yanyan Lan}$^{4\dagger}$, \textbf{Yang Liu}$^{1,4\dagger}$,
\textbf{Zhiyuan Liu}$^{1\dagger}$, \textbf{Zhiwu Lu}$^{3\dagger}$,
\textbf{Xipeng Qiu}$^{5\dagger}$, \\
\textbf{Ruihua Song}$^{3\dagger}$, \textbf{Jie Tang}$^{1\dagger}$, \textbf{Ji-Rong Wen}$^{3\dagger}$, 
\textbf{Jinhui Yuan}$^{6\dagger}$, \textbf{Wayne Xin Zhao}$^{3\dagger}$, \textbf{Jun Zhu}$^{1\dagger}$ \\
$^{1}$ Department of Computer Science and Technology, Tsinghua University, Beijing, China \\
$^{2}$ School of Information, Renmin University of China, Beijing, China \\
$^{3}$ Gaoling School of Artificial Intelligence, Renmin University of China, Beijing, China \\
$^{4}$ Institute for AI Industry Research, Tsinghua University, Beijing, China \\
$^{5}$ School of Computer Science, Fudan University, Shanghai, China \\
$^{6}$ OneFlow Inc., Beijing, China \\
\small{\texttt{\{hanxu17,zy-z19,dingn18,gu-yx17,liuxiao17,qiujz16,yuan-yao18\}@mails.tsinghua.edu.cn}}, \\
\small{\texttt{\{hanwentao,aihuang,lanyanyan,liuyang2011,liuzy,jietang,dcszj\}@tsinghua.edu.cn}}, \\
\small{\texttt{\{bnhony,zhangliang00,qjin,luzhiwu,jrwen,batmanfly\}@ruc.edu.cn}}, \\
\small{\texttt{xpqiu@fudan.edu.cn, songruihua\_bloon@outlook.com, yuanjinhui@oneflow.org}}
}
\date{}

\begin{document}
\maketitle
\begin{abstract}

Large-scale pre-trained models (PTMs) such as BERT and GPT have recently achieved great success and become a milestone in the field of artificial intelligence (AI). Owing to sophisticated pre-training objectives and huge model parameters, large-scale PTMs can effectively capture knowledge from massive labeled and unlabeled data. By storing knowledge into huge parameters and fine-tuning on specific tasks, the rich knowledge implicitly encoded in huge parameters can benefit a variety of downstream tasks, which has been extensively demonstrated via experimental verification and empirical analysis. It is now the consensus of the AI community to adopt PTMs as backbone for downstream tasks rather than learning models from scratch. In this paper, we take a deep look into the history of pre-training, especially its special relation with transfer learning and self-supervised learning, to reveal the crucial position of PTMs in the AI development spectrum. Further, we comprehensively review the latest breakthroughs of PTMs. These breakthroughs are driven by the surge of computational power and the increasing availability of data, towards four important directions: designing effective architectures, utilizing rich contexts, improving computational efficiency, and conducting interpretation and theoretical analysis. Finally, we discuss a series of open problems and research directions of PTMs, and hope our view can inspire and advance the future study of PTMs.

\end{abstract}

\section{Introduction}
\label{sec:introduction}
Deep neural networks, such as convolutional neural networks (CNNs)~\cite{krizhevsky2012imagenet,kim2014convolutional,kalchbrenner2014convolutional,he2016deep}, recurrent neural networks (RNNs)~\cite{sutskever2014sequence,donahue2015long,liu2016recurrent,wu2016google}, graph neural networks (GNNs)~\cite{kipf2016semi,velivckovic2017graph,schlichtkrull2018modeling}, and attention neural networks~\cite{jaderberg2015spatial,wang2017residual}, have been widely applied for various artificial intelligence (AI) tasks in recent years. Different from previous non-neural models that largely relied on hand-crafted features and statistical methods, neural models can automatically learn low-dimensional continuous vectors (\emph{a.k.a.,} distributed representations) from data as task-specific features, thereby getting rid of complex feature engineering. Despite the success of deep neural networks, a number of studies have found that one of their critical challenges is data hungry. Since deep neural networks usually have a large number of parameters, they are thus easy to overfit and have poor generalization ability~\cite{belkin2019reconciling,xu2020neural} without sufficient training data.

\begin{figure*}[t]
\centering
\subfigure[Evaluation on language understanding benchmark GLUE.]{
\label{fig:change1}
\begin{minipage}[t]{0.48\linewidth}
\centering
\includegraphics[width=0.9\linewidth]{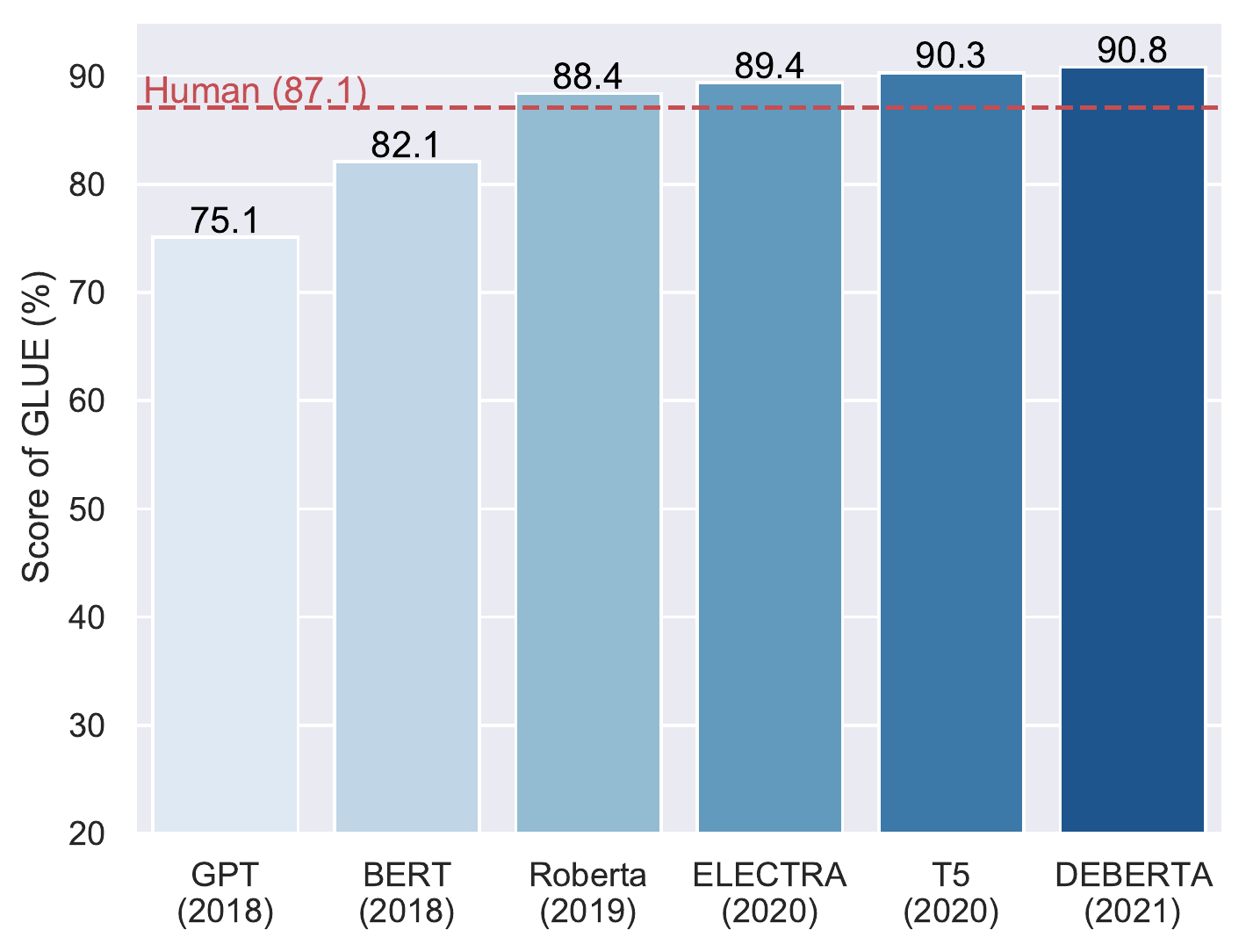}
\end{minipage}
}
\hfill
\subfigure[Manual evaluation on dialogue systems.]{
\label{fig:change2}
\begin{minipage}[t]{0.48\linewidth}
\centering
\includegraphics[width=0.9\linewidth]{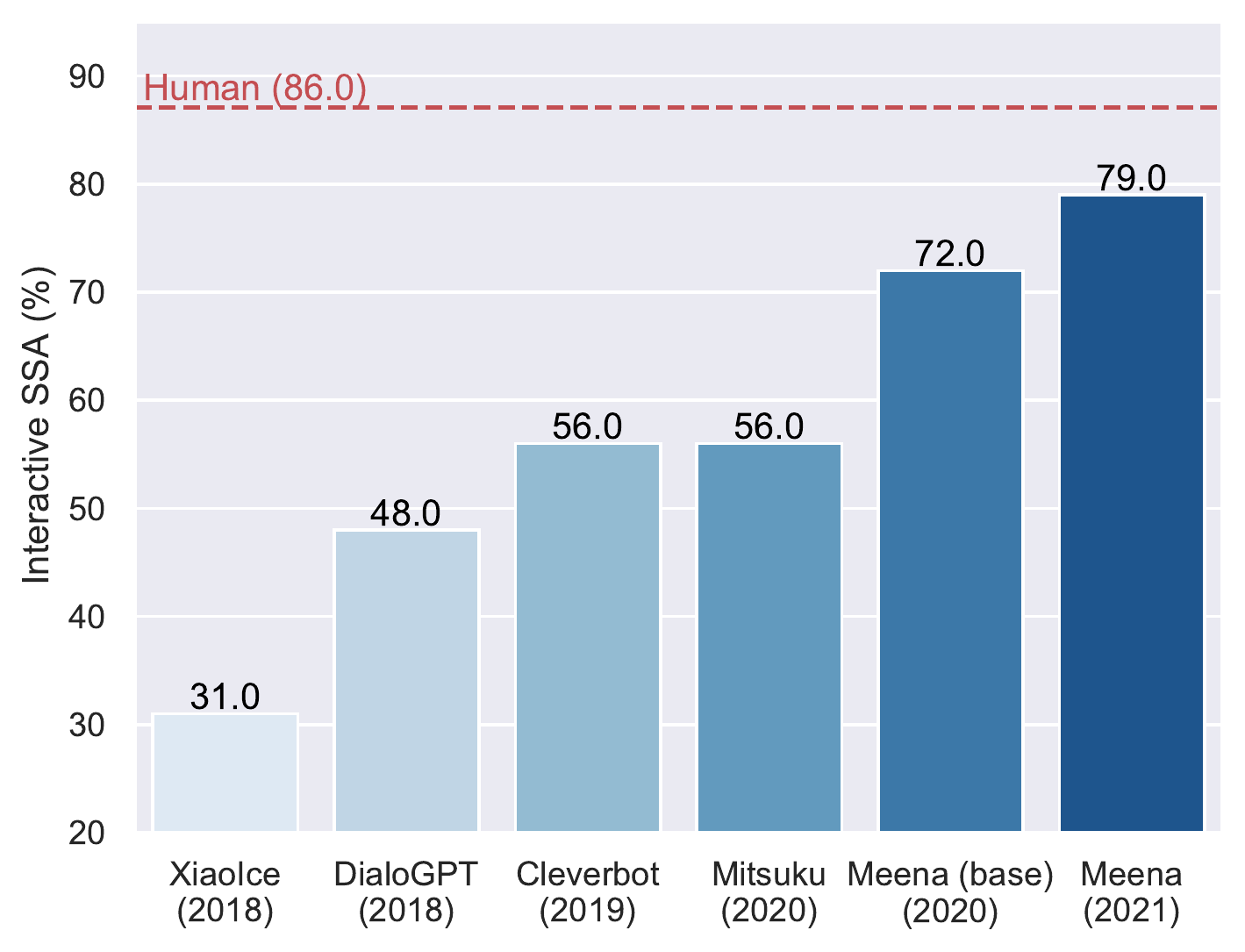}
\end{minipage}
}
\caption{The two figures show the significant improvement on performance of both language understanding and language generation after using large-scale PTMs.}
\end{figure*}

Considering this issue, over the same period of developing deep neural networks, massive efforts have been devoted to manually constructing high-quality datasets for AI tasks~\cite{deng2009imagenet,lin2014microsoft,bojar2014findings}, making it possible to learn effective neural models for specific tasks that are superior to conventional non-neural models. However, it is expensive and time-consuming to  manually annotate large-scale data. For example, utilizing crowdsourcing to segment images costs about \$6.4 per image~\cite{liu2020self}. Some complex tasks that require expert annotations may charge much more to build their datasets. Several tasks such as visual recognition~\cite{deng2009imagenet} and machine translation~\cite{bojar2014findings} have datasets containing millions of samples, yet it is impossible to build such large-scale datasets for all AI tasks. More generally, the dataset of a specific AI task usually has a limited size. Hence, for a long time until now, it has been a key research issue: \textit{how to train effective deep neural models for specific tasks with limited human-annotated data}.

One milestone for this issue is the introduction of transfer learning~\cite{thrun1998learning,pan2009survey}. Instead of training a model from scratch with large amounts of data, human beings can learn to solve new problems with very few samples. This amazing learning process is motivated by the fact that human beings can use previously learned knowledge to handle new problems. Inspired by this, transfer learning formalizes a two-phase learning framework: a pre-training phase to capture knowledge from one or more source tasks, and a fine-tuning stage to transfer the captured knowledge to target tasks. Owing to the wealth of knowledge obtained in the pre-training phase, the fine-tuning phase can enable models to well handle target tasks with limited samples. 

Transfer learning provides a feasible method for alleviating the challenge of data hungry, and it has soon been widely applied to the field of computer vision (CV). A series of CNNs~\cite{krizhevsky2012imagenet,simonyan2014very,szegedy2015going,he2016deep} are pre-trained on the human-annotated visual recognition dataset ImageNet~\cite{deng2009imagenet}. Benefiting from the strong visual knowledge distributed in ImageNet, fine-tuning these pre-trained CNNs with a small amount of task-specific data can perform well on downstream tasks. This triggers the first wave of exploring pre-trained models (PTMs) in the era of deep learning. In this wave, PTMs are used for almost all CV tasks such as image classification~\cite{he2016deep}, object detection~\cite{sermanet2013overfeat,ren2015faster}, image segmentation~\cite{long2015fully}, and image captioning~\cite{vinyals2015show}. 

\begin{figure*}[t]
\centering
\subfigure[The number of publications on ``language models'' and their citations in recent years.]{
\label{fig:change3}
\begin{minipage}[t]{0.48\linewidth}
\centering
\includegraphics[width=1.0\linewidth]{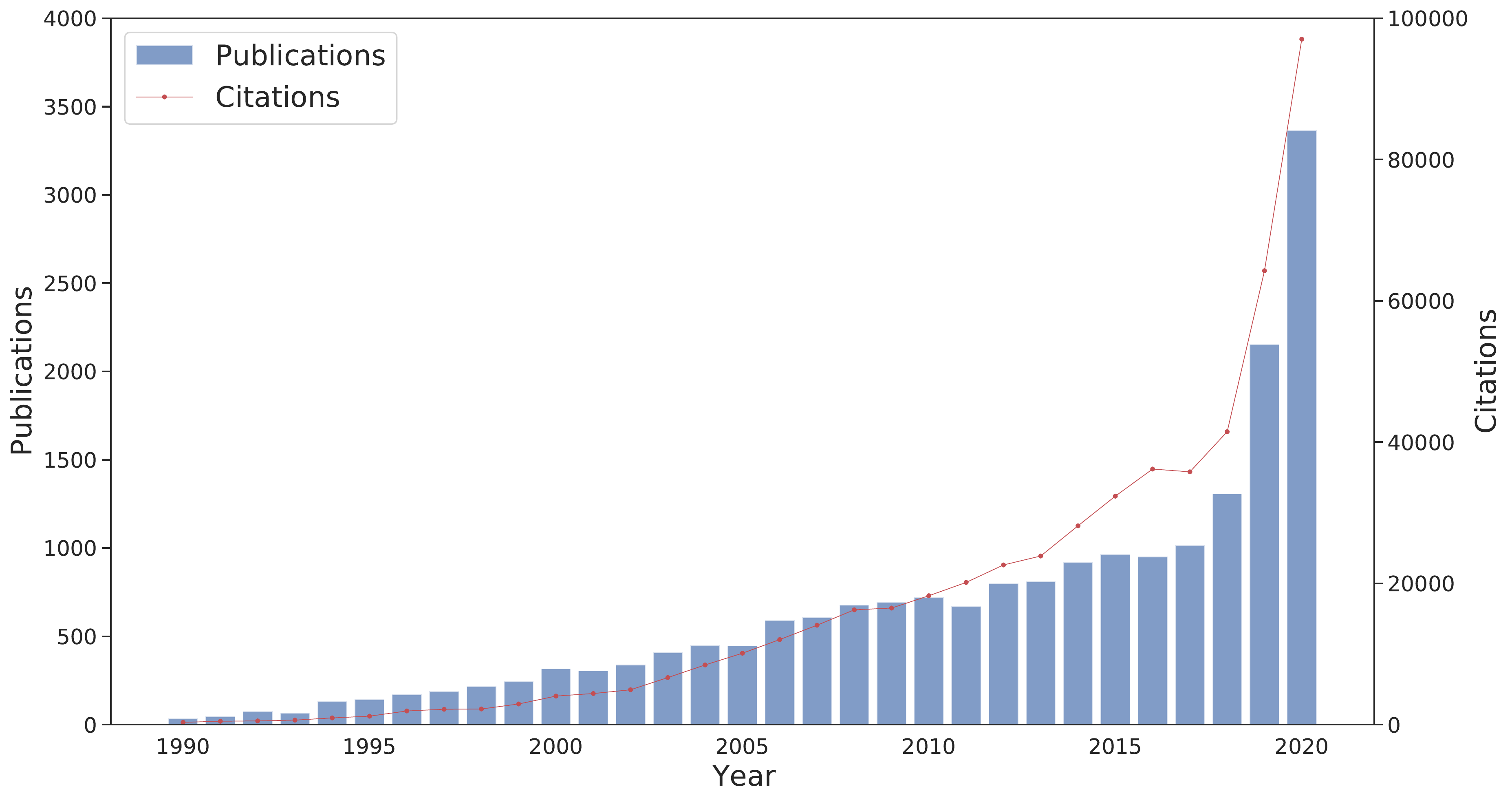}
\end{minipage}
}
\hfill
\subfigure[The model size and data size applied by recent NLP PTMs. A base-10 log scale is used for the figure.]{
\label{fig:change4}
\begin{minipage}[t]{0.48\linewidth}
\centering
\includegraphics[width=0.97\linewidth]{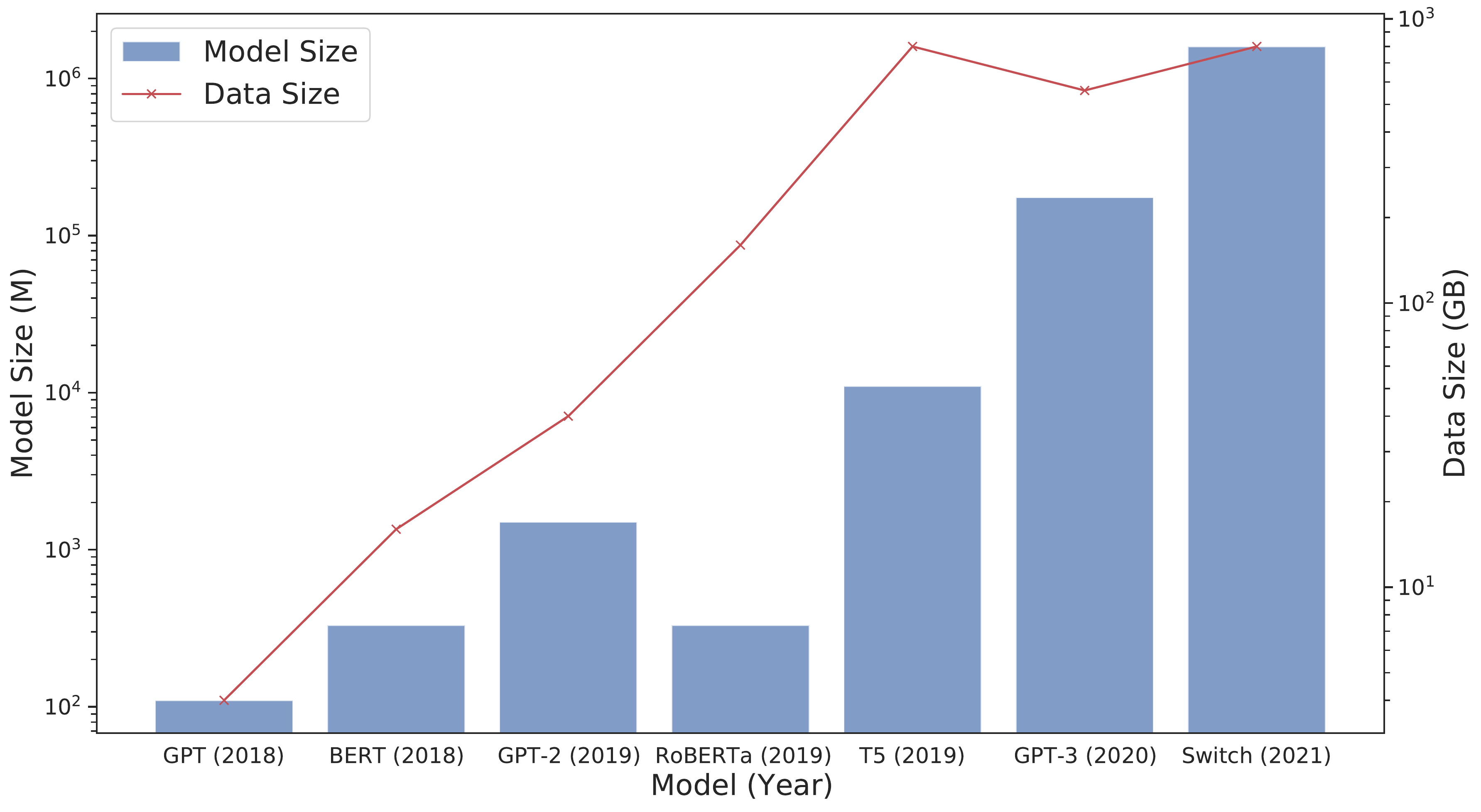}
\end{minipage}
}
\caption{Figure~\ref{fig:change3} shows the number of publications with the keyword ``language model'' as well as their citations in different years. Figure~\ref{fig:change4} shows the parameter size of large-scale PTMs for NLP tasks and the pre-training data size are increasing by $10$ times per year. From these figures, we can find that, after 2018, when large-scale NLP PTMs begin to be explored, more and more efforts are devoted to this field, and the model size and data size used by the PTMs are also getting larger.}
\end{figure*}

The natural language processing (NLP) community was also aware of the potential of PTMs and started to develop PTMs for NLP tasks~\cite{qiu2020pre}. 
To take full advantage of large-scale unlabeled corpora to provide versatile linguistic knowledge for NLP tasks, the NLP community adopts self-supervised learning~\cite{liu2020self} to develop PTMs.
The motivation of self-supervised learning is to leverage intrinsic correlations in the text as supervision signals instead of human supervision. For example, given the sentence ``Beijing is the capital of China'', we mask the last word in the sentence, and then require models to predict the masked position with the word ``China''. Through self-supervised learning, tremendous amounts of unlabeled textual data can be utilized to capture versatile linguistic knowledge without labor-intensive workload. This self-supervised setting in essence follows the well-known language model learning~\cite{bengio2003neural}. 

For a long time, the problem of vanishing or exploding gradients~\cite{bengio1994learning} is the pain point of using deep neural networks for NLP tasks. Therefore, when the CV community advances the research of deep PTMs, the early exploration of the NLP community focuses on pre-training shallow networks to capture semantic meanings of words, like Word2Vec~\cite{mikolov2013distributed,mikolov2013efficient,mikolov2013linguistic} and GloVe~\cite{pennington2014glove}. Although these pre-trained word embeddings play an important role in various NLP tasks, they still face a major limitation to represent polysemous words in different contexts, as each word is represented by only one dense vector. A famous example in NLP is that the word ``bank'' has entirely different meanings in the sentences ``open a bank account'' and ``on a bank of the river''. This motivates pre-training RNNs to provide contextualized word embeddings~\cite{melamud2016context2vec,peters2018deep,howard2018universal}, yet the performance of these models is still limited by their model size and depth. 

With the development of deep neural networks in the NLP community, the introduction of Transformers~\cite{vaswani2017attention} makes it feasible to train very deep neural models for NLP tasks. With Transformers as architectures and language model learning as objectives, deep PTMs GPT~\cite{radfordimproving} and BERT~\cite{devlin2019bert} are proposed for NLP tasks in 2018. From GPT and BERT, we can find that when the size of PTMs becomes larger, large-scale PTMs with hundreds of millions of parameters can capture polysemous disambiguation, lexical and syntactic structures, as well as factual knowledge from the text. By fine-tuning large-scale PTMs with quite a few samples, rich linguistic knowledge of PTMs brings awesome performance on downstream NLP tasks. As shown in Figure~\ref{fig:change1} and Figure~\ref{fig:change2}, large-scale PTMs well perform on both language understanding and language generation tasks in the past several years and even achieve better results than human performance. As shown in Figure~\ref{fig:change3}, all these efforts and achievements in the NLP community let large-scale PTMs become the focus of AI research, after the last wave that PTMs allow for huge advances in the CV community. 

\begin{figure*}[t]
		\centering  
		\includegraphics[width=0.9\linewidth]{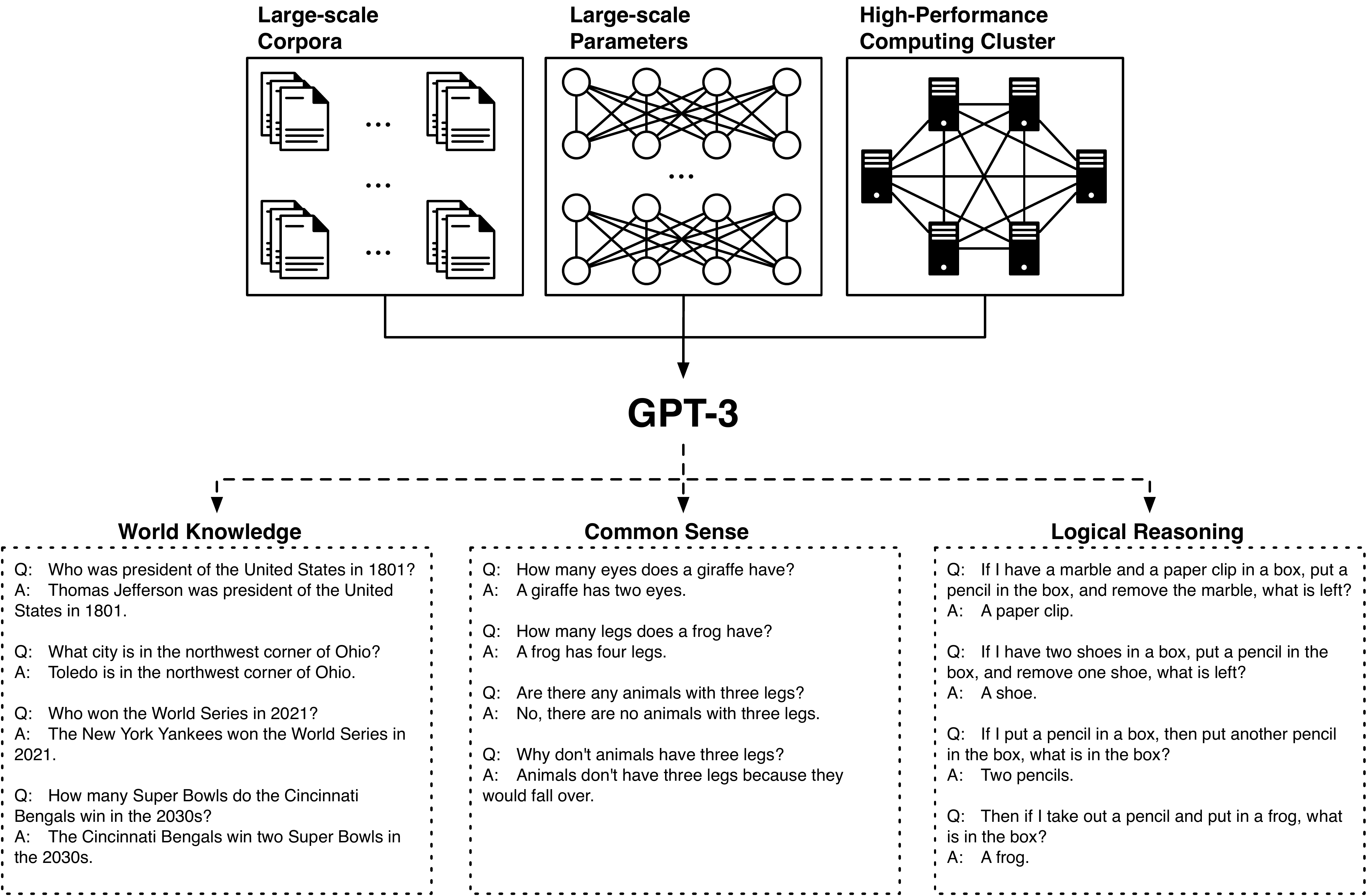}
		\caption{GPT-3, with 175 billion parameters, uses 560 GB data and 10,000 GPUs for its training. It has shown the abilities of learning world knowledge, common sense, and logical reasoning.}
		\label{fig:gpt3example} 
\end{figure*}

Up to now, various efforts have been devoted to exploring large-scale PTMs, either for NLP~\cite{radford2019language,liu2019roberta,raffel2020exploring,lewis2020bart}, or for CV~\cite{lu2019vilbert,li2019visualbert,tan2019lxmert}. Fine-tuning large-scale PTMs for specific AI tasks instead of learning models from scratch has also become a consensus~\cite{qiu2020pre}. As shown in Figure~\ref{fig:change4}, with the increasing computational power boosted by the wide use of distributed computing devices and strategies, we can further advance the parameter scale of PTMs from million-level to billion-level~\cite{brown2020language,lepikhin2020gshard,zeng2021pangu,zhang2020cpm,cpm-v2} and even trillion-level~\cite{fedus2021switch}. And the emergence of GPT-3~\cite{brown2020language}, which has hundreds of billions of parameters, enables us to take a glimpse of the latent power distributed in massive model parameters, especially the great abilities of few-shot learning like human beings (shown in Figure~\ref{fig:gpt3example}).

The existing large-scale PTMs have improved the model performance on various AI tasks and even subverted our current perception of the performance of deep learning models. However, several fundamental issues about  PTMs still remain: it is still not clear for us the nature hidden in huge amounts of model parameters, and huge computational cost of training these behemoths also prevents us from further exploration. At this moment, these PTMs have pushed our AI researchers to a crossroad, with a number of open directions to go. 

``\emph{Rome wasn't built in a day}''--- PTMs also experience a long development before achieving the latest success. To this end, we try to trace the development history of PTMs and draw their positions in the AI spectrum, which can give us a clear understanding of the core research issues of PTMs. Then, we introduce the details of various latest PTMs, following four important lines that are currently being advanced, including designing effective architectures, utilizing rich contexts, improving computational efficiency, and conducting interpretation and theoretical analysis. By integrating the current development of PTMs into the context of the historical spectrum, we discuss several open problems and conclude promising future directions for PTMs. We hope our efforts in this paper can advance further development of PTMs. In what follows, we will introduce the background of pre-training in Section~\ref{sec:background} and Section~\ref{sec:plm}, the model architectures of PTMs in Section~\ref{sec:arch}, using multi-source heterogeneous data for PTMs in Section~\ref{sec:rich-data}, the computational efficiency optimization of PTMs in Section~\ref{sec:efficiency}, and the theoretical analysis of PTMs in Section~\ref{sec:analysis}. Finally, we will briefly discuss a series of open problems and promising directions towards better PTMs in the future.

\section{Background}
\label{sec:background}

Although effective PTMs have recently gained the attention of researchers, pre-training is not a novel machine learning tool. In fact, pre-training has been developed for decades, as a typical machine learning paradigm. In this section, we introduce the development of pre-training in the AI spectrum, from early supervised pre-training to current self-supervised pre-training, which can lead to a brief understanding of the background of PTMs.

\subsection{Transfer Learning and Supervised Pre-Training}

The early efforts of pre-training are mainly involved in transfer learning~\cite{thrun1998learning}. The study of transfer learning is heavily motivated by the fact that people can rely on previously learned knowledge to solve new problems and even achieve better results. More formally, transfer learning aims to capture important knowledge from multiple source tasks and then apply the knowledge to a target task.

In transfer learning, source tasks and target tasks may have completely different data domains and task settings, yet the knowledge required to handle these tasks is consistent~\cite{pan2009survey}. It is thus important to select a feasible method to transfer knowledge from source tasks to target tasks. To this end, various pre-training methods have been proposed to work as the bridge between source and target tasks. Specifically, these methods first pre-train models on the data of multiple source tasks to pre-encode knowledge and then transfer the pre-encoded knowledge to train models for target tasks.

Generally, two pre-training approaches are widely explored in transfer learning: feature transfer and parameter transfer. Feature transfer methods pre-train effective feature representations to pre-encode knowledge across domains and tasks~\cite{johnson2005high,evgeniou2007multi,dai2007co,raina2007self}. By injecting these pre-trained representations into target tasks, model performance of target tasks can be significantly improved. Parameter transfer methods follow an intuitive assumption that source tasks and target tasks can share model parameters or prior distributions of hyper-parameters. Therefore, these methods pre-encode knowledge into shared model parameters~\cite{lawrence2004learning,evgeniou2004regularized,williams2007multi,gao2008knowledge}, and then transfer the knowledge by fine-tuning pre-trained parameters with the data of target tasks. 

To some extent, both representation transfer and parameter transfer lay the foundation of PTMs. Word embeddings, widely used as the input of NLP tasks, are built on the framework of feature transfer. Inspired by parameter transfer, pre-trained CNNs are applied as the backbone of most state-of-the-art CV models. Some recent well-known PTMs are also based on representation transfer and parameter transfer, e.g., ELMo~\cite{peters2018deep} and BERT apply representation transfer and parameter transfer respectively.


\begin{figure*}[t]
		\centering  
		\includegraphics[width=1.0\linewidth]{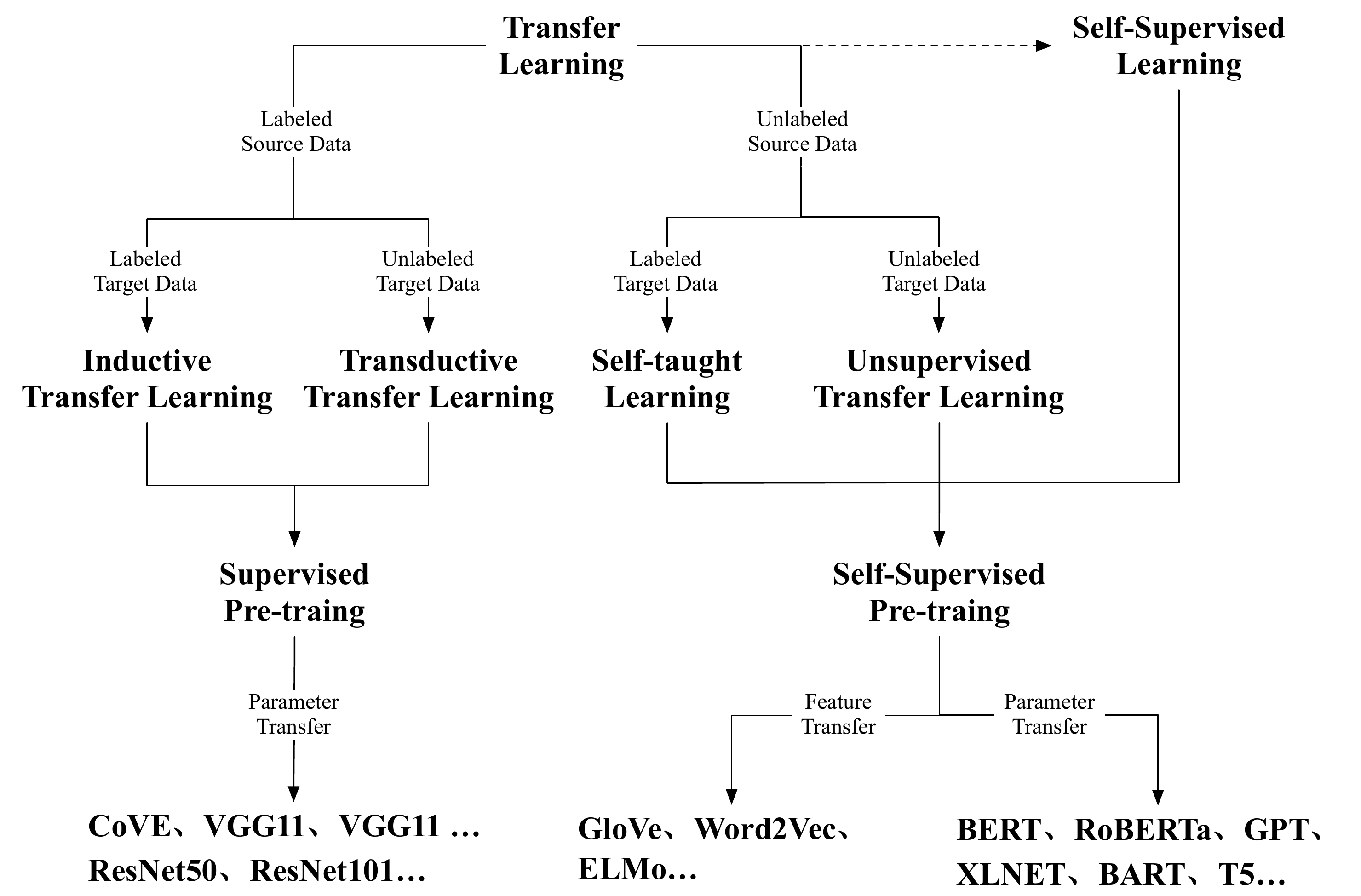}
		\caption{The spectrum of pre-training methods from transfer learning, self-supervised learning to the latest pre-training neural models.}
		\label{fig:history} 
\end{figure*}

Since AlexNet~\cite{krizhevsky2012imagenet}, a series of deep neural networks have been developed for AI tasks. As compared with those conventional machine learning models, deep neural models have more parameters and show better capabilities of fitting complex data. Therefore, from AlexNet to later VGG~\cite{simonyan2014very} and GoogleNet~\cite{szegedy2015going}, the architecture of these neural networks becomes deeper and deeper, and their performance accordingly becomes better and better. Although the network depth is important, training a deep network is not easy, as stacking more network layers inevitably brings the problem of vanishing or exploding gradients~\cite{bengio1994learning}. Besides the gradient issues, model performance may soon meet a ceiling and then degrade rapidly with continually increasing network depths.

By adding normalization to parameter initialization~\cite{lecun2012efficient,saxe2013exact} and hidden states~\cite{ioffe2015batch}, and introducing shortcut connections with residual layers, ResNet~\cite{he2016deep} effectively tackles these problems. As we mentioned before, deep neural networks require large amounts of data for training. To provide sufficient data to train deep models, some large-scale supervised datasets have also been built~\cite{russakovsky2015imagenet, lin2014microsoft,krishna2017visual,chen2015microsoft,cordts2016cityscapes}, and the most representative one is ImageNet. ImageNet contains millions of images divided into thousands of categories, representing a wide variety of everyday objects. Based on the combination of effective model ResNet, informative dataset ImageNet, as well as mature knowledge transfer methods, a wave of pre-training models on labeled data emerges.


The CV community benefits a lot from this wave. By applying ResNet pre-trained on ImageNet as the backbone, various CV tasks have been quickly advanced, like image classification~\cite{he2016deep,lee2015deeply}, object detection~\cite{ren2015faster,sermanet2013overfeat,gidaris2015object}, image segmentation~\cite{long2015fully,zheng2015conditional}, image caption~\cite{vinyals2015show,johnson2016densecap}, visual question answering~\cite{antol2015vqa,gao2015you,xiong2016dynamic}, etc. Utilizing PTMs like ResNet50~\footnote{ResNet50 is a PTM with 50 layers.} has proven to be a crucial step to obtain highly accurate results on most CV tasks. Inspired by the success of PTMs for CV tasks, some NLP researchers also explore supervised Pre-training, and the most representative work is CoVE~\cite{DBLP:conf/nips/McCannBXS17}. CoVE adopts machine translation as its pre-training objective. After pre-training, the encoder of source languages can work as a powerful backbone for downstream NLP tasks.

\subsection{Self-Supervised Learning and Self-Supervised Pre-Training}

As shown in Figure~\ref{fig:history}, transfer learning can be categorized under four sub-settings, inductive transfer learning~\cite{lawrence2004learning,mihalkova2007mapping,evgeniou2007multi}, transductive transfer learning~\cite{shimodaira2000improving,zadrozny2004learning,daume2006domain}, self-taught learning~\cite{raina2007self,dai2008self}~\footnote{Self-study learning can be viewed as a variant of inductive transfer learning without available labeled data}, and unsupervised transfer learning~\cite{wang2008transferred}. 

Among these four settings, the inductive and transductive settings are the core of research, as these two settings aim to transfer knowledge from supervised source tasks to target tasks. Although supervised learning is always one of the core issues of machine learning research, the scale of unlabeled data is much larger than that of manually labeled data. Recently, more and more researchers have noticed the importance of large-scale unlabeled data and are committed to extracting information from unlabeled data. Self-supervised learning has been proposed to extract knowledge from large-scale unlabeled data by leveraging input data itself as supervision. 

Self-supervised learning and unsupervised learning have many similarities in their settings. To a certain extent, self-supervised learning can be regarded as a branch of unsupervised learning because they both apply unlabeled data. However, unsupervised learning mainly focuses on detecting data patterns (e.g., clustering, community discovery, and anomaly detection), while self-supervised learning is still in the paradigm of supervised settings (e.g., classification and generation)~\cite{liu2020self}. 

The development of self-supervised learning makes it possible to perform pre-training on large-scale unsupervised data. Compared to supervised pre-training working as the cornerstone of CV in the deep learning era, self-supervised pre-training allows for huge advances in the field of NLP. Although some supervised pre-training methods like CoVE have achieved promising results on NLP tasks, it is nearly impossible to annotate a textual dataset as large as ImageNet, considering annotating textual data is far more complex than annotating images. Hence, applying self-supervised learning to utilize unlabeled data becomes the best choice to pre-train models for NLP tasks. The recent stunning breakthroughs in PTMs are mainly towards NLP tasks, more specifically pre-trained language models.

The early PTMs for NLP tasks exist in the form of well-known word embeddings~\cite{collobert2008unified,mikolov2013distributed,pennington2014glove}, which apply self-supervised methods to transform words into distributed representations. As these pre-trained word representations capture syntactic and semantic information in the text, they are often used as input embeddings and initialization parameters for NLP models and offer significant improvements over random initialization parameters~\cite{turian2010word}. Since these word-level models often suffer from the word polysemy, \citet{peters2018deep} further adopt a sequence-level neural model to capture complex word features across different linguistic contexts and generates context-aware word embeddings. Using word embeddings as the input of neural models has almost become the common mode for NLP tasks. 

After \citet{vaswani2017attention} propose Transformers to deal with sequential data, PTMs for NLP tasks have entered a new stage, because it is possible to train deeper language models compared to conventional CNNs and RNNs. Different from those word-level PTMs used as input features, the Transformer-based PTMs such as GPT and BERT can be used as the model backbone of various specific tasks. After pre-training these Transformer-based PTMs on large-scale textual corpora, both the architecture and parameters of PTMs can serve as a starting point for specific NLP tasks, i.e., just fine-tuning the parameters of PTMs for specific NLP tasks can achieve competitive performance. So far, these Transformer-based PTMs have achieved state-of-the-art results on almost all NLP tasks. Inspired by GPT and BERT, many more effective PTMs for NLP tasks have also been proposed, like XLNET~\cite{yang2019xlnet}, RoBERTa~\cite{liu2019roberta}, BART~\cite{lewis2020bart}, and T5~\cite{raffel2020exploring}.

With the recent advance of PTMs for NLP tasks, applying Transformer-based PTMs as the backbone of NLP tasks has become a standard procedure. Motivated by the success of self-supervised learning and Transformers in NLP, some researchers explore self-supervised learning~\cite{wu2018unsupervised,chen2020simple,chen2020exploring,he2020momentum} and Transformers~\cite{carion2020end,liu2021swin} for CV tasks. These preliminary efforts have shown that self-supervised learning and Transformers can outperform conventional supervised CNNs. Furthermore, Transformer-based multimodal PTMs~\cite{lu2019vilbert,li2019visualbert,tan2019lxmert} have also been proposed and shown promising results. After the last wave of supervised pre-training, self-supervised pre-training has become the focus of current AI research.

Looking back at the pre-training in the AI spectrum, it is not difficult to find that pre-training has been developed for decades, focusing on how to acquire versatile knowledge for various downstream tasks. Next, we will comprehensively introduce the latest breakthroughs of PTMs in this wave of self-supervised pre-training. Considering that almost all the latest PTMs are related to pre-trained language models, ``PTMs'' in the following sections refers to pre-trained language models or multimodal models. For those conventional PTMs based on supervised pre-training, we refer to the papers of \citet{he2019rethinking} and \citet{zoph2020rethinking}.

\section{Transformer and Representative PTMs}
\label{sec:plm}
\begin{figure*}[t]
		\centering  
		\includegraphics[width=1.0\linewidth]{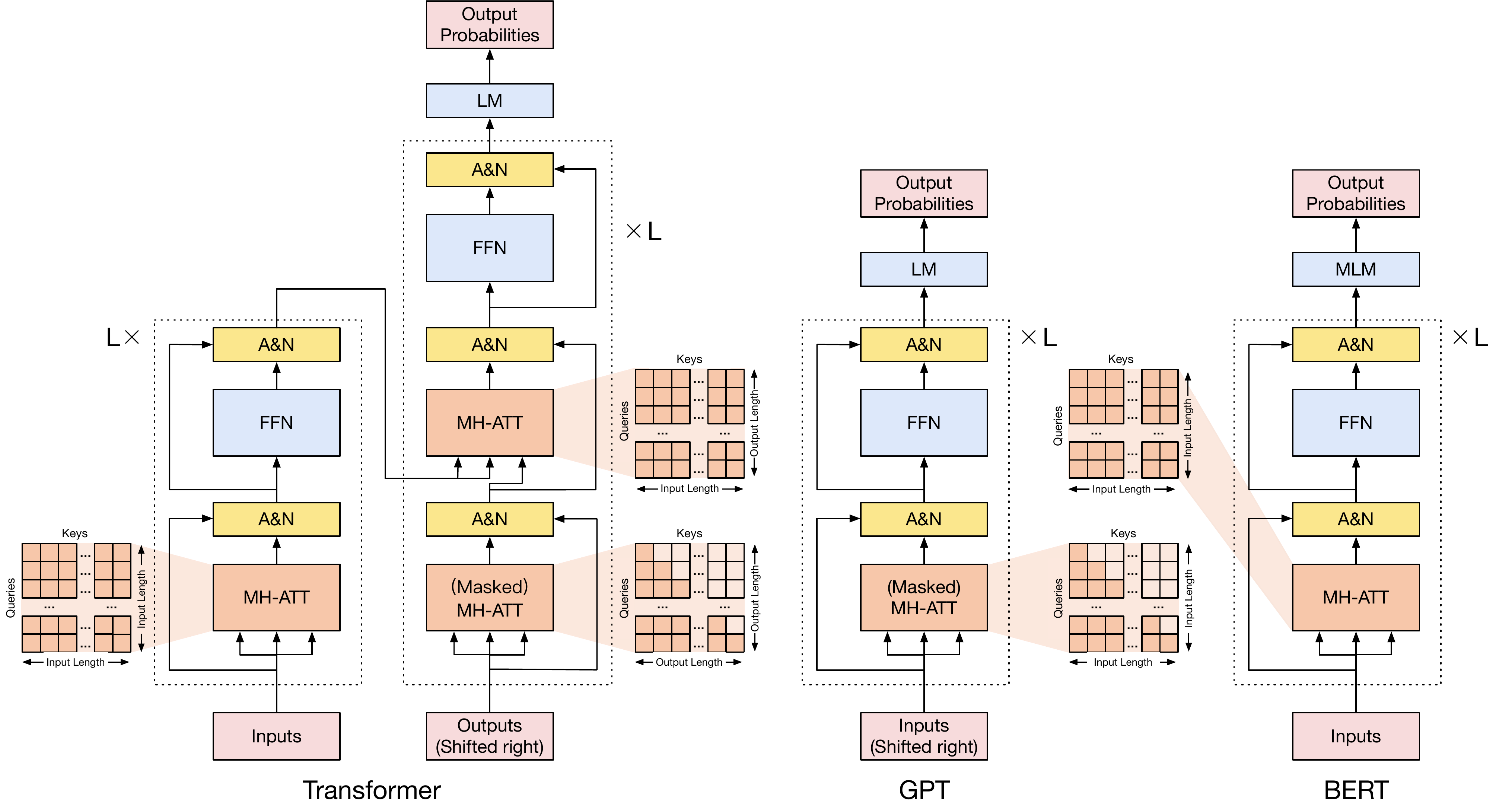}
		\caption{The architecture of Transformer, GPT, and BERT.}
		\label{Transformer_GPT_BERT} 
\end{figure*}

As we mentioned before, the key to the success of recent PTMs is an integration of self-supervised learning and Transformer. Hence, this section begins with the dominant basic neural architecture, Transformer. Then, we will introduce two landmark Transformer-based PTMs, GPT and BERT. These two PTMs respectively use autoregressive language modeling and autoencoding language modeling as pre-training objectives. All subsequent PTMs are variants of these two models. The final part of this section gives a brief review of typical variants after GPT and BERT to reveal the recent development of PTMs.

\subsection{Transformer}

Before Transformer, RNNs have long been a typical tool for processing sequential data, especially for processing natural languages. As RNNs are equipped with sequential nature, they read a word at each time step in order. For each word, RNNs refer to all hidden states of its previous words to process it. Such a mechanism is considered to be difficult to take advantage of the parallel capabilities of high-performance computing devices such as GPUs and TPUs.

As shown in Figure~\ref{Transformer_GPT_BERT}, Transformer is a non-recurrent sequence-to-sequence (seq2seq) architecture consisting of an encoder and a decoder. The encoder and decoder of a Transformer are both stacked by several identical blocks. Each encoder block is composed of a multi-head self-attention layer and a position-wise feed-forward layer. Compared with the encoder block, each decoder block has an additional cross-attention layer since the decoder requires to consider the output of the encoder as a context for generation. Between neural layers, residual connection~\cite{he2016deep} and layer normalization~\cite{ba2016layer} are employed, making it possible to train a deep Transformer. 

\vpara{Attention Layer.} Self-attention layers are the key to the success of Transformer. Formally, given a query set $\mathcal{Q} = \{\mathbf{q}_1, \ldots, \mathbf{q}_n\}$, a key set $\mathcal{K} = \{\mathbf{k}_1, \ldots, \mathbf{k}_m \}$, a value set $\mathcal{V} = \{\mathbf{v}_1, \ldots, \mathbf{v}_m\}$, each query vector $\mathbf{q}_i \in \mathbb{R}^{d_k}$, each key vector $\mathbf{k}_i \in \mathbb{R}^{d_k}$, and each value vector $\mathbf{v}_i \in \mathbb{R}^{d_v}$, the scaled dot-product attention is defined as 
\begin{equation}
\begin{aligned}
& \{\mathbf{h}_1,  \ldots, \mathbf{h}_n\} = \texttt{ATT}(\mathcal{Q}, \mathcal{K}, \mathcal{V}),\\
& \mathbf{h}_i = \sum_{j=1}^{m} a_{ij} \mathbf{v}_j,\\
& a_{ij} = \frac{\exp ( \texttt{ATT-Mask}( 
\frac{\mathbf{q}_i \cdot \mathbf{k}_j}{\sqrt{d_k}} ) )}{\sum_{l=1}^{m} \exp (\texttt{ATT-Mask}( 
\frac{\mathbf{q}_i \cdot \mathbf{k}_l}{ \sqrt{d_k} }) )}.
\end{aligned}
\end{equation}
Intuitively, $\mathcal{Q}$ is the set of vectors to calculate the attention for, $\mathcal{K}$ is the set of vectors to calculate the attention against. As a result of dot-product multiplication, we can get the weight $a_{ij}$ to indicate how attended the query vector $\mathbf{q}_i$ against the key vector $\mathbf{k}_j$. Finally, we can calculate the weighted mean of value vectors as the final result of the attention layer. Note that, the masking function $\texttt{ATT-Mask}(\cdot)$ is used to restrict which key-value pairs each query vector can attend. If we do not want $\mathbf{q}_{i}$ to attend $\mathbf{k}_{j}$, $\texttt{ATT-Mask}(x) = -\infty$, otherwise $\texttt{ATT-Mask}(x) = x$.

By respectively packing $\mathcal{Q}, \mathcal{K}, \mathcal{V}$ into matrix representations $\mathbf{Q}\in \mathbb{R}^{n \times d_k}, \mathbf{K} \in \mathbb{R}^{m \times d_k}, \mathbf{V}\in\mathbb{R}^{m \times d_v}$, the attention can be simplified to
\begin{equation}
\begin{aligned}
    & \mathbf{H} = \texttt{ATT} (\mathbf{Q},\mathbf{K},\mathbf{V}) = \mathbf{A}\mathbf{V},\\
    & \mathbf{A}  = \texttt{Softmax}(\texttt{ATT-Mask}(\frac{\mathbf{Q}\mathbf{K}^\top}{\sqrt{d_k}})),
\end{aligned}
\end{equation}
where $\texttt{Softmax}(\cdot)$ is applied in a row-wise manner, $\mathbf{A}\in \mathbb{R}^{n \times m}$ is the attention matrix, $\mathbf{H}\in \mathbb{R}^{n \times d_v}$ is the result. 

Instead of using the vanilla scaled dot-product attention, Transformer applies a multi-head attention layer defined as follows,
\begin{equation}
\begin{aligned}
     \mathbf{H} &= \texttt{MH-ATT}  (\mathbf{Q},\mathbf{K},\mathbf{V}) \\
    &= \texttt{Concat}(\mathbf{H}_1,\ldots,\mathbf{H}_h)\mathbf{W}^O,\\
     \mathbf{H}_i &= \texttt{ATT}(\mathbf{Q}\mathbf{W}_i^Q, \mathbf{K}\mathbf{W}_i^K, \mathbf{V}\mathbf{W}_i^V),
\end{aligned}
\end{equation}
where $h$ is the head number. $\mathbf{W}_i^Q$, $\mathbf{W}_i^K$, $\mathbf{W}_i^V$ are respectively used to project the input $\mathbf{Q}$, $\mathbf{K}$, $\mathbf{V}$ into the feature space of the $i$-th head attention. After concatenating all head outputs by $\texttt{Concat}(\cdot)$, the multi-head attention layer applies $\mathbf{W}^O$ to project the concatation into the final output space. 

\vpara{Position-Wise Feed-Forward Layer.} Besides attention layers, each block of Transformer also contains a position-wise feed-forward layer. Given the packed input matrix $\mathbf{X} \in \mathbb{R}^{n \times d_i}$ indicating a set of input vectors, $d_i$ is the vector dimension, a position-wise feed-forward layer is defined as 
\begin{equation}
    \mathbf{H} = \texttt{FFN}(\mathbf{X}) = \sigma(\mathbf{X}\mathbf{W}_1+\mathbf{b}_1)\mathbf{W}_2+\mathbf{b}_2,
\end{equation}
where $\sigma(\cdot)$ is the activation function (usually the ReLU function). $\mathbf{W}_1 \in \mathbb{R}^{d_i \times d_f}$, $\mathbf{b}_1 \in \mathbb{R}^{d_f}$, $\mathbf{W}_2\in \mathbb{R}^{d_f \times d_o}$, $\mathbf{b}_2 \in \mathbb{R}^{d_o}$ are all learnable parameters for projection. $\mathbf{H} \in \mathbb{R}^{n \times d_o}$ is the final result of the feed-forward layer. Empirically, $d_i$ is set equal to $d_o$, $d_f$ is set to be much larger than $d_i$ and $d_o$.

\vpara{Residual Connection and Normalization} As we mentioned before, Transformer applies residual connection and layer normalization between various neural layers, making the architecture of Transformer possible to be deep. Formally, given a neural layer $f(\cdot)$, the residual connection and normalization layer is defined as
\begin{equation}
\mathbf{H} = \texttt{A\&N}(\mathbf{X}) = \texttt{LayerNorm}(f(\mathbf{X})+\mathbf{X}),
\end{equation}
where $\texttt{LayerNorm}(\cdot)$ denotes the layer normalization operation.

\begin{figure}[t]
		\centering  
		\includegraphics[width=0.8\linewidth]{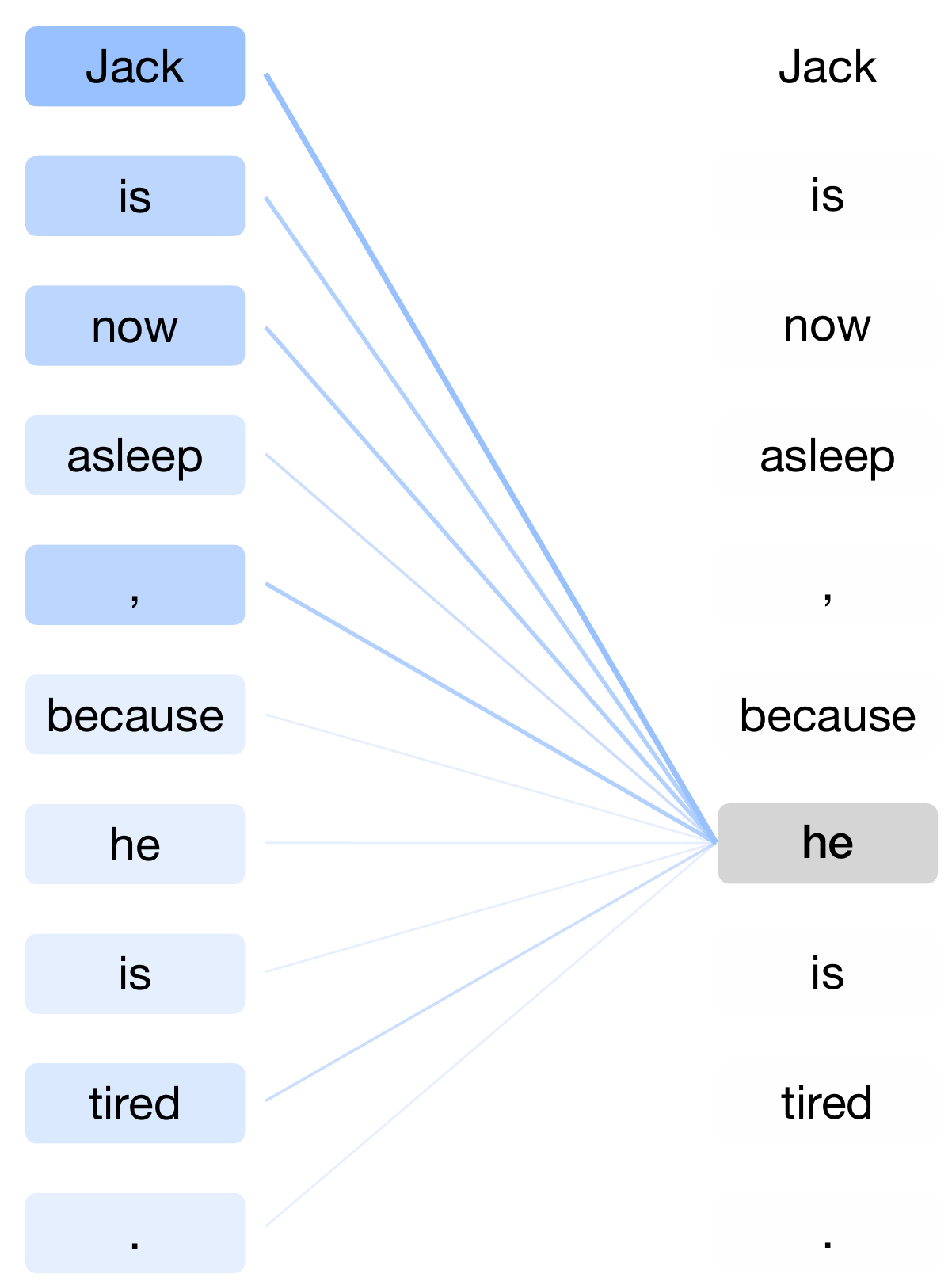}
		\caption{An illustration of the self-attention mechanism of Transformer. The figure shows the self-attention results when encoding the word ``he'', where the darker the color of the square is, the larger the corresponding attention score is.}
		\label{self-att} 
\end{figure}

As shown in Figure~\ref{Transformer_GPT_BERT}, there are three variants of the multi-head attention in Transformer: 

(1) Self-attention is used in the encoder, which uses the output of the previous layer as $\mathbf{Q}$, $\mathbf{K}$, $\mathbf{V}$. In the encoding phase, given a word, the self-attention computes its attention scores by comparing it with all words in the input sequence. And such attention scores indicate how much each of the other words should contribute to the next representation of the given word. We give an example in Figure~\ref{self-att}, where the self-attention accurately captures the referential relationships between ``Jack'' and ``he'', generating the highest attention score. 

(2) Masked self-attention is used in the decoder, whose attention matrix satisfies $\mathbf{A}_{ij}=0, i > j$. This attention is beneficial to autoregressive language modeling. In the decoding phase, the self-attention is similar to the encoding, except that it only decodes one representation from left to right at one time. Since each step of the decoding phase only consults the previously decoded results, we thus require to add the masking function into the self-attention.

(3) Cross-attention is also used in the decoder, which uses the output of the previous decoder block as $\mathbf{Q}$ as well as the output of the encoder as $\mathbf{K}$ and $\mathbf{V}$. Such a procedure is essentially an aggregation of the information of the whole input sequence, and it will be applied to all the words to generate in the decoding phase. Taking advantage of the input context is of great significance to some seq2seq tasks such as machine translation and text summarization.

For more details of Transformer, please refer to its original paper~\cite{vaswani2017attention} and the survey paper~\cite{lin2021surveytransformers}.
Due to the prominent nature, Transformer gradually becomes a standard neural structure for natural language understanding and generation. Moreover, it also serves as the backbone neural structure for the subsequently derived PTMs. Next, we will introduce two landmarks that completely open the door towards the era of large-scale self-supervised PTMs, GPT and BERT. In general, GPT is good at natural language generation, while BERT focuses more on natural language understanding.

\begin{figure*}[t]
		\centering  
		\includegraphics[width=1.0\linewidth]{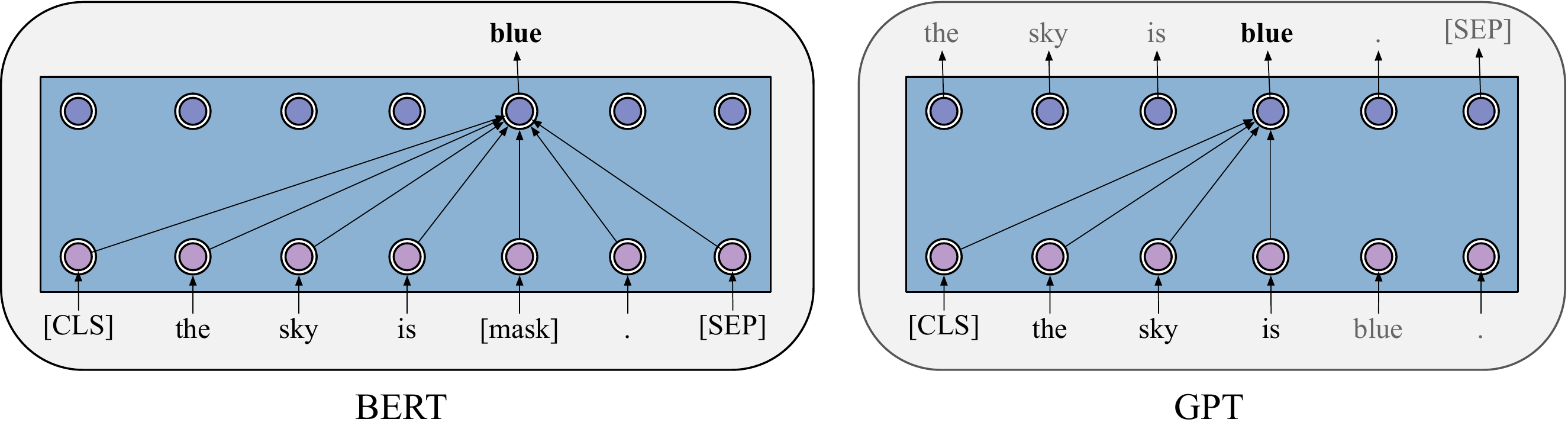}
		\caption{The difference between GPT and BERT in their self-attention mechanisms and pre-training objectives.}
		\label{fig:bertvsgpt} 
\end{figure*}

\subsection{GPT}

As introduced in Section~\ref{sec:background}, PTMs typically consist of two phases, the pre-training phase and the fine-tuning phase. Equipped by the Transformer decoder as the backbone~\footnote{Since GPT uses autoregressive language modeling for the pre-training objective, the cross-attention in the original Transformer decoder is removed.}, GPT applies a generative pre-training and a discriminative fine-tuning. Theoretically, compared to precedents of PTMs, GPT is the first model that combines the modern Transformer architecture and the self-supervised pre-training objective. Empirically, GPT achieves significant success on almost all NLP tasks, including natural language inference, question answering, commonsense reasoning, semantic similarity and classification.

Given large-scale corpora without labels, GPT optimizes a standard autoregressive language modeling, that is, maximizing the conditional probabilities of all the words by taking their previous words as contexts. In the pre-training phase of GPT, the conditional probability of each word is modeled by Transformer. As shown in Figure~\ref{Transformer_GPT_BERT} and Figure~\ref{fig:bertvsgpt}, for each word, GPT computes its probability distributions by applying masked multi-head self-attention operations over its previous words. Formally, given a corpus consisting of tokens $\mathcal{X} = \{x_0, x_1, \ldots, x_n, x_n+1\}$, GPT applies a standard language modeling objective by maximizing the following log-likelihood:
\begin{equation}
    \mathcal{L}(\mathcal{X}) = \sum_{i=1}^{n+1} \log P(x_i|x_{i-k},...,x_{i-1}; \Theta),
\end{equation}
where $k$ is the window size, the probability $P$ is modeled by the Transformer decoder with parameters $\Theta$, $x_0$ is the special token $\texttt{[CLS]}$, $x_{n+1}$ is the special token $\texttt{[SEP]}$. 
    
The adaptation procedure of GPT to specific tasks is fine-tuning, by using the pre-trained parameters of GPT as a start point of downstream tasks. In the fine-tuning phase, passing the input sequence through GPT, we can obtain the representations of the final layer of the GPT Transformer. By using the representations of the final layer and task-specific labels, GPT optimizes standard objectives of downstream tasks with simple extra output layers. As GPT has hundreds of millions of parameters, it is trained for 1 month on 8 GPUs, which is fairly the first ``large-scale'' PTM in the history of NLP. And undoubtedly, the success of GPT pave the way for the subsequent rise of a series of large-scale PTMs. In the next part, we will introduce another most representative model BERT.

\begin{figure*}[t]
		\centering  
		\includegraphics[width=1.0\linewidth]{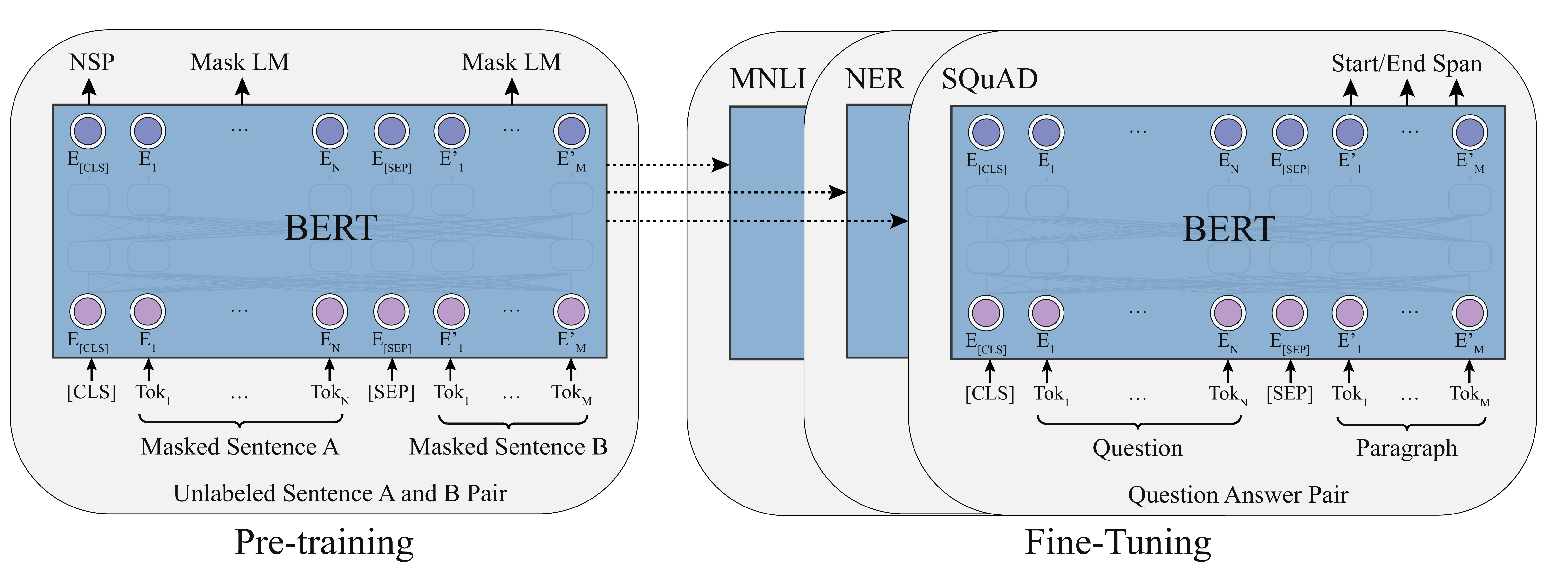}
		\caption{The pre-training and fine-tuning phases for BERT.}
		\label{BERT} 
\end{figure*}

\subsection{BERT}
	
The emergence of BERT has also greatly promoted the development of the PTM field. Theoretically, compared with GPT, BERT uses a bidirectional deep Transformer as the main structure. There are also two separate stages to adapt BERT for specific tasks, pre-training and fine-tuning (see Figure~\ref{Transformer_GPT_BERT} and Figure~\ref{BERT}). 

In the pre-training phase, BERT applies autoencoding language modeling rather than autoregressive language modeling used in GPT. More specifically, inspired by cloze~\cite{taylor1953cloze}, the objective masked language modeling (MLM) is designed. As shown in Figure~\ref{fig:bertvsgpt}, in the procedure of MLM, tokens are randomly masked with a special token $\texttt{[MASK]}$, the objective is to predict words at the masked positions with contexts. Compared with standard unidirectional autoregressive language modeling, MLM can lead to a deep bidirectional representation of all tokens. Formally, given a corpus consisting of tokens $\mathcal{X} = \{x_0, x_1, \ldots, x_n, x_n+1\}$, BERT randomly masks $m$ tokens in $\mathcal{X}$ and then maximizes the following log-likelihood:
\begin{equation}
    \mathcal{L}(\mathcal{X}) = \sum_{i=1}^{m} \log P(\texttt{[Mask]}_i=y_i|\tilde{\mathcal{X}}; \Theta),
\end{equation}
where the probability $P$ is modeled by the Transformer encoder with parameters $\Theta$, $\tilde{\mathcal{X}}$ is the result after masking some tokens in $\mathcal{X}$, $\texttt{[Mask]}_i$ is the $i$-th masked position, and $y_i$ is the original token at this position.

Besides MLM, the objective of next sentence prediction (NSP) is also adopted to capture discourse relationships between sentences for some downstream tasks with multiple sentences, such as natural language inference and question answering. For this task, a binary classifier is used to predict whether two sentences are coherent. In the pre-training phase, MLM and NSP work together to optimize the parameters of BERT.

After pre-training, BERT can obtain robust parameters for downstream tasks. By modifying inputs and outputs with the data of downstream tasks, BERT could be fine-tuned for any NLP tasks. As shown in Figure~\ref{BERT}, BERT could effectively handle those applications with the input of a single sentence or sentence pairs. For the input, its schema is two sentences concatenated with the special token $\texttt{[SEP]}$, which could represent: (1) sentence pairs in paraphrase, (2) hypothesis-premise pairs in entailment, (3) question-passage pairs in question answering, and (4) a single sentence for text classification or sequence tagging. For the output, BERT will produce a token-level representation for each token, which can be used to handle sequence tagging or question answering, and the special token $\text{[CLS]}$ can be fed into an extra layer for classification. After GPT, BERT has further achieved significant improvements on 17 different NLP tasks, including SQuAD (better than human performance), GLUE (7.7\% point absolute improvements), MNLI (4.6\% point absolute improvements), etc.

\subsection{After GPT and BERT}

\begin{figure*}[t]
    \centering  
    \includegraphics[width=1.0\linewidth]{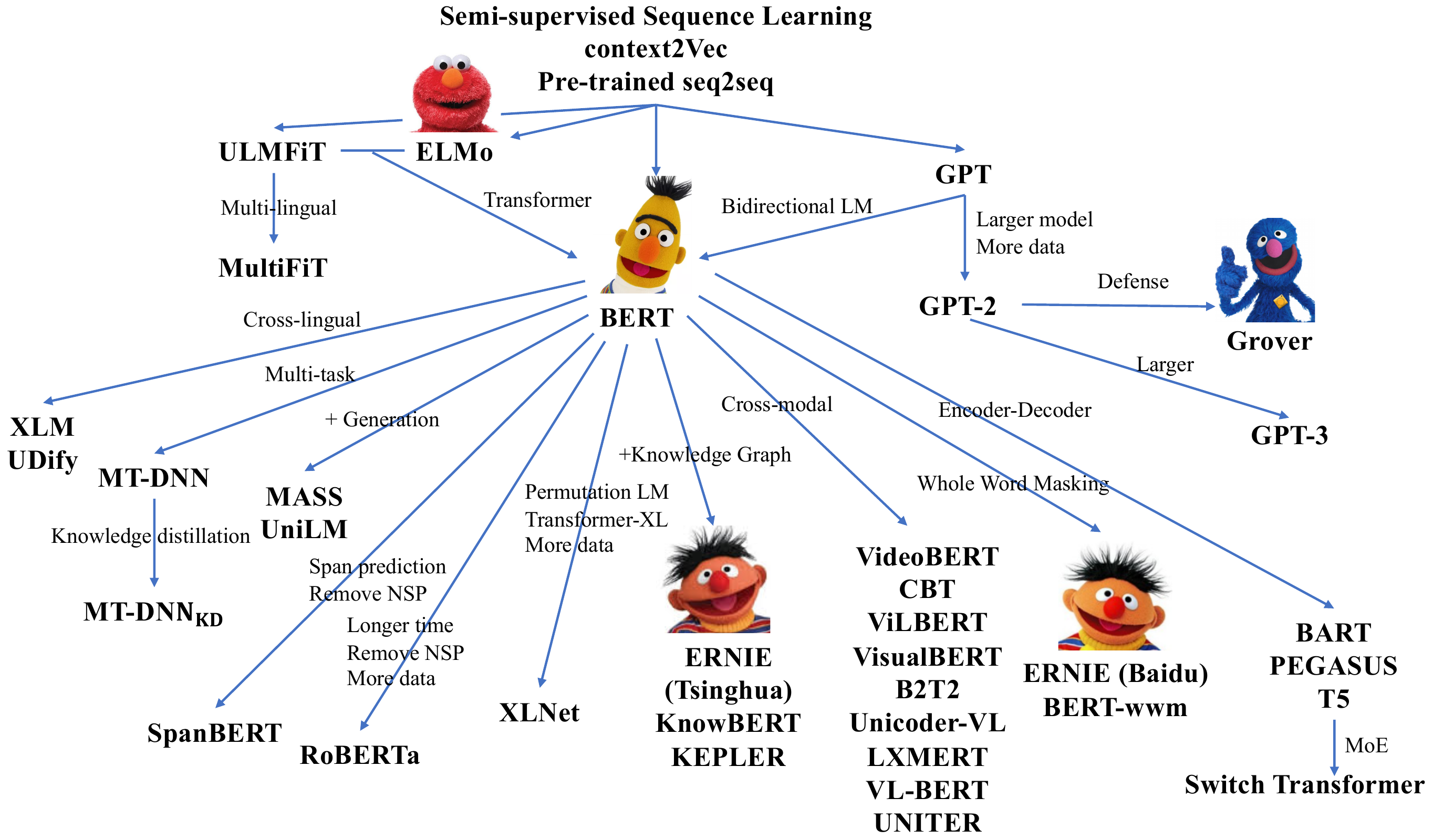}
    \caption{The family of recent typical PTMs, including both pre-trained language models and multimodal models.}
    \label{family} 
\end{figure*}

After GPT and BERT, some of their improvements have been proposed, such as RoBERTa and ALBERT. RoBERTa\cite{liu2019roberta} is one of the success variants of BERT, which mainly has four simple and effective changes: (1) Removing the NSP task; (2) More training steps, with bigger batch size and more data; (3) Longer training sentences; (4) Dynamically changing the $\texttt{[MASK]}$ pattern. RoBERTa achieves impressive empirical results on the basis of BERT. Moreover, RoBERTa has pointed out that the NSP task is relatively useless for the training of BERT. ALBERT\cite{lan2019albert} is another important variant of BERT, which provides several interesting observations on reducing parameters. First, it factorizes the input word embedding matrix into two smaller ones. Second, it enforces parameter-sharing between all Transformer layers to significantly reduce parameters. Third, it proposes the sentence order prediction (SOP) task to substitute BERT's NSP task. As a sacrifice to its space efficiency, ALBERT has a slower fine-tuning and inference speed.

As shown in Figure~\ref{family}, besides RoBERTa and ALBERT, there are various PTMs being proposed in recent years towards better capturing knowledge from unlabeled data. Some work improves the model architectures and explores novel pre-training tasks, such as XLNet~\cite{yang2019xlnet}, UniLM~\cite{dong2019unified}, MASS~\cite{song2019mass}, SpanBERT~\cite{joshi2020spanbert} and ELECTRA~\cite{clark2020electra}. Besides, incorporating rich data sources is also an important direction, such as utilizing multilingual corpora, knowledge graphs, and images. Since the model scale is a crucial success factor of PTMs, researchers also explore to build larger models to reach over hundreds of billions of parameters, such as the series of GPT~\cite{radford2019language, brown2020language}, Switch Transformer~\cite{fedus2021switch}, and meanwhile conduct computational efficiency optimization for training PTMs~\cite{shoeybi2019megatron,rajbhandari2020zero,ren2021zerooffload}. In the following sections, we will further introduce all these efforts for PTMs in detail.

\section{Designing Effective Architectures}
\label{sec:arch}

In this section, we dive into the after-BERT PTMs deeper. The success of Transformer-based PTMs has stimulated a stream of novel architectures for modeling sequences for natural language and beyond. Generally, all the after-BERT Transformer architectures for language pre-training could be categorized according to two motivations: toward \textbf{unified sequence modeling} and \textbf{cognitive-inspired architectures}. 
Besides, we also take a glimpse over other important BERT variants in the third subsection, which mostly focus on improving natural language understanding.

\subsection{Unified Sequence Modeling}

Why is NLP so challenging? One of the fundamental reasons is that it has versatile downstream tasks and applications, which could be generally categorized into three genres:

\begin{itemize}
    \item Natural language understanding: includes grammatical analysis, syntactic analysis, word/sentence/paragraph classification, question answering, factual/commonsense knowledge inference and etc.
    \item Open-ended language generation: includes dialog generation, story generation, data-to-text generation and etc.
    \item Non-open-ended language generation: includes machine translation, abstract summarizing, blank filling and etc.
\end{itemize}

Nevertheless, the differences between them are not so significant. As Feynman's saying goes, ``What I cannot create, I do not understand''. On one hand, a model that can not understand must not fluently generate; on the other hand, we can easily turn understanding tasks into generation tasks~\cite{schick2020s}. Recent studies also show that GPTs can achieve similar and even better performance on understanding benchmarks than BERTs~\cite{liu2021gpt}. The boundary between understanding and generation is vague.

Based on the observation, a bunch of novel architectures has been seeking for unifying different types of language tasks with one PTM. We will take a look over its development and discuss the inspirations they bring towards a unified foundation of natural language processing.

\vpara{Combining Autoregressive and Autoencoding Modeling.} The pioneer work to unify GPT-style unidirectional generation and BERT-style bidirectional understanding is XLNet~\cite{yang2019xlnet}, which proposes the permutated language modeling. The masked-recover strategy in BERT naturally contradicts with its downstream application, where there is no $\texttt{[MASK]}$ in input sentences. XLNet solves the problem by permutating tokens' order in the pre-training and then applying the autoregressive prediction paradigm, which endows XLNet with the ability for both understanding and generation. An important follower of permutation language modeling is MPNet~\cite{song2020mpnet}, which amends the XLNet's discrepancy that in pre-training XLNet does not know the sentence's length while in downstream it knows.

Besides permutated language modeling, another stream would be multi-task training. UniLM~\cite{dong2019unified} proposes to jointly train different language modeling objectives together, including unidirectional, bidirectional, and seq2seq objectives. This can be achieved by changing the attention masks in Transformers. UniLM performs quite well in generative question answering and abstract summarization.

Recently, GLM~\cite{du2021all} proposes a more elegant approach for combining autoregressive and autoencoding. Given a variable-length masked span, instead of providing the number of $\texttt{[MASK]}$ to model as BERT and SpanBERT~\cite{joshi2020spanbert} do, GLM asks Transformer blocks to autoregressively generate the masked tokens. And to preserve the information of $\texttt{[MASK]}$s' number, GLM proposes a 2D positional encoding strategy. GLM is the first model to achieve the best performance on all types of tasks including natural language understanding, conditional generation, and unconditional generation at the same time.

\begin{table}[t]
    \small
    \caption{Three fundamental types of framework and their suitable downstream tasks. ``NLU'' refers to natural language understanding. ``Cond.~Gen.'' and ``Uncond.~Gen.'' refer to conditional and unconditional text generation, respectively. ``\checkmark'' means ``is good at'', ``---'' means ``could be adapted to'', and ``$\times$'' means ``cannot be directly applied to''. We define unconditional generation as the task of generating text without further training as in a standard language model, while conditional generation refers to seq2seq tasks such as text summarization. Taken from~\cite{du2021all}.}
    \begin{tabular}{c|c|c|c}
    \toprule
        Framework & NLU & Cond.~Gen. & Uncond.~Gen.  \\
        \midrule
        Autoregressive & --- & --- & \checkmark \\
        Autoencoding & \checkmark & $\times$ & $\times$ \\
        Encoder-Decoder & --- & \checkmark & --- \\
        \bottomrule
    \end{tabular}
    \label{tab:summary}
\end{table}

\vpara{Applying Generalized Encoder-Decoder.} Before GLM, both encoder structure (e.g., BERT) or decoder structure (e.g., GPT) can not solve an important problem: to fill in blanks with variable lengths~\cite{du2021all,shen2020blank}. The decoder-based models can not make it because they can only generate at the end of the sequence and neither the encoder-based models because the number of $\texttt{[MASK]}$s will leak information. A natural idea is to turn to encoder-decoder architectures originally designed for machine translation, which would produce variable lengths of target sequences conditioned on the sources.

The pioneer of this genre is MASS~\cite{song2019mass}, which introduces the masked-prediction strategy into the encoder-decoder structure. However, MASS does not touch the problem of filling variable-length blanks. T5~\cite{raffel2020exploring} solves the problem by masking a variable-length of span in text with only one mask token and asks the decoder to recover the whole masked sequence. BART~\cite{lewis2020bart} introduces the interesting idea of corrupting the source sequence with multiple operations such as truncation, deletion, replacement, shuffling, and masking, instead of mere masking. There are following works that specify in typical seq2seq tasks, such as PEGASUS~\cite{zhang2020pegasus} and PALM~\cite{bi2020palm}.

However, several challenges lie in front of encoder-decoder architectures. First, the encoder-decoder introduces much more parameters compared to a single encoder/decoder. Although this problem could be alleviated by parameter-sharing of the encoder and decoder, its parameter-efficiency is still doubtful. Second, encoder-decoder structures generally do not perform very well on natural language understanding. Despite reported improvements over similar-sized vanilla BERT, well-trained RoBERTa or GLM encoder performs much better than them.

\subsection{Cognitive-Inspired Architectures}

Is the current Transformer a good enough implementation of human beings' cognitive system? Of course not. Attention mechanism, the core module in the Transformer architecture, is inspired by the micro and atom operation of the human's cognitive system and only responsible for the perceptive function. However, human-level intelligence is far more complex than the mere understanding of the association between different things.

In pursuit for human-level intelligence, understanding the macro architecture of our cognitive functions including decision making, logical reasoning, counterfactual reasoning and working memory~\cite{baddeley1992working} is crucial. In this subsection, we will take a look over the novel attempts inspired by advances of cognitive science, especially on maintainable working memory and sustainable long-term memory.

\vpara{Maintainable Working Memory.} A natural problem of Transformer is its fixed window size and quadratic space complexity, which significantly hinders its applications in long document understanding and generation. 

Despite the bunch of modifications on approximate computing of the quadratic growing point-wise attention~\cite{tay2020efficient}, a question is that we humans do not present such a long-range attention mechanism. As an alternative, cognitive scientists have revealed that humans could maintain a working memory~\cite{baddeley1992working,brown1958some,barrouillet2004time,wharton1994below}, which not only memorizes and organizes but also forgets. The conventional long-short term memory (LSTM) network is an exemplar practice for such a philosophy.

For Transformer-based architectures, the Transformer-XL~\cite{dai2019transformer} is the first to introduce segment-level recurrence and relative positional encoding to fulfill this goal. However, the recurrence only implicitly models the working memory. As a more explicit solution, CogQA~\cite{ding2019cognitive} proposes to maintain a cognitive graph in the multi-hop reading. It is composed of two systems: the System 1 based on PTMs and the System 2 based on GNNs to model the cognitive graph for multi-hop understanding.

A limitation of CogQA is that its use of the System 1 is still based on fixed window size. To endow working memory with the ability to understand long documents, CogLTX~\cite{ding2020cogltx} leverages a MemRecall language model to select sentences that should be maintained in the working memory and task-specific modules for answering or classification.

\vpara{Sustainable Long-Term Memory. } The success of GPT-3 and recent studies on language models' ability in recalling factual knowledge~\cite{Petroni:EMNLP2019,wang2020language,liu2021gpt} has revealed the fact that Transformers can memorize. How does Transformers make it?

In~\citet{lample2019large}, the authors provide some inspiring evidences on how Transformers memorize. They replace the feed-forward networks in a Transformer layer with large key-value memory networks, and find it to work pretty well. This somehow proves that the feed-forward networks in Transformers is equivalent to memory networks.

Nevertheless, the memory capacity in Transformers is quite limited. For human intelligence, besides working memory for deciding and reasoning, the long-term memory also plays a key role in recalling facts and experiences. REALM~\cite{guu2020realm} is a pioneer to explore how to construct a sustainable external memory for Transformers. The authors tensorize the whole Wikipedia sentence by sentence, and retrieve relevant sentences as context for masked pre-training. The tensorized Wikipedia is asynchronously updated for a given number of training steps. RAG~\cite{lewis2020retrieval} extends the masked pre-training to autoregressive generation, which could be better than extractive question answering.

Besides tensorizing the textual corpora, ~\cite{verga2020facts,fevry2020entities} propose to tensorize entities and triples in existing knowledge bases. When entities appear in contexts, they replace entity tokens' embedding in an internal Transformer layer with the embedding from outer memory networks. ~\cite{dhingra2020differentiable,sun2021reasoning} maintain a virtual knowledge from scratch, and propose a differentiable reasoning training objective over it. All of these methods achieve promising improvement on many open-domain question answering benchmarks.

\subsection{More Variants of Existing PTMs}

Besides the practice to unify sequence modeling and construct cognitive-inspired architectures, most current studies focus on optimizing BERT's architecture to boost language models' performance on natural language understanding.

A stream of work aims at improving the masking strategy, which could be regarded as a certain kind of data augmentation~\cite{gu2020train}. SpanBERT~\cite{joshi2020spanbert} shows that masking a continuous random-length span of tokens with a span boundary objective (SBO) could improve BERT's performance. Similar ideas have also been explored in ERNIE~\cite{sun2019ernie,sun2019ernie20} (where a whole entity is masked), NEZHA~\cite{wei2019nezha}, and Whole Word Masking~\cite{cui2019pre}.

Another interesting practice is to change the masked-prediction objective to a harder one. ELECTRA~\cite{clark2020electra} transform MLM to a replace token detection (RTD) objective, in which a generator will replace tokens in original sequences and a discriminator will predict whether a token is replaced.

\section{Utilizing Multi-Source Data}
\label{sec:rich-data}

In this section, we introduce some typical PTMs that take advantage of multi-source heterogeneous data, including multilingual PTMs, multimodal PTMs, and knowledge-enhanced PTMs.

\subsection{Multilingual Pre-Training}

Language models trained on large-scale English corpora have achieved great success in many benchmarks. However, we live in a multilingual world, and training a large language model for each language is not an elegant solution because of the cost and the amount of data required. In fact, although people from all over the world use different languages, they can express the same meaning. This may indicate that semantics is independent of symbol systems. Additionally, some researchers found that they could get even better performance on benchmarks when training one model with several languages comparing with training several monolingual models~\cite{lample2019cross, huang2020m3p}. Hence, training one model to learn multilingual representations rather than monolingual representations may be a better way. 

Before BERT, some researchers have explored multilingual representations. There are mainly two ways to learn multilingual representations. One way is to learn through parameter sharing. For example, training multilingual LSTMs with several language pairs together achieves multilingual translation. Another way is to learn language-agnostic constraints, such as decoupling language representations into language-specific and language-agnostic representations utilizing the WGAN~\cite{arjovsky2017wasserstein} framework. Both of these two ways enable models to be applied to multilingual scenarios, but only for specific tasks. The model in each of them is trained with one specific task from beginning to end, and cross-lingual knowledge cannot be generalized to other tasks. Hence, for any other multilingual tasks, training new models from scratch is still required. Learning new models from scratch needs a large volume of task-specific data.

The appearance of BERT shows that the framework of pre-training with general self-supervised tasks and then fine-tuning on specific downstream tasks is feasible. This motivates researchers to design tasks to pre-train versatile multilingual models. Multilingual tasks could be divided into understanding tasks and generation tasks according to task objectives. Understanding tasks focus on sentence-level or word-level classification, and are of help for downstream classification tasks such as natural language inference~\cite{conneau2018xnli}. Generation tasks focus on sentence generation, and are crucial in downstream generation tasks such as machine translation.

Some understanding tasks are first used to pre-train multilingual PTMs on non-parallel multilingual corpora. For example, multilingual BERT (mBERT) released by~\citet{devlin2019bert} is pre-trained with the multilingual masked language modeling (MMLM) task using non-parallel multilingual Wikipedia corpora in 104 languages. The research conducted by~\citet{pires-etal-2019-multilingual} shows that mBERT has the ability to generalize cross-lingual knowledge in zero-shot scenarios. This indicates that even with the same structure of BERT, using multilingual data can enable the model to learn cross-lingual representations. XLM-R~\cite{conneau2020unsupervised} builds a non-parallel multilingual dataset called CC-100, which supports 100 languages. The scale of CC-100 is much larger than the Wikipedia corpora used by mBERT, especially for those low-resource languages. XLM-R is pre-trained with MMLM as the only task on CC-100 and gets better performance on several benchmarks than mBERT, which indicates that a larger scale of multilingual corpora can bring better performance.

However, the MMLM task cannot well utilize parallel corpora. In fact, parallel corpora are quite important for some NLP tasks such as machine translation. Intuitively, parallel corpora are very helpful to directly learn cross-lingual representations for those sentences in different languages with the same meanings. From this point, XLM~\cite{lample2019cross} leverages bilingual sentence pairs to perform the translation language modeling (TLM) task. Similar to MLM in BERT, TLM combines two semantically matched sentences into one and randomly masks tokens in both parts. Compared with MLM, TLM requires models to predict the masked tokens depending on the bilingual contexts. This encourages models to align the representations of two languages together. 

Besides TLM, there are some other effective methods to learn multilingual representations from parallel corpora. Unicoder~\cite{huang-etal-2019-unicoder} provides two novel pre-training tasks based on parallel corpora: cross-lingual word recovery (CLWR) and cross-lingual paraphrase classification (CLPC). CLWR uses target language embeddings to represent source language embeddings by leveraging attention mechanisms, and its objective is to recover the source language embeddings. This task enables models to learn word-level alignments between different languages. CLPC treats aligned sentences as positive pairs and samples misaligned sentences as negative pairs to perform sentence-level classification, letting models predict whether the input pair is aligned or not. With CLPC, models can learn sentence-level alignments between different languages. ALM~\cite{Yang2020AlternatingLM} automatically generates code-switched sequences from parallel sentences and performs MLM on it, which forces models to make predictions based only on contexts of other languages. InfoXLM~\cite{chi2020infoxlm} analyzes MMLM and TLM from the perspective of information theory, and encourages models to distinguish aligned sentence pairs with misaligned negative examples under the framework of contrastive learning. HICTL~\cite{wei2021on} extends the idea of using contrastive learning to learn both sentence-level and word-level cross-lingual representations. ERNIE-M~\cite{2020arXiv201215674O} proposes back-translation masked language modeling (BTMLM), and expands the scale of parallel corpora through back-translation mechanisms. These works show that leveraging parallel corpora can bring much help towards learning cross-lingual representations. 

Researches have also widely explored generative models for multilingual PTMs. Normally, a generative model consists of a Transformer encoder and a Transformer decoder. For example, MASS~\cite{song2019mass} extends MLM to language generation. It randomly masks a span of tokens in the input sentence and predicts the masked tokens in an autoregressive manner. Denoising autoencoding (DAE) is a typical generation task, which applies noise functions to the input sentence and then restores the original sentence with the decoder. The noise functions of DAE usually contain two operations: replacing a span of tokens with a mask token as well as permuting the order of tokens. mBART \cite{10.1162/tacl_a_00343} extends DAE to support multiple languages by adding special symbols. It adds a language symbol both to the end of the encoder input and the beginning of the decoder input. This enables models to know the languages to be encoded and generated.

Although DAE in mBART~\cite{10.1162/tacl_a_00343} is trained with multiple languages, the encoding input and the decoding output are always in the same language. This leads models to capture spurious correlations between language symbols and generated sentences. In other words, models may ignore the given language symbols and directly generate sentences in the same language of the input. To address this issue, XNLG~\cite{xnlg} proposes the cross-lingual autoencoding (XAE) task. Different from DAE, the encoding input and the decoding output of XAE are in different languages, which is similar to machine translation. In addition, XNLG optimizes parameters in a two-stage manner. It trains the encoder with the MLM and TLM tasks in the first stage. Then, it fixes the encoder and trains the decoder with the DAE and XAE tasks in the second stage. All parameters are well pre-trained by this way, and the gap between pre-training with MLM and fine-tuning with autoregressive decoding is also filled.

\subsection{Multimodal Pre-Training}
\label{sec:multimodal}


Large-scale pre-training and its downstream applications have cascaded impactful research and development with diverse real-world modalities. 
As human beings, we are exposed to different modalities---we see objects, hear sounds and speak languages. Modalities, such as audio, video, image and text, refer to how something happens or is experienced. Recent years have witnessed an upsurging interest in cross-modal tasks that involves multiple modalities. More recently, large-scale PTMs have enhanced research interests in the intersection of multiple modalities, such as the intersection of image and text, or the intersection of video and text. Most of these cross-modal works can be classified as vision and language (V\&L), considering that images and videos belong to vision as well as text and speech (audio) belong to language. Specifically, V\&L tasks can be further divided into image-text-based tasks, video-text-based tasks, and video-audio-based tasks according to their specific modalities being used. In this section, we present an overview of existing works in pre-training on V\&L modalities. Existing cross-modal pre-training PTMs mainly focus on (1) improving model architecture, (2) utilizing more data, and (3) designing better pre-training tasks.


For image-text-based PTMs, most current works are based on the architecture of visual-linguistic BERT. The main challenge lies in the alignment of visual and textual content in a unified semantic space (i.e. V\&L grounding). To this end, there are mainly two kinds of model architecture designs: two-stream and single-stream. As a representative work of two-stream models, ViLBERT~\cite{lu2019vilbert} processes image regions and text tokens with two separate streams, and fuses them with specifically designed co-attention transformer blocks. In comparison, LXMERT~\cite{tan2019lxmert} first processes two modalities separately and then conducts a late fusion with a cross-modality encoder. In single-stream models, such as VisualBERT~\cite{li2019visualbert}, Unicoder-VL~\cite{li2020unicoder}, B2T2~\cite{alberti2019fusion}, the image region features and word embeddings are usually concatenated and fed into a single transformer. Researchers have not reached a consensus on which design is better~\cite{lu2019vilbert,su2019vl} on the V\&L grounding ability. Considering model simplicity and parameter efficiency, current works mainly adopt the single-stream design. 

In cross-modal pre-training, data resources are also of vital significance. The most widely used corpora are image-text pairs collected from web including Conceptual Captions~\cite{sharma2018conceptual}, SBU Captions~\cite{ordonez2011sbu} or existing V\&L datasets designed for specific tasks including COCO~\cite{lin2014microsoft}, Flicker30K~\cite{plummer2015flickr30k}, GQA~\cite{hudson2019gqa}, VQA~\cite{antol2015vqa} and Visual Genome~\cite{krishna2017visual}. Directly increasing the scale of image-text data is useful for better V\&L grounding. UNITER~\cite{chen2020uniter} combines several above-mentioned datasets, resulting in 5.6 million image-text pairs for training. Sufficient training data helps UNITER achieve impressive results on downstream tasks. Similar to UNITER in architecture and pre-training tasks, ImageBERT~\cite{qi2020imagebert} further constructs a dataset containing 10 million web image-text pairs and uses it as a pre-training dataset, leading to a better performance than UNITER on image-text retrieval tasks. In addition to parallel image-text data, VL-BERT~\cite{su2019vl} finds that incorporating extra text-only corpora like BooksCorpus~\cite{zhu2015book} and Wikipedia is helpful for text understanding, especially for tasks with long and complex sentences like visual commonsense reasoning. Different from works using only easily collected data like image-text pairs or textual corpora, \citet{lu202012} identifies the contribution of dedicated datasets by conducting a joint multi-task training on nearly all kinds of V\&L tasks. 

Given data resources, it is also important to design corresponding pre-training tasks or strategies to utilize the information efficiently. For V\&L understanding tasks, the most widely used pre-training tasks are MLM, sentence-image alignment (SIA), masked region classification (MRC), masked region feature regression (MRFR), and directly incorporating downstream tasks. Similar to MLM for NLP, MLM for V\&L aims to recover masked tokens in captions with the help of visual and textual context. SIA is designed to judge whether image-text pairs are matched. MRC can be considered as the visual MLM, requiring V\&L models to predict the categories of masked objects. MRFR further requires V\&L models to recover the visual features of masked object regions. There are also models directly conducting downstream V\&L understanding tasks in the pre-training stage. For example, LXMERT employs VQA as a pre-training task. \citet{lu202012} trains all downstream tasks jointly. To learn the fine-grained alignment between image regions and words, UNITER further proposes a word-region alignment task in the way of Optimal Transport~\cite{chen2020got}, which first finds a sparse matching between image regions and words, and then minimizes the alignment distance. However, most of these works ignore the object tags' function as a kind of explicit bridges between image regions and text tokens. Therefore, Oscar~\cite{li2020oscar} proposes to concatenate the object tags with original image-text pairs as anchors to learn the alignment between V\&L modalities, and designs a new pre-training task for image-tag sequence-caption alignment judgment. In this way, Oscar achieves SOTA results on most V\&L tasks compared with the aforementioned models on both V\&L understanding and generation tasks. Besides pre-training tasks designed for V\&L understanding tasks, there are also some pre-training tasks targeting at V\&L generation tasks. For example, VLP~\cite{zhou2020unified} and X-GPT~\cite{xia2020xgpt} employ seq2seq MLM as their pre-training tasks. 

Instead of designing delicate pre-training tasks, recent works CLIP~\cite{radford2021learning} and WenLan~\cite{huo2021wenlan} choose to grasp the V\&L grounding ability in a simple and holistic regime. They encode images and captions into holistic visual and text representations rather than separated region features and word embeddings, and then only conduct an image-text retrieval task. The success of this kind of holistic alignment can be largely attributed to the enlarging scale of web data, which is $400$ million image-text pairs for CLIP and $30$ million for WenLan.

Previous works mentioned above are specialized for V\&L understanding or only image captioning tasks, but are not capable of image generation. Recently, a bigger step towards conditional image generation is taken by DALLE~\cite{ramesh2021zero} and CogView~\cite{ding2021cogview}. DALLE is the first transformer-based text-to-image PTM, with around $10$ billion parameters. It shows the potential of multi-modal PTMs in bridging the gap between text descriptions and image generation, especially the excellent ability in combining different objects, such as ``an armchair in the shape of an avocado". CogView further improves the numerical precision and training stability by introducing sandwich transformer and sparse attention mechanism, and surpasses the DALLE in Fréchet Inception Distance (FID)~\cite{heusel2017gans} on blurred COCO. 

In addition to image-text PTMs, there are also PTMs for other modalities, such as video and audio. VideoBERT~\cite{sun2019videobert} conducts pre-training on Cooking312K video dataset~\cite{sun2019videobert} and validates the model on zero-shot action classification task and video captioning task.
SpeechBERT~\cite{chuang2019speechbert} first encodes the continuous audio signal into several phonetic-semantic word embeddings, and then uses MLM on both text and audio modalities as pre-training tasks. After pre-training, spoken question answering (SQA) task is used for evaluation.

\begin{figure*}[t]
		\centering  
		\includegraphics[width=0.95\linewidth]{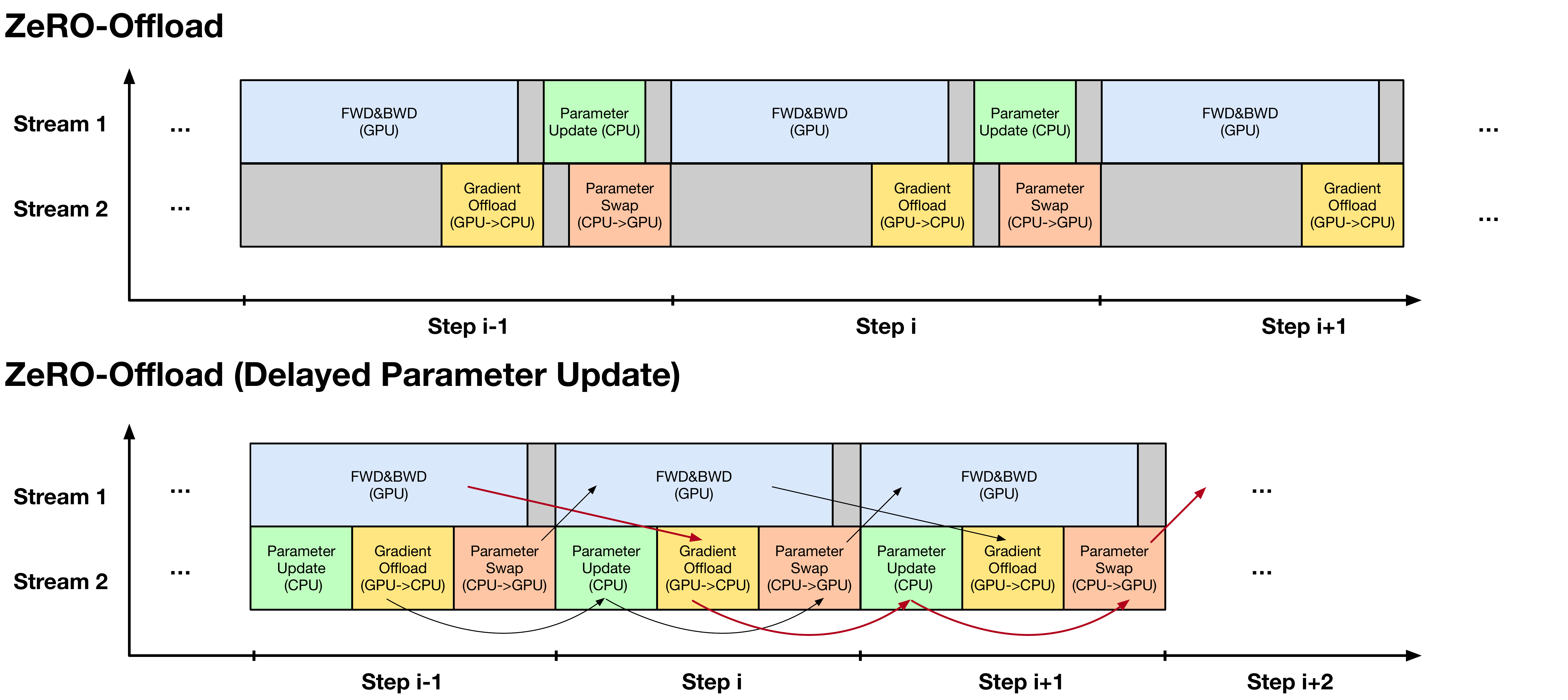}
		\caption{An illustration of ZeRO-Offload and ZeRO-Offload with delayed parameter update.}
		\label{fig:zero} 
\end{figure*}

\subsection{Knowledge-Enhanced Pre-Training}

PTMs can extract plenty of statistical information from large amounts of data. Besides, external knowledge, such as knowledge graphs, domain-specific data and extra annotations of pre-training data, is the outcome of human wisdom which can be a good prior to the modeling of statistics. In this subsection, we classify external knowledge according to the knowledge format and introduce several methods attempting to combine knowledge with PTMs.

The typical form of structured knowledge is knowledge graphs. Many works try to enhance PTMs by integrating entity and relation embeddings~\cite{zhang2019ernie, liu2020k, peters2019knowledge, sun2020colake, rosset2020knowledge,qin2020erica} or their alignments with the text~\cite{xiong2019pretrained, sun2019ernie}. However, real-world knowledge graphs like Wikidata contain more information than entities and relations. \citet{wang2021kepler} pre-train models based on the descriptions of Wikidata entities, by incorporating a language model loss and a knowledge embedding loss together to get knowledge-enhanced representations. Some works regard the paths and even sub-graphs in knowledge graphs as a whole, and directly model them and the aligned text to retain more structural information. Since aligning entities and relations to raw text is often troublesome and can introduce noise in data pre-processing, another line of works~\cite{bosselut2019comet, guan2020knowledge, chen2020kgpt} can directly convert structural knowledge into the serialized text and let models learn knowledge-text alignments by themselves. An interesting attempt is OAG-BERT~\cite{liu2021oag}, which integrates heterogeneous structural knowledge in the open academic graph (OAG)~\cite{zhang2019oag}, which covers 0.7 billion heterogeneous entities and 2 billion relations.

Compared to structured knowledge, unstructured knowledge is more intact but also noisier. How to effectively model this kind of knowledge from the data is also worth being explored. The data of a specific domain or task can be considered as a kind of unstructured knowledge. Many works~\cite{beltagy2019scibert,lee2020biobert} further pre-train the general PTMs on this data to get better domain-specific or task-specific models. Since there are some domain-specific and task-specific human annotations, \citet{ke2020sentilare} incorporate these extra annotations to get better domain-specific and task-specific language representations. For all the above-mentioned works, knowledge is implicitly stored in their model parameters. To model external knowledge in a more interpretable way, some works~\cite{lewis2020retrieval,guu2020realm} design retrieval-based methods to use structured knowledge on downstream tasks. Another kind of works~\cite{wang2020k} can use adapters trained on different knowledge sources with extra annotations to distinguish where the knowledge is from.

\section{Improving Computational Efficiency}
\label{sec:efficiency}

\begin{figure*}[t]
		\centering  
		\includegraphics[width=0.8\linewidth]{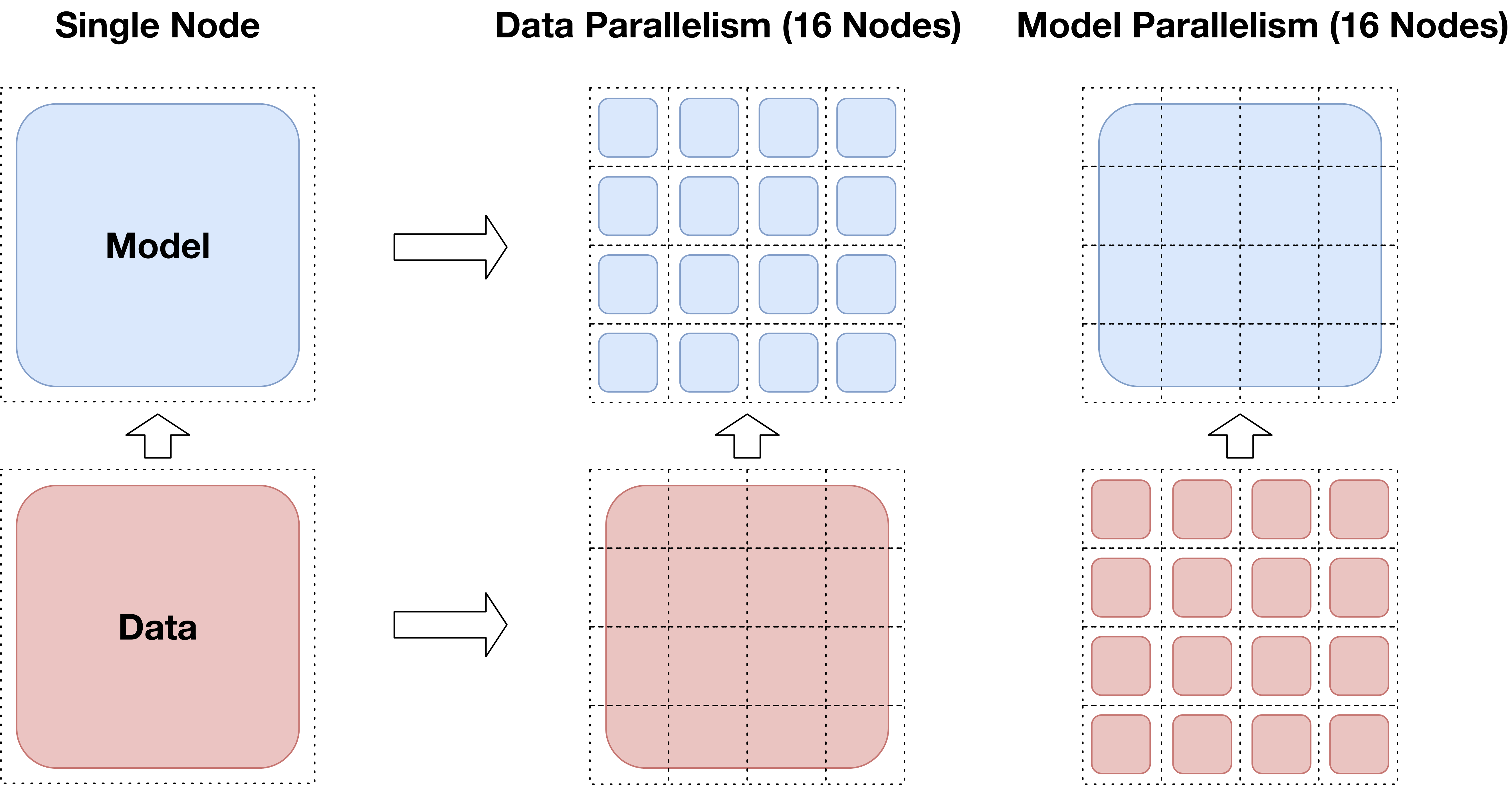}
		\caption{An illustration of the data parallelism and model parallelism with 16 nodes.}
		\label{fig:ppp} 
\end{figure*}

As introduced in Section~\ref{sec:introduction}, a major trend of PTMs is that the number of parameters is getting larger and larger. Increasing the size of a neural network typically improves accuracy, but it also increases the memory and computational requirements for training the model. In this section, we will introduce how to improve computational efficiency from the following three aspects: system-level optimization, efficient learning algorithms, and model compression strategies.

\subsection{System-Level Optimization}

An effective and practical way to reduce computational requirements is system-level optimization towards computational efficiency and memory usage. System-level optimization methods are often model-agnostic and do not change underlying learning algorithms. Therefore, they are widely used in training large-scale PTMs. Generally, these methods can be divided into single-device optimization methods and multi-device optimization ones.

\vpara{Single-Device Optimization.} Current large-scale PTMs usually cost a lot of memory for pre-training. This is mainly due to the redundant representation of floating-point numbers. Modern deep learning systems are mainly based on a single-precision floating-point format (FP32). However, the weights of models usually fall in a limited range, and using a half-precision floating-point format (FP16) can accomplish most of the computation with little precision loss~\cite{pmlr-v37-gupta15}. 

However, in some cases, training models in FP16 may fail because of the floating-point truncation and overflow. To tackle this problem, mixed-precision training methods~\cite{micikevicius2018mixed} have been proposed, which preserve some critical weights in FP32 to avoid the floating-point overflow and use dynamic loss scaling operations to get rid of the floating-point truncation. Sufficient experiments have shown that mixed-precision training methods are more stable than directly training models in FP16. Although mixed-precision training methods can significantly reduce the training time and memory usage, they still face some challenges. When model parameters are not initialized well, mixed-precision methods may still cause unstable training. All these challenges still require to be further explored.

Besides the redundant representation of floating-point numbers, the activation states saved for computing gradients are also redundant. For example, in Transformer-based models, apart from the weights of attention layers and linear layers, computational devices also store the hidden states of each layer for the efficiency of the chain rule used in the gradient back-propagation. As compared with model parameters, these hidden states can consume even much more memory. To handle redundant activation states, gradient checkpointing methods~\cite{rasley2020deepspeed} have been used to save memory by storing only a part of the activation states after forward pass. The discarded activation states are recomputed during the backward steps if necessary.

\begin{figure*}[t]
		\centering  
		\includegraphics[width=1.0\linewidth]{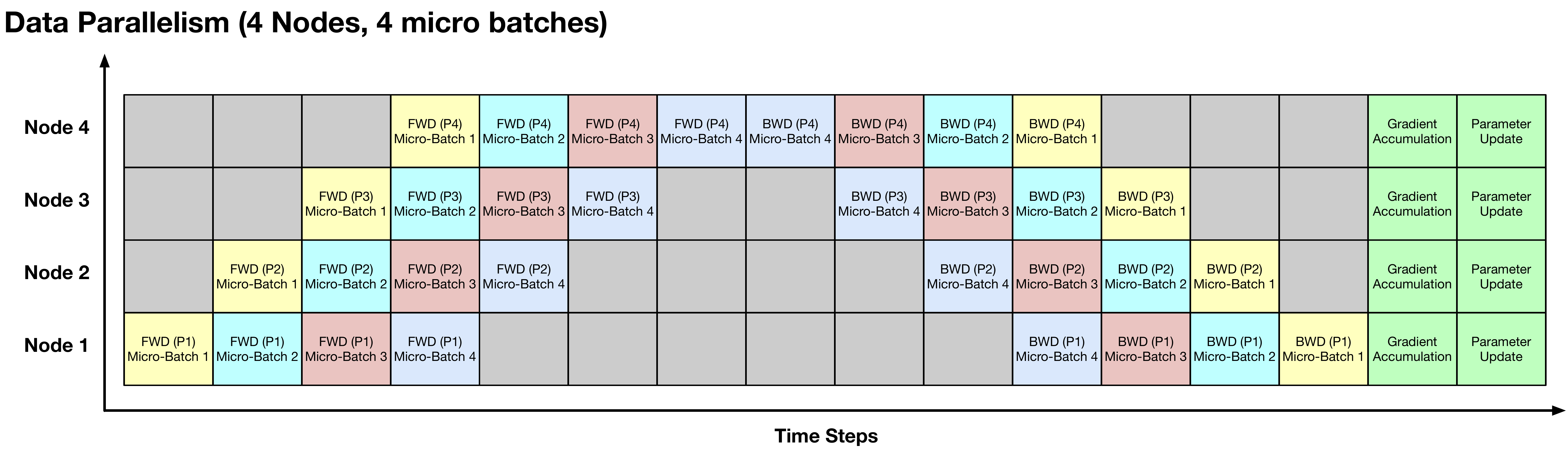}
		\caption{An illustration of the pipeline parallelism with 4 nodes and 4 micro batches.}
		\label{fig:pipeline} 
\end{figure*}

When pre-training recent large-scale PTMs, the memory consumption can be too large to fit in a single GPU. Therefore, some works~\cite{Huang2020SwapAdvisorPD} attempt to store model parameters and activation states with the CPU memory rather than the GPU memory, since the CPU memory is usually much larger. As shown in Figure~\ref{fig:zero}, some works such as ZeRO-Offload~\cite{ren2021zerooffload} design delicate strategies to schedule the swap between the CPU memory and the GPU memory so that memory swap and device computation can be overlapped as much as possible.

\vpara{Multi-Device Optimization.} Recently, distributed training is commonly used in pre-training, where multiple GPUs distributed in many computational nodes are used together to train a single model. Data parallelism~\cite{li2020pytorch} is a simple and effective approach to accelerate training a model. As shown in Figure~\ref{fig:ppp}, when we use data parallelism, a large batch is partitioned to different nodes and thus forward pass can be parallelized. At backward pass, the gradients on different nodes should be aggregated with all-reduce operations to ensure the consistency of parameter optimization, which may introduce additional communication overhead.

When pre-training models with billions to trillions of parameters, traditional data parallelism brings challenges of fitting whole model parameters into a single GPU, even with half-precision or mixed-precision training. Although this problem can be solved by using a GPU with larger memory, the expenses can be hard to afford, limiting the use of PTM by ordinary researchers. Model parallelism is an effective way to tackle this problem~\cite{shazeer-2018-nips}. As shown in Figure~\ref{fig:ppp}, when conducting model parallelism, model parameters can be distributed to multiple nodes. The communication operations between these nodes like reduce-scatter and all-gather guarantee the correctness of forward pass and backward pass. Megatron-LM~\cite{shoeybi2019megatron} adopts model parallelism to Transformer-based PTMs. It splits self-attention heads as well as feed-forward layers into different GPUs, reducing the memory burden of a single GPU. Mesh-Tensorflow~\cite{shazeer-2018-nips} also enables users to split tensors along any tensor dimensions, which can bring more customized options for model parallelism.

Although model parallelism enables different computational nodes to store different parts of model parameters, it has to insert collective communication primitives during both forward pass and backward pass, which can not be overlapped by device computation. On the contrary, the all-reduce collective communication operation in data parallelism usually can be overlapped by the backward computation. As a result, data parallelism is preferred as long as it can conquer the excessive requirement of memory capacity. In the standard implementation of data parallelism, optimizer states are usually copied along different nodes to guarantee synchronized optimization across data parallelism units. This redundancy leads to the additional overhead of GPU memory, especially when models are trained in a mixed-precision manner because the optimizer needs to store 32-bit master states of these models to ensure accuracy. To eliminate the redundancy brought by optimizer states and parameters, ZeRO optimizer~\cite{rajbhandari2020zero} methods equally partition and distribute optimizer states to each node of data parallelism, such that each node only updates the optimizer states corresponding to its partition. At the end of a training step, all optimizer states are gathered across data parallelism nodes.

The above-mentioned model parallelism techniques mainly focus on partitioning and parallelizing matrix operations across different nodes. As shown in Figure~\ref{fig:pipeline}, another effective method for model parallelism is pipeline parallelism, which partitions a deep neural network into multiple layers and then puts different layers onto different nodes. After the computation of each node, the output is sent to the next node where the next layer computation takes place. Since pipeline parallelism only needs to communicate the intermediate activation states between nodes performing adjacent stages of the pipeline, the communication cost is relatively small. Existing pipeline methods include GPipe~\cite{huang2018gpipe} which can send smaller parts of samples within a mini-batch to different nodes, and TeraPipe~\cite{li2021terapipe} which can apply token-level pipeline mechanisms for Transformer-based models to make each token in a sequence be processed by different nodes. Both of these pipeline methods speed up the large-scale PTMs. However, they should be stopped at the end of each batch until the gradient back-propagation is complete, which can lead to pipeline bubbles. 

\subsection{Efficient Pre-Training}

Besides some system-level optimization methods, various efforts have been devoted to exploring more efficient pre-training methods, so that we can pre-train large-scale PTMs with a lower cost solution.

\vpara{Efficient Training Methods.} Conventional pre-training tasks can be sample-inefficient. For example, for MLM which is widely used to pre-train recent PTMs, models are required to predict masked tokens according to contexts. The masked tokens are usually a subset (typically 15\%) of input tokens, i.e., models can only learn from a small set of input tokens. To tackle this problem, ELECTRA~\cite{clark2020electra} applies the replaced token detection task. This task forces models to distinguish whether an input token is replaced by a generator. This task can leverage more supervision information from each sample since all input tokens need to be distinguished. ELECTRA takes much fewer pre-training steps when it reaches similar performance to those MLM models. Furthermore, traditional MLM randomly masks tokens in a document to predict. Since the difficulty of predicting different tokens varies a lot, the random masking strategy makes the training process aimless and inefficient. Therefore, some works selectively mask tokens based on their importance~\cite{gu2020train} or gradients~\cite{chen2020variance} in back-propagation to speed up model training.

Apart from the pre-training tasks, the current pre-training dynamics are also sub-optimal. Recent large-scale PTMs usually require a large batch size. But in an early work~\cite{goyal2017accurate}, researchers find that naively increasing the batch size may cause difficulty in optimization. Therefore, they propose a warmup strategy that linearly increases the learning rate at the beginning of training. This strategy is commonly used in recent large-scale PTMs. Another feature of recent PTMs is that they are usually composed of multiple stacks of a base structure like Transformers. The conventional training paradigm optimizes each layer simultaneously using the same hyper-parameters. However, some recent works study Transformer-based models and claim that different layers can share similar self-attention patterns. Therefore, a shallow model can firstly be trained and then duplicated to construct a deep model~\cite{pmlr-v97-gong19a}. Some layers can also be dropped during training to reduce the complexity of back-propagation and weight update~\cite{NEURIPS2020_a1140a3d}. In addition, \citet{you2017scaling} and \citet{You2020Large} find that adaptively using different learning rates at different layers can also speed up convergence when the batch size is large.   

\vpara{Efficient Model Architectures.} Besides efficient pre-training methods, more variants of model architectures can also reduce the computational complexity to improve the efficiency of training PTMs. For most Transformer-based PTMs, as their input sequence goes longer, their efficiency is limited by the computation of attention weights due to its quadratic time and space complexity of the sequence length. Therefore, many works attempt to reduce the complexity of Transformers. Some works~\cite{peng2021random, choromanski2020rethinking, wang2020linformer, katharopoulos2020transformers} design low-rank kernels to theoretically approximate the original attention weights and result in linear complexity. Some works~\cite{child2019generating} introduce sparsity into attention mechanisms by limiting the view of each token to a fixed size and separating tokens into several chunks so that the computation of attention weights takes place in every single chunk rather than a complete sequence. Compared to predefined chunks, some works~\cite{roy2021efficient, kitaev2020reformer} find that using learnable parameters to assign tokens into chunks results in better performance. Another kind of methods~\cite{guo2019star, lee2019set, beltagy2020longformer, ainslie2020etc, zaheer2020big} combine global and local attention mechanisms, and then use global nodes to gather tokens in a sequence. In this way, the long sequence is compressed into a small number of elements so that we can reduce the complexity.

Keeping the same theoretical computation complexity as the original Transformer, more variants of the model structure can also accelerate the model convergence. Mix-of-experts (MoE) has been proved early~\cite{shazeer2017outrageously} to increase the parameters of deep neural models while keeping the computational overhead nearly unchanged. Recently, Switch Transformers~\cite{fedus2021switch} employ this technique in pre-training. They add multiple experts to each layer of Transformers. During each forward and backward step, they select only one expert for computation, and thus the training and inference time remain similar to the ordinary Transformers without experts. Some experimental results show that MoE-based models converge faster than the ordinary ones due to the significantly larger model capacity brought by multiple experts. Some efficient open-source toolkits~\cite{he2021fastmoe} are also developed to train large-scale MoE-based models.

\subsection{Model Compression}
\label{sec:compression}

Another important approach to improve the efficiency of PTMs is model compression. In this setting, large models are compressed to small ones to meet the demand for faster inference and deployment on resource-constrained devices.

\vpara{Parameter Sharing.} PTMs can be compressed with sharing parameters across similar units. ALBERT~\cite{lan2019albert} uses factorized embedding parameterization and cross-layer parameter sharing to reduce the parameters of PTMs. Using same weights across all Transformer layers, ALBERT achieves a significant parameter reduction based on the BERT model, and meanwhile has the same or even better performance. This indicates that PTMs can be extremely over-parameterized.  

\vpara{Model Pruning.} To take more advantage of the over-parameterized feature of current PTMs, another method to reduce model parameters is model pruning, which cuts off some useless parts in PTMs to achieve accelerating while maintaining the performance. In~\cite{fan2019reducing}, Transformer layers are selectively dropped during training, resulting in a more shallow model during inference. In \cite{DBLP:conf/nips/MichelLN19}, \cite{voita2019analyzing} and \cite{zhang_know_2021}, researchers study the redundancy of the attention heads in Transformers and find that only a small part of them is enough for good performance. Most of these heads can be removed with little impact on the accuracy. Other trials such as CompressingBERT~\cite{DBLP:conf/rep4nlp/GordonDA20} try to prune the weights of attention layers and linear layers to reduce the number of parameters in PTMs, while maintaining the comparable performance to the original model.

\vpara{Knowledge Distillation.} Although ALBERT saves the memory usage of PTMs, its inference time is not significantly decreased since features still need to go through its layers with the same number as the original model. Knowledge distillation aims at training a small model to reproduce the behavior of a large teacher model. The memory usage and the time overhead are both decreased when using a small distilled model for inference. There are some typical works employing knowledge distillation for PTMs, such as DistillBERT~\cite{sanh2019distilbert}, TinyBERT~\cite{jiao2019tinybert}, BERT-PKD~\cite{sun2019patient} and MiniLM~\cite{wang2020minilm}. In these works, a small student model is trained to mimic the output probability, the hidden states, and the attention matrices of a large teacher model during both the pre-training and fine-tuning stages. With knowledge distillation, the modelegy in the teacher model is transferred into the student model, which can lead to increasing performance compared to training a student model alone. However, the knowledge distillation methods mentioned above require the data used for pre-training the teacher model, which is usually not released in consideration of the data copyright and privacy. Moreover, the teacher model needs to forward over the entire pre-training data to produce logits or intermediate representations for knowledge distillation, causing an even longer training time.

\vpara{Model Quantization.} To get a more compressed model, model quantization is also a useful technique, which has been widely explored in some CNN-based models~\cite{stock2019and, polino2018model}. Model quantization refers to the compression of higher-precision floating-point parameters to lower-precision floating-point ones. Conventional PTMs are usually represented in 32 bits or 16 bits, while models after quantization can be in 8 bits or even 1 or 2 bits. For recent Transformer-based models, 8-bit quantization has been proved to be effective for model compression in Q8BERT~\cite{zafrir2019q8bert}, with little impact on the model performance. Despite this, training 1 or 2 Bits models remains challenging due to the significant decrease in model capacity. To alleviate the performance degradation, other methods to preserve the accuracy can also be employed. Q-BERT~\cite{shen2020q} uses mixed-bits quantization in which the parameters with higher Hessian spectrum require higher precision while those parameters with lower Hessian spectrum need lower precision. TernaryBERT~\cite{zhang2020ternarybert} applies knowledge distillation in quantization, forcing low-bit models to imitate full-precision models. Both Q-BERT and TernaryBERT result in ultra low-bit models. However, low-bit representation is a highly hardware-related technique, which means quantization often requires specific hardware and can not generalize to other devices.

\section{Interpretation and Theoretical Analysis}
\label{sec:analysis}

Beyond the superior performance of PTMs on various NLP tasks, researchers also explore to interpret the behaviors of PTMs, including understanding how PTMs work and uncovering the patterns that PTMs capture. These works cover several important properties of PTMs: knowledge, robustness, and structural sparsity/modularity. Moreover, there are some pioneering works on building the theoretical analysis for PTMs.

\subsection{Knowledge of PTMs}

The implicit knowledge captured by PTMs can be roughly divided into two categories: linguistic knowledge and world knowledge.

\vpara{Linguistic Knowledge.} The linguistic knowledge of PTMs attracts most of attentions among all topics of PTMs' interpretation. Compared to conventional neural models such as CNNs and RNNs which have fewer layers and parameters, large-scale PTMs can learn rich linguistic knowledge from massive pre-training data. In order to study PTMs' linguistic knowledge, researcher design several approaches: (1) Representation Probing: Fix the parameters of PTMs and train a new linear layer on the hidden representations of PTMs for a specific probing task. It is the most popular approach because it can be easily adapted to any probing task without particular design. (2) Representation Analysis: Use the hidden representations of PTMs to compute some statistics such as distances or similarities. According to these statistics, we can construct the relation between different words, phrases, or sentences. (3) Attention analysis: similar to representation analysis, attention analysis compute statistics about attention matrices and is more suitable to discover the hierarchical structure of texts. (4) Generation Analysis: Use language models to directly estimate the probabilities of different sequences or words. The target texts could be correct or incorrect in some linguistic phenomenons.

Representation probing have been widely applied to analyze NLP neural models from word embeddings to PTMs~\cite{DBLP:conf/emnlp/Kohn15,DBLP:conf/repeval/EttingerER16,DBLP:conf/emnlp/ShiPK16,DBLP:conf/iclr/AdiKBLG17,DBLP:conf/acl/BaroniBLKC18,DBLP:conf/naacl/HewittM19,DBLP:journals/corr/abs-2008-06788}. \citet{DBLP:conf/naacl/Liu0BPS19} conduct comprehensive probing experiments on 11 linguistic tasks and find that the representations given by large-scale PTMs are competitive compared to previous task-specific models, which indicates that the models have already learned knowledge about tokens, chunks, and pairwise relations. To further investigate how PTMs represent sentence structures about syntactic, semantic, local, and long-range 
information, \citet{tenney2018what} design a new edge probing task and examine PTMs on a broad suite of sub-sentence tasks and show that PTMs have strong ability to encode syntactic informative while they bring little improvement on semantic tasks. Similarly, several works also reveal the strong syntax encoding of PTMs~\cite{DBLP:conf/aaai/VilaresSSG20,DBLP:conf/cogsci/WarstadtB20,DBLP:conf/naacl/HewittM19}. To analyze the function of different layers, \citet{DBLP:conf/acl/JawaharSS19} and \citet{DBLP:conf/acl/TenneyDP19} show that PTMs encode linguistic information with phrase features at the bottom, syntactic features in the middle and semantic features at the top. Compared to non-contextual representations (e.g., word2vec), PTMs' representations are better in encoding sentence-level properties~\cite{DBLP:conf/rep4nlp/MiaschiD20}. Furthermore, \citet{DBLP:journals/pnas/ManningCHKL20} explore to reconstruct the sentence tree structures given by linguists using a linear transformation of PTMs' embeddings and achieve promising results.

Besides representation probing, researchers try to uncover the structure and relation among different representations. \citet{DBLP:conf/iclr/KimCEL20} propose to leverage the concept of Syntactic Distance to construct the constituency trees of sentences from word representations. \citet{DBLP:journals/corr/abs-1906-11511} analyze how the deletion of one word in a sentence changes representations of other words to reveal the influence of one word on other words.

There are also several works on interpreting PTMs via attention matrices. \citet{DBLP:journals/corr/abs-1906-01698} quantitatively evaluate attention matrices for subject-verb agreement and anaphor-antecedent dependencies, and show that PTMs tend to encode positional information in lower layers and capture hierarchical information in higher layers. To better characterize the behaviors of PTMs' attention matrices, \citet{DBLP:journals/corr/abs-1911-12246} propose to take the maximum attention weight and compute the maximum spanning tree as two statistics. Based on the experimental results, they find that fine-tuning has little impact on the self-attention patterns.

Since PTMs can be directly used to generate tokens or estimate the probabilities of different sentences, it is intuitive to construct analysis tasks based on generation~\cite{DBLP:journals/corr/abs-1901-05287}. Perturbed Masking~\cite{DBLP:conf/acl/WuCKL20} recovers syntactic trees from PTMs without any extra parameter and the structure given by PTMs are competitive with a human-designed dependency schema in some downstream tasks. To analysis the gain of pre-training on estimating the probabilities of ungrammatical words, Schijndel~\cite{DBLP:conf/emnlp/SchijndelML19} show that expanding the training corpora yields diminishing returns and the training corpora would need to be unrealistically large to make PTMs match human performance.

\vpara{World Knowledge.} In addition to linguistic knowledge, PTMs also learn rich world knowledge from pre-training, mainly including commonsense knowledge and factual knowledge~\cite{DBLP:conf/aaai/ZhouZCH20,DBLP:conf/aaai/BouraouiCS20}. 

For the commonsense knowledge, Ettinger~\cite{DBLP:journals/tacl/Ettinger20} first evaluates PTMs' knowledge in the aspect of psycholinguists and find that the models perform well in the situation of shared category or role reversal but fail with challenging inferences and role-based event. Then, to extract commonsense from PTMs, \citet{DBLP:conf/emnlp/DavisonFR19} propose to first transform relational triples into masked sentences and then rank these sentences according to the mutual information given by PTMs. In the experiments, the PTM-based extraction method without further training even generalizes better than current supervised approaches. Similarly, \citet{da-kasai-2019-cracking} also find that PTMs have learned various commonsense features in their representation space based on a series of probing tasks. In addition to the commonsense features/attributes, the implicit relations between different attributes are important and \citet{DBLP:conf/cogsci/ForbesHC19} show that current PTMs' representations cannot model the implicit relations well, which requires further exploration. 

For factual knowledge, \citet{Petroni:EMNLP2019} propose to formulate the relational knowledge generation as the completion of fill-in-the-blank statements. According to the experimental results, they find that PTMs significantly outperform previous supervised baselines on this task without any fine-tuning. However, the construction of these fill-in-the-blank statements is non-trivial. To extract more factual knowledge from PTMs, LPAQA~\cite{DBLP:journals/tacl/JiangXAN20} have been propose to automatically search better statements/prompts through mining-based and paraphrasing-based methods. AutoPrompt~\cite{shin2020autoprompt} proposes to train discrete prompts for knowledge probing. In P-tuning~\cite{liu2021gpt}, the authors discover that the better prompts lie in continuous embedding space, rather than discrete space. The P-tuning boosts the P@1 performance on LAMA to 64\%, which is 20\% higher than AutoPrompt. Moreover, \citet{DBLP:conf/emnlp/RobertsRS20} fine-tune PTMs for the task of open-domain question answering and find that fine-tuning can further benefit the knowledge generation of PTMs. However, \citet{DBLP:conf/emnlp/PornerWS20} find that the success of knowledge generation may rely on learning neural stereotypical associations, i.e., a person with an Italian-sounding name will be predicted to Italian by PTMs. For understanding the number in texts, \citet{DBLP:conf/emnlp/WallaceWLSG19} find that ELMo captures numeracy the best for all pre-trained methods, which is a character-based model, but BERT, which uses sub-word units, is less exact. \cite{wang2020language} investigates the knowledge stored in Transformer's feed-forward attention matrices and proposes a framework to construct open knowledge graphs using PTMs.

\subsection{Robustness of PTMs} 

Recent works have identified the severe robustness problem in PTMs using adversarial examples. Adversarial attacks aims to generate new samples, which are mis-classified by models, by small perturbation on the original inputs. For example, PTMs can be easily fooled by synonym replacement ~\cite{Jin2019textfooler,zang2020word, wang2021cline}. Meanwhile, irrelevant artifacts such as form words can mislead the PTMs into making wrong predictions ~\cite{niven2019probing,wallace2019trigger}. Current works mainly utilize the model prediction, prediction probabilities, and model gradients of the models to search adversarial examples. However, it is difficult to maintain the quality of the adversarial examples generated by machines. Recently, human-in-the-loop methods~\cite{wallace2019trick,nie2020adversarial} have been applied to generate more natural, valid, and diverse adversarial examples, which brings larger challenge and expose more properties and problems of PTMs. In conclusion, the robustness of PTMs has become a serious security threat when people deploy PTMs for real-world applications.

\subsection{Structural Sparsity of PTMs}

Following BERT, most PTMs adopt Transformer as the architecture backbone. Although people can easily train a deep Transformer and achieve significant improvement over previous works using CNN and RNN, Transformer meets the problem of over-parameterization. Researchers have shown that the multi-head attention structures are redundant in the tasks of machine translation~\cite{DBLP:conf/nips/MichelLN19}, abstractive summarization~\cite{DBLP:journals/corr/abs-1911-03898}, and language understanding~\cite{DBLP:conf/emnlp/KovalevaRRR19}, i.e., when removing part of attention heads, we can achieve better performance. This phenomenon is consistent to the observation in~\cite{DBLP:journals/corr/abs-1906-04341} where they find that most heads in the same layer have similar self-attention patterns. Furthermore, \citet{DBLP:conf/emnlp/KovalevaRRR19} conduct a qualitative and quantitative analysis of the information encoded by PTMs' heads. Their findings suggest that the attention behaviors of different heads can be categorized into a limited set of patterns. Besides the multi-head attention, several other works explore to identify the sparsity of parameters. \citet{DBLP:conf/rep4nlp/GordonDA20} show that low levels of pruning (30-40\%) do not affect pre-training loss or the performance on downstream tasks at all. Targeting the sparsity during fine-tuning, \citet{DBLP:conf/emnlp/PrasannaRR20} validate the lottery ticket hypothesis on PTMs and find that it is possible to find sub-networks achieving performance that is comparable with that of the full model. Surprisingly, \citet{kao2021berts} show that we can improvement the performance by simply duplicating some hidden layers to increase the model capacity, which suggests that the redundant parameters may benefit the fine-tuning.

\subsection{Theoretical Analysis of PTMs}

Since pre-training has achieved great success in deep learning, researchers try to investigate how pre-training works, especially unsupervised pre-training. In the early days of deep learning, people found that it is effective to train a deep belief network by greedy layer-wise unsupervised pre-training followed by supervised fine-tuning~\cite{DBLP:journals/neco/HintonOT06}. Recently, pre-training based on contrast learning including language modeling has become the mainstream approach. In this section, we will introduce some theoretical explanatory hypotheses or frameworks for pre-training.

\citet{DBLP:journals/jmlr/ErhanBCMVB10} propose two hypotheses to explain the effect of pre-training: (1) better optimization and (2) better regularization. In the aspect of better optimization, the network with pre-training is closer to the global minimum compared to the models randomly initialized. In the aspect of better regularization, the training error of PTMs is not necessarily better than the random models while the test error of PTMs is better, which means better generalization ability. Then, the experimental results lean towards the second hypothesis. They find that the PTM doesn't achieve lower training error. Moreover, compared to other regularization approaches such as L1/L2, the unsupervised pre-training regularization is much better.

Towards the recent development of pre-training objective, \citet{DBLP:conf/icml/SaunshiPAKK19} conduct a theoretical analysis of contrastive unsupervised representation learning. Contrastive learning treats the pairs of text/images appearing in the same context as the semantically similar pairs and the randomly sampled pairs as the semantically dissimilar pairs. Then, the distance between the similar pair should be close and the distance between the dissimilar pair should be distant. In the prediction process of language modeling, the context and the target word are the similar pair and the other words are negative samples~\cite{DBLP:conf/iclr/KongdYLDY20}. \citet{DBLP:conf/icml/SaunshiPAKK19} first provide a new conceptual framework to bridge the gap between pre-training and fine-tuning. Specifically, they introduce the concept of latent classes and the semantically similar pairs are from the same latent class. For example, the latent class can be ``'happy'' to include all texts including happy sentiments. The latent classes cover all possible classes and the classes defined by downstream tasks are from the set of latent classes. Then, they prove that the loss of contrastive learning is the upper bound of the downstream loss. Hence, when optimizing the pre-training loss, we can expect a lower loss in downstream tasks.

\section{Future Directions}
\label{sec:future}

So far, we have comprehensively reviewed the past and present of PTMs. In the future, on the basis of existing works, PTMs can be further developed from the following aspects: 
architectures and pre-training methods (section~\ref{subsec:arc}), 
multilingual and multimodal pre-Training (section~\ref{subsec:multi}), 
computational efficiency (section~\ref{subsec:comp}), 
theoretical foundation (section~\ref{subsec:theo}), 
modeledge learning (section~\ref{subsec:kb}), 
cognitive learning (section~\ref{subsec:cog}),
and novel applications (section~\ref{subsec:app}). 
In fact, researchers have made lots of efforts in the above directions, and we have also introduced the latest breakthroughs in the previous sections. However, there are still some open problems in these directions that need to be further addressed. We mainly focus on discussing these open problems in this section.

\subsection{Architectures and Pre-Training Methods}
\label{subsec:arc}

From the aspect of architectures and pre-training methods, we believe the following problems worth further exploring in the future:

\vpara{New Architectures.} Transformers have been proved to be an effective architecture for pre-training. However, the main limitation of Transformers is its computational complexity. Limited by the memory of GPUs, most current PTMs cannot deal with sequences containing more than 512 tokens. Therefore, it is important to search for more efficient model architectures to capture longer-range contextual information. However, the design of deep architecture is challenging, and we may seek help from some automatic methods, such as neural architecture search (NAS). Besides, although larger PTMs can usually lead to better performance, a practical problem is how to leverage these huge PTMs on some special scenarios, such as low-capacity devices and low-latency applications, where the efficiency of PTMs is a key factor. Moreover, different downstream tasks prefer different architectures. For example, the Transformer encoder is suitable for natural language understanding tasks while the Transformer decoder is suitable for natural language generation tasks. Therefore, we may need to carefully design task-specific architectures according to the type of downstream tasks.

\vpara{New Pre-Training Tasks.} The general-purpose PTMs are always our pursuits for learning the intrinsic universal knowledge of languages (even world knowledge). However, such PTMs usually need deeper architecture, larger corpora and challenging pre-training tasks. All these requirements further result in higher training costs. Moreover, training huge models is also a challenging problem, which needs sophisticated and efficient training techniques such as distributed training, mixed-precision training, etc. Therefore, a more practical direction is to design more efficient self-supervised pre-training tasks and training methods according to the capabilities of existing hardware and software. ELECTRA~\cite{clark2020electra} is a good attempt towards this direction.

\vpara{Beyond Fine-Tuning.} Currently, fine-tuning is the dominant method to transfer the knowledge of PTMs to downstream tasks but one deficiency is its parameter inefficiency: every downstream task has its own fine-tuned parameters. An improved solution is to fix the original parameters of PTMs and add small fine-tunable adaption modules for specific tasks. Thus, we can use a shared PTM to serve multiple downstream tasks. Recently, with the emerging of GPT-3, a novel genre for model tuning, namely prompt tuning, is getting more and more attention. By designing, generating and searching discrete~\cite{Petroni:EMNLP2019, gao2021making, hu2021knowledgeable} or continuous~\cite{liu2021gpt,han2021ptr,lester2021power} prompts and using MLM for specific downstream tasks, these models could (1) bridge the gap between pre-training and fine-tuning, and thereby perform better on downstream tasks; (2) reduce the computational cost on fine-tuning the tremendous amounts of parameters. To sum up, prompt tuning is a promising way to stimulate the linguistic and world knowledge distributed in PTMs.


\vpara{Reliability.} The reliability of PTMs is also becoming an issue of great concern with the extensive use of PTMs in production systems.
The studies of adversarial attacks~\cite{li2020bert,li2020generating,zhang2021red} against PTMs help us understand their capabilities by fully exposing their vulnerabilities. Adversarial defenses~\cite{si2020better,yao2019adversarial,li2021token} for PTMs are also promising, which can improve the robustness of PTMs and make them immune against adversarial attacks.
Overall, as a key component in many NLP applications, the interpretability and reliability of PTMs remain to be further explored, which will help us
understand how PTMs work and provide guidance for better use and further improvement of PTMs.

\subsection{Multilingual and Multimodal Pre-Training}
\label{subsec:multi}

Although multimodal and multilingual PTMs have witnessed numerous advances in the last two years, they still have the following ongoing research lines: 

\vpara{More Modalities.} In addition to image and text, video and audio can also be exploited for multimodal pre-training. The main challenge thus lies in how to model temporal contexts involved in these two modalities. In particular, for large-scale pre-training over video-text pairs, the conventional self-supervised learning methods are not suitable due to their high computational costs. To handle this problem, it is important to develop more effective and efficient self-supervised learning methods for more complex modalities.

\vpara{More Insightful Interpretation.} It is still unknown why bridging vision and language works. For example, regardless of the advantages brought by multimodal pre-training, does it lead to any harm to the single modality (image or text)? If the answer is yes, can we overcome this drawback during multimodal pre-training? Along this research line, the latest visualization tools for deep learning can be exploited for the interpretation of multimodal pre-training. 

\vpara{More Downstream Applications.}  It is well-known that multimodal pre-training can be applied to image-text retrieval, image-to-text generation, text-to-image generation and other downstream tasks. However, it is still challenging to find a ``true'' real-world application scenario for multimodal pre-training, since many effective engineering tricks can be leveraged instead (even with less cost). A closer collaboration with the industry is thus needed. 

\vpara{Transfer Learning.} Currently, to make multimodal multilingual models handle different languages, data for each language is required during pre-training. It is not flexible to add unseen languages during pre-training. Therefore, a new pre-training framework should be explored to easily adapt to those unseen languages. Besides, current multimodal multilingual models are not able to process audio data. For example, to translate English audio to Chinese audio, we need to first transfer English audio to English text by an extra speech recognition system. After translation with a cross-lingual model, we need to further transfer Chinese text to Chinese audio by an extra text-to-speech tool. How to directly transfer the source language audio to the target language text or target language audio by multimodal multilingual PTMs is also worth exploring.

\subsection{Computational Efficiency}
\label{subsec:comp}

Deep learning models have become increasingly complicated and large~\cite{devlin2019bert,brown2020language,kaplan-scaling-2020,fedus2021switch} in the recent years. The novel requirements of large-scale deep learning models bring severe challenges to the existing deep learning frameworks such as TensorFlow \cite{abadi-2016-tensorflow} and PyTorch \cite{paszke-2019-pytorch}, which were designed in the early days without initially foreseeing the emerging requirements such as model/pipeline parallelism of large models~\cite{brown2020language,huang2018gpipe,wang-2019-eurosys}. To develop more efficient frameworks, the following directions are helpful.

\vpara{Data Movement.} 
Developing an efficient distributed deep learning framework faces various challenges. 
One has to carefully manage the data movement between devices, which may otherwise become the performance bottleneck~\cite{narayanan-2019-sosp, jiang-osdi-2020}. A well-defined parallelism strategy is needed to place and schedule computational tasks on inter-connected devices, by minimizing the communication cost, maximizing the computational and memory resources, and optimizing the computation-communication overlap. In the best case, this efficient parallelism strategy can be generated automatically. 

\vpara{Parallelism Strategies.} Particular to the choice of parallelism strategy, data parallelism, model parallelism, pipeline parallelism, and various hybrid parallelism approaches can find their best usage depending on the structure of neural networks and hardware configuration~\cite{tal-ddl-2019}.
Data parallelism is especially suitable for deep learning models with a relatively small set of parameters (usually less than tens of million parameters) where near-linear speed-up can be achieved when the back-propagation maximally overlaps with the gradient/parameter communication~\cite{hashemi19tictac,peng-2019-sosp, jiang-osdi-2020}. Model parallelism and pipeline parallelism are for models with a more significant number of parameters, which probably cannot fit into a single device. In current practice, a user must thoroughly consider the network structure given a deep learning model and the inter-device communication bandwidth to decide the most appropriate parallelism strategies or switch between different strategies~\cite{shazeer-2018-nips}. 

\vpara{Large-Scale Training.} Given the poor support to model parallelism and pipeline parallelism by existing deep learning frameworks, some emerging open-source projects develop dedicated frameworks for large-scale training. For example, HugeCTR~\cite{Oldridge2020MerlinAG} is used for large-scale click-through rate estimation. Megatron-LM~\cite{shoeybi2019megatron,narayanan-megatron-2021} and DeepSpeed~\cite{rajbhandari-zeroinfinity-2021,rajbhandari2020zero} target at training large-scale NLP PTMs. InsightFace~\cite{insightface} trains large-scale face recognition models. However, these frameworks are restricted to limited application cases and cannot serve as a general solution. Further, these approaches cannot work together to constitute a complete solution due to the compatibility issue. 

\vpara{Wrappers and Plugins.} Without a mechanism to support model parallelism and pipeline parallelism, one has to develop various libraries dedicated to some particular algorithms via inserting the data routing operations by hand between computing operations on top of existing frameworks. Further, communication and computation need to be manually overlapped to maximize the system throughput. Manually programming communication operations is prohibitively complicated and only can solve problems case by case, leading to a significant obstacle in applying parallelism strategies to new deep learning models. If communication operations can be automatically managed transparently to users by deep learning frameworks, more models and applications can benefit from the distributed training.

To support more complicated parallelism strategies, many schemes are used as wrappers or plugins based on some mainstream deep learning frameworks such as TensorFlow and PyTorch. Mesh-TensorFlow \cite{shazeer-2018-nips}, FlexFlow \cite{jia-2019-sysml}, OneFlow~\cite{oneflow}, MindSpore~\cite{mindspore} and GShard \cite{lepikhin2020gshard} provide APIs for developers to express a wide range of parallel computation patterns for different components of deep neural models. The SBP configuration in OneFlow could be still too complex for users to set. However, directly programming with communication primitives for a different kind of parallelism is more complicated. OneFlow transforms the manually programming to just setting \textit{SBP} signatures. Moreover, in OneFlow, the user could just set the \textit{SBP} signatures of a subset of operations instead of the whole set, and leave the rest \textit{SBP} to be inferred with heuristic approaches like GShard~\cite{lepikhin2020gshard}, in which users provide some initial annotations or use default annotations as seed, then the algorithm propagates the sharding information to the un-annotated tensors. The approach in FlexFlow~\cite{jia-2019-sysml} can also be used here. The automatic scheduling of parallelism strategies is the trend of distributed training in the future.

\subsection{Theoretical Foundation}
\label{subsec:theo}

In this subsection, we analyze the future directions in a more fundamental way. In the aspect of theoretical foundation, we discuss the following research problems.

\vpara{Uncertainty.} One under-addressed issue with PTMs (as well as other deep neural networks) is that they are often over-confident in predictions, i.e., these models do not know what they do not know. For instance, GPT-3 can be used to answer questions with promising performance on benchmark datasets. However, if you ask a simple question like ``How many eyes does my foot have?'', GPT-3 would certainly produce an answer like ``Your foot has two eyes'', which looks counter-intuitive. \footnote{More examples of the Turing test of GPT-3 can be found at \url{https://lacker.io/ai/2020/07/06/giving-gpt-3-a-turing-test.html}} Of course, the above question is not often asked by normal human beings. It is generally a challenging task to deal with such out-of-distribution (OOD) data in machine learning. 

To address the above challenge, one promising direction is to adopt Bayesian methods that explore probabilistic tools to capture the uncertainty of both data and model (also known as aleatoric uncertainty and epistemic uncertainty respectively)~\cite{der2009aleatory} or derive some testing statistics. Such uncertainty or statistics is helpful to detect outliers~\cite{wang2020ood}. Recently, much work has been done on the theory, algorithms and programming libraries of Bayesian deep learning, which conjoins Bayesian methods and deep networks (e.g., see~\cite{zhusuan2017} for more details). Such progress can be further extended to large-scale PTMs to properly characterize uncertainty and avoid over-confident outputs. Of course, improving the computational efficiency of Bayesian deep learning is a key factor to address the above challenge. 

\vpara{Generalization and Robustness.} Another important issue with PTMs is on generalization. As an important advancement of deep learning, it inherits the advantages as well as challenges of deep neural networks. It has been observed that classical learning theory is not sufficient to understand the behavior of deep networks~\cite{zhang2017understanding}, thereby calling for new tools in learning theory. As for PTMs, besides theoretical understanding of the neural models themselves (e.g., Transformer and BERT), new questions arise. For example, it is important to theoretically understand the roles of pre-training in improving the generalization of downstream tasks. The recent work \cite{DBLP:conf/icml/SaunshiPAKK19} provides a fruitful attempt at understanding contrastive learning with particular assumptions. However, it is still largely open to analyze PTMs under more realistic settings.

As we mentioned before, the adversarial robustness also raises new questions. In previous work, it was shown that a higher sample complexity is needed in order to achieve adversarial robustness for neural networks~\cite{Schmidt2018samplecomplexity}. Such analysis has inspired further improvements~(e.g., \cite{pang2019rethinking}). However, it is generally unknown how large-scale PTMs can help in this aspect. Are there effective ways to explore PTMs as extra data resources to improve the robustness of downstream tasks? Also, the robustness of PTMs themselves is an unsolved issue, as mentioned before.

\subsection{Modeledge Learning}
\label{subsec:kb}

As introduced in section~\ref{sec:analysis}, PTMs can achieve a surge of improvements for a wide range of NLP tasks because they learn versatile knowledge from large unlabeled corpora. As opposed to the knowledge represented by discrete symbols, which is interpretable to human beings, the knowledge stored in PTMs is represented as real-valued vectors. For example, given a triple $\langle \mathit{h}, \mathit{r}, \mathit{t} \rangle$ in a knowledge graph, it is easy to know that the head entity $\mathit{h}$ has a relation $\mathit{r}$ to the tail entity $\mathit{t}$. In contrast, you seem to have difficulty knowing what a representation produced by a PTM means. Therefore, we can refer to the knowledge stored in PTMs as ``modeledge'', which is distinguished from the discrete symbolic knowledge formalized by human beings.

\vpara{Knowledge-Aware Tasks.} While the use of symbolic knowledge is effective, it is time-consuming and labor-intensive to manually organize this discrete knowledge such as building various knowledge bases. With the rapid advance of researches on PTMs, there emerge various PTMs such as GPT, BERT and BART. More and more researchers have probed into what knowledge do PTMs learn from the data, and why they perform so well on downstream tasks~\cite{Jawahar:ACL2019,Ethayarajh:EMNLP2019}. \citet{Petroni:EMNLP2019} state that PTMs can be seen as knowledge bases and study how to apply PTMs to the knowledge completion task. \citet{Ethayarajh:EMNLP2019} also claim that PTMs would be open knowledge graphs and propose an unsupervised method to build knowledge graphs based on PTMs. From all these knowledge-aware tasks, we can find that a wealth of human knowledge is captured by PTMs and stored in the form of modeledge. How to stimulate the modeledge of PTMs is worth further exploring in the future.

\vpara{Modeledge Storage and Management.} As existing PTMs are built on varying architectures and may be trained with different corpora, they contain diverse modeledge. As a result, how to store and manage various continuous modeledge in PTMs becomes a new challenge. There are two kinds of straightforward ideas. The first is to pre-train a huge model on extra-large scale data. Then, PTMs will have the extraordinary ability to cover almost all modeledge in existing PTMs. This method is simple and effective while it requires extremely high computational power and storage resources. For example, GPT-3 uses about 175 billion parameters. The second is to combine multiple models into one large model based on the mixture of experts (MoE)~\cite{Jacobs:1991}. For example, \citet{fedus2021switch} improve MoE to propose Switch Transformers. This method is easy to contain new models but the requirement of memory grows as the number of models increases.

Considering that there are both similarities and differences among existing PTMs, we have an important question that needs to be answered: is it possible to build a universal continuous knowledge base (UCKB) that stores modeledge from various PTMs? The UCKB can not only store continuous modeledge imported from existing PTMs but also can blend different modeledge and then export the fused modeledge to a model to make it more powerful. \citet{Chen:ArXiv2020} first propose the concept of UCKB and make some preliminary explorations. They regard neural networks as parameterized functions and use knowledge distillation~\cite{Hinton:NeurIPS2015} to import and export modeledge. UCKB overcomes the redundancy of model storage and stores the modeledge of various models into a common continuous knowledge base. However, how to design more effective architectures for the storage and interface of UCKB still remains challenging.

\subsection{Cognitive and Knowledgeable Learning}
\label{subsec:cog}

Making PTMs more knowledgeable is an important topic for the future of PTMs. We divide the future development of knowledgeable PTMs into the following three approaches:

\vpara{Knowledge Augmentation.} For an input text, there is rich related external knowledge, which can be used to augment the input. Considering the formats of knowledge and plain text are very different, it is important to bridge the gap between text representations and knowledge representations (including symbols or vectors) and use their information uniformly as input. The solution to this problem requires both unified model architectures and knowledge-guided pre-training objectives. 

\vpara{Knowledge Support.} Current model architectures are manually designed and usually very regular. With prior knowledge about the input, we can train different sub-module to process different kinds of input, which may accelerate the process of training and inference and benefit the model efficiency. This process is similar to human behavior where different brain regions correspond to different activity functions. 

\vpara{Knowledge Supervision.} Knowledge bases store amounts of structural data, which can be used as a complementary source during pre-training. By learning from both knowledge bases and large-scale corpora, PTMs can have better language understanding and generation abilities compared to only using plain text. Through these three directions, we hope the future PTMs can easily understand the meanings beyond words and achieve better performance on various downstream tasks.

In terms of cognitive PTMs, we believe the following approaches would be helpful:  

\vpara{Cognitive Architecture.} Since neural networks are inspired by the micro structure of the human neural system, it is expected to see how the macro function and organization of human cognitive system can enlighten the design of the next generation of intelligence system, such as the Global Working Theory (GWT). The success of CogQA and CogLTX may provide some thoughts on this challenge. 

\vpara{Explicit and Controllable Reasoning.} While deep learning has achieved success in many perceptive tasks, how to conduct complex decision making and efficient multi-step reasoning is still unsolved, which may require machines to automatically plan the decision making process into a cognitive graph and do explicit reasoning over the factors in graphs as human do. Methods such as InversePrompting~\cite{zou2021controllable} which shows supreme ability in controlling theme-related text generation would provide some thoughts. 

\vpara{Interactions of Knowledge.} Though our PTMs are getting bigger and more general, what knowledge it has learned from pre-training is largely unexplored. Moreover, since our brains are working with the collaboration of different function zones, it is important to see if our PTMs have shaped different inner function modules and how they would interact with each other.

\subsection{Applications}
\label{subsec:app}

PTMs have been successfully applied in a wide variety of domains and tasks. In this section, we will highlight some of these applications. 

\vpara{Natural Language Generation.} Many natural language generation tasks have been dominated by PTMs, such as GPT-2, BART, T5, UniLM and many more. These tasks include machine translation, summarization, dialog generation, story generation, poetry generation and other long text generation. Since the prevalent trend of PTMs, the backbone models have moved from CNNs/RNNs to transformers or transformer-based PTMs. 
PTMs have also been successfully applied to multimodal generation. Trained on text-image parallel data, these models have been shown strong in applications such as visual question answering, image-to-text generation and text-to-image generation. 
As large-scale PTMs have been trained on so large-scale data, they have innate advantages for natural language generation, particularly low-resourced natural language generation.

\vpara{Dialog Systems.} Many recent open-domain dialog systems are built upon large-scale transformer structures. These examples include Meena~\cite{adiwardana2020towards}, Blender~\cite{roller-etal-2021-recipes}, CDial-GPT~\cite{wang2020chinese}, Plato~\cite{bao-etal-2020-plato} and Plato-2~\cite{bao2021plato}, which are trained on large-scale conversation data, commonly with the seq2seq framework. These models have shown capabilities of delivering natural and engaging conversations, some of which have been reported to be close to human-level performance~\cite{adiwardana2020towards}. However, dialog-specific pre-training tasks are yet to be explored, comparing to pre-training tasks for other applications. 

\vpara{Domain-Specific PTMs.} When large-scale domain-specific corpora are cheaply available, we can train domain-specific PTMs on such data. Some notable works include BioBERT \cite{lee2020biobert}  and SciBERT \cite{beltagy2019scibert}, which are trained respectively on the biological and scientific literature text. These models are expected and verified to learn more domain-specific knowledge and language use than those trained on the general text. Such domain expertise is usually regarded as important for solving many domain-specific problems.

\vpara{Domain Adaptation and Task Adaptation.} Large-scale PTMs learn general knowledge from the large-scale general text, providing a good initial point to further learn domain-specific knowledge by fine-tuning or other techniques. Although PTMs are becoming larger and larger, the domain-specific data are always limited. Therefore, domain adaptation is becoming crucial for domain-specific applications. It has been evident that the simple fine-tuning of large-scale PTMs is not sufficient for domain-specific applications~\cite{gururangan-etal-2020-dont, ke2020sentilare}. The most essential reason for this is the distribution shift: the data distribution in a specific domain may be substantially different from that in the general pre-training text. 
Another important issue for the success of domain-specific applications goes to task adaptation. 
Most often, domain applications have a small set of labeled data, which can empower supervised learning to learn domain expertise more efficiently. However, for super-large PTMs, simply fine-tuning on labeled data seems to be inefficient in computation, nor effective in performance. Thus, how to bridge the gap between pre-training and task-specific fine-tuning becomes crucial. Moreover, efficient and effective task-specific fine-tuning is also an important research direction for the future application of PTMs~\cite{soares2019matching, ding2021prototypical}.

\section{Conclusion}

In this paper, we take a look into the history of pre-training to indicate the core issue of PTMs, and meanwhile reveal the crucial position of PTMs in the AI development spectrum. Furthermore, we comprehensively review the latest efforts towards better PTMs, including designing effective architectures, utilizing rich contexts, improving computational efficiency, and conducting interpretation and theoretical analysis. All these works contribute to the recent wave of developing PTMs. Although existing PTMs have achieved promising results, especially those large-scale PTMs showing amazing abilities in zero/few-shot learning scenarios, how to develop PTMs next is still an open question. The knowledge stored in PTMs is represented as real-valued vectors, which is quite different from the discrete symbolic knowledge formalized by human beings. We name this continuous and machine-friendly knowledge ``modeledge'' and believe that it is promising to capture the modeledge in a more effective and efficient way and stimulate the modeledge for specific tasks. We hope our view could inspire more efforts in this field and advance the development of PTMs.

\section*{Note and Contribution}

This paper originates from a 3-day closed-door workshop initiated by Jie Tang, Ji-Rong Wen and Minlie Huang held in Beijing WTown from January 1 to January 3, 2021, supported by China Computer Federation (CCF). All authors of this paper organized or participated in this workshop, and this paper can be regarded as a summary and extension of the discussion in the workshop.

The contributions of all authors are listed as follows: Zhiyuan Liu and Xu Han designed the structure of this paper; Xu Han drafted the abstract, Section~\ref{sec:introduction}, Section~\ref{sec:background}; Ning Ding and Xu Han drafted Section~\ref{sec:plm}; Xiao Liu and Jiezhong Qiu drafted Section~\ref{sec:arch}; Yuqi Huo, Yuan Yao, Ao Zhang and Liang Zhang drafted Section~\ref{sec:rich-data}; Yuxian Gu drafted Section~\ref{sec:efficiency}; Zhengyan Zhang drafted Section~\ref{sec:analysis}. All faculty authors drafted various topics in Section~\ref{sec:future}, including Xipeng Qiu for Section~\ref{subsec:arc}, Ji-Rong Wen, Ruihua Song and Yang Liu for Section~\ref{subsec:multi}, Jinhui Yuan and Wentao Han for Section~\ref{subsec:comp}, Jun Zhu and Yanyan Lan for Section~\ref{subsec:theo}, Yang Liu for Section~\ref{subsec:kb}, Jie Tang and Zhiyuan Liu for Section~\ref{subsec:cog}, Minlie Huang and Jie Tang for Section~\ref{subsec:app}. Wayne Xin Zhao, Xipeng Qiu provided comments to the manuscript, and Xu Han, Ning Ding and Zhengyan Zhang proofread the whole paper.

\bibliographystyle{acl_natbib}
\bibliography{arr}

\begin{thebibliography}{322}
\expandafter\ifx\csname natexlab\endcsname\relax\def\natexlab#1{#1}\fi

\bibitem[{ins(2021)}]{insightface}
 2021.
\newblock Insightface project.
\newblock \url{https://github.com/deepinsight/insightface}.

\bibitem[{min(2021)}]{mindspore}
 2021.
\newblock {MindSpore Deep Learning Framework}.
\newblock \url{https://github.com/mindspore-ai/mindspore}.

\bibitem[{one(2021)}]{oneflow}
 2021.
\newblock {OneFlow Deep Learning Framework}.
\newblock \url{https://github.com/Oneflow-Inc/oneflow}.

\bibitem[{Abadi et~al.(2016)Abadi, Barham, Chen, Chen, Davis, Dean, Devin,
  Ghemawat, Irving, Isard, Kudlur, Levenberg, Monga, Moore, Murray, Steiner,
  Tucker, Vasudevan, Warden, Wicke, Yu, and Zheng}]{abadi-2016-tensorflow}
Mart\'{\i}n Abadi, Paul Barham, Jianmin Chen, Zhifeng Chen, Andy Davis, Jeffrey
  Dean, Matthieu Devin, Sanjay Ghemawat, Geoffrey Irving, Michael Isard,
  Manjunath Kudlur, Josh Levenberg, Rajat Monga, Sherry Moore, Derek~G. Murray,
  Benoit Steiner, Paul Tucker, Vijay Vasudevan, Pete Warden, Martin Wicke, Yuan
  Yu, and Xiaoqiang Zheng. 2016.
\newblock Tensorflow: A system for large-scale machine learning.
\newblock In \emph{Proceedings of OSDI}, pages 265--283.

\bibitem[{Adi et~al.(2017)Adi, Kermany, Belinkov, Lavi, and
  Goldberg}]{DBLP:conf/iclr/AdiKBLG17}
Yossi Adi, Einat Kermany, Yonatan Belinkov, Ofer Lavi, and Yoav Goldberg. 2017.
\newblock Fine-grained analysis of sentence embeddings using auxiliary
  prediction tasks.
\newblock In \emph{Proceedings of ICLR}.

\bibitem[{Adiwardana et~al.(2020)Adiwardana, Luong, So, Hall, Fiedel,
  Thoppilan, Yang, Kulshreshtha, Nemade, Lu et~al.}]{adiwardana2020towards}
Daniel Adiwardana, Minh-Thang Luong, David~R So, Jamie Hall, Noah Fiedel, Romal
  Thoppilan, Zi~Yang, Apoorv Kulshreshtha, Gaurav Nemade, Yifeng Lu, et~al.
  2020.
\newblock Towards a human-like open-domain chatbot.
\newblock \emph{arXiv preprint arXiv:2001.09977}.

\bibitem[{Ainslie et~al.(2020)Ainslie, Ontanon, Alberti, Pham, Ravula, and
  Sanghai}]{ainslie2020etc}
Joshua Ainslie, Santiago Ontanon, Chris Alberti, Philip Pham, Anirudh Ravula,
  and Sumit Sanghai. 2020.
\newblock {ETC}: Encoding long and structured inputs in transformers.
\newblock In \emph{Proceedings of EMNLP}, pages 268--284.

\bibitem[{Alberti et~al.(2019)Alberti, Ling, Collins, and
  Reitter}]{alberti2019fusion}
Chris Alberti, Jeffrey Ling, Michael Collins, and David Reitter. 2019.
\newblock Fusion of detected objects in text for visual question answering.
\newblock In \emph{Proceedings of EMNLP-IJCNLP}, pages 2131--2140.

\bibitem[{Antol et~al.(2015)Antol, Agrawal, Lu, Mitchell, Batra, Zitnick, and
  Parikh}]{antol2015vqa}
Stanislaw Antol, Aishwarya Agrawal, Jiasen Lu, Margaret Mitchell, Dhruv Batra,
  C~Lawrence Zitnick, and Devi Parikh. 2015.
\newblock Vqa: Visual question answering.
\newblock In \emph{Proceedings of ICCV}, pages 2425--2433.

\bibitem[{Arjovsky et~al.(2017)Arjovsky, Chintala, and
  Bottou}]{arjovsky2017wasserstein}
Martin Arjovsky, Soumith Chintala, and L{\'e}on Bottou. 2017.
\newblock Wasserstein generative adversarial networks.
\newblock In \emph{Proceedings of ICML}, pages 214--223.

\bibitem[{Ba et~al.(2016)Ba, Kiros, and Hinton}]{ba2016layer}
Jimmy~Lei Ba, Jamie~Ryan Kiros, and Geoffrey~E Hinton. 2016.
\newblock Layer normalization.
\newblock In \emph{Proceedings of NeurIPS}.

\bibitem[{Baan et~al.(2019)Baan, ter Hoeve, van~der Wees, Schuth, and
  de~Rijke}]{DBLP:journals/corr/abs-1911-03898}
Joris Baan, Maartje ter Hoeve, Marlies van~der Wees, Anne Schuth, and Maarten
  de~Rijke. 2019.
\newblock Understanding multi-head attention in abstractive summarization.
\newblock \emph{arXiv preprint arXiv:1911.03898}.

\bibitem[{Baddeley(1992)}]{baddeley1992working}
Alan Baddeley. 1992.
\newblock Working memory.
\newblock \emph{Science}, 255(5044):556--559.

\bibitem[{Bao et~al.(2020)Bao, He, Wang, Wu, and Wang}]{bao-etal-2020-plato}
Siqi Bao, Huang He, Fan Wang, Hua Wu, and Haifeng Wang. 2020.
\newblock {PLATO}: Pre-trained dialogue generation model with discrete latent
  variable.
\newblock In \emph{Proceedings of ACL}.

\bibitem[{Bao et~al.(2021)Bao, He, Wang, Wu, Wang, Wu, Guo, Liu, and
  Xu}]{bao2021plato}
Siqi Bao, Huang He, Fan Wang, Hua Wu, Haifeng Wang, Wenquan Wu, Zhen Guo,
  Zhibin Liu, and Xinchao Xu. 2021.
\newblock Plato-2: Towards building an open-domain chatbot via curriculum
  learning.
\newblock In \emph{Proceedings of ACL}.

\bibitem[{Barrouillet et~al.(2004)Barrouillet, Bernardin, and
  Camos}]{barrouillet2004time}
Pierre Barrouillet, Sophie Bernardin, and Val{\'e}rie Camos. 2004.
\newblock Time constraints and resource sharing in adults' working memory
  spans.
\newblock \emph{Journal of Experimental Psychology: General}, 133(1):83--100.

\bibitem[{Belkin et~al.(2019)Belkin, Hsu, Ma, and
  Mandal}]{belkin2019reconciling}
Mikhail Belkin, Daniel Hsu, Siyuan Ma, and Soumik Mandal. 2019.
\newblock Reconciling modern machine-learning practice and the classical
  bias--variance trade-off.
\newblock \emph{PNAS}, 116(32):15849--15854.

\bibitem[{Beltagy et~al.(2019)Beltagy, Lo, and Cohan}]{beltagy2019scibert}
Iz~Beltagy, Kyle Lo, and Arman Cohan. 2019.
\newblock Scibert: A pretrained language model for scientific text.
\newblock In \emph{Proceedings of EMNLP-IJCNLP}, pages 3615--3620.

\bibitem[{Beltagy et~al.(2020)Beltagy, Peters, and
  Cohan}]{beltagy2020longformer}
Iz~Beltagy, Matthew~E Peters, and Arman Cohan. 2020.
\newblock Longformer: The long-document transformer.
\newblock \emph{arXiv preprint arXiv:2004.05150}.

\bibitem[{Ben-Nun and Hoefler(2019)}]{tal-ddl-2019}
Tal Ben-Nun and Torsten Hoefler. 2019.
\newblock Demystifying parallel and distributed deep learning: An in-depth
  concurrency analysis.
\newblock \emph{ACM Computing Surveys (CSUR)}, 52(4):1--43.

\bibitem[{Bengio et~al.(2003)Bengio, Ducharme, Vincent, and
  Janvin}]{bengio2003neural}
Yoshua Bengio, R{\'e}jean Ducharme, Pascal Vincent, and Christian Janvin. 2003.
\newblock A neural probabilistic language model.
\newblock \emph{JMLR}, 3:1137--1155.

\bibitem[{Bengio et~al.(1994)Bengio, Simard, and Frasconi}]{bengio1994learning}
Yoshua Bengio, Patrice Simard, and Paolo Frasconi. 1994.
\newblock Learning long-term dependencies with gradient descent is difficult.
\newblock \emph{IEEE TNNLS}, 5(2):157--166.

\bibitem[{Bi et~al.(2020)Bi, Li, Wu, Yan, and Wang}]{bi2020palm}
Bin Bi, Chenliang Li, Chen Wu, Ming Yan, and Wei Wang. 2020.
\newblock Palm: Pre-training an autoencoding\&autoregressive language model for
  context-conditioned generation.
\newblock In \emph{Proceedings of EMNLP}, pages 8681--8691.

\bibitem[{Bojar et~al.(2014)Bojar, Buck, Federmann, Haddow, Koehn, Leveling,
  Monz, Pecina, Post, Saint-Amand et~al.}]{bojar2014findings}
Ond{\v{r}}ej Bojar, Christian Buck, Christian Federmann, Barry Haddow, Philipp
  Koehn, Johannes Leveling, Christof Monz, Pavel Pecina, Matt Post, Herve
  Saint-Amand, et~al. 2014.
\newblock Findings of the 2014 workshop on statistical machine translation.
\newblock In \emph{Proceedings of WMT}, pages 12--58.

\bibitem[{Bosselut et~al.(2019)Bosselut, Rashkin, Sap, Malaviya, Celikyilmaz,
  and Choi}]{bosselut2019comet}
Antoine Bosselut, Hannah Rashkin, Maarten Sap, Chaitanya Malaviya, Asli
  Celikyilmaz, and Yejin Choi. 2019.
\newblock Comet: Commonsense transformers for automatic knowledge graph
  construction.
\newblock In \emph{Proceedings of ACL}, pages 4762--4779.

\bibitem[{Bouraoui et~al.(2020)Bouraoui, Camacho{-}Collados, and
  Schockaert}]{DBLP:conf/aaai/BouraouiCS20}
Zied Bouraoui, Jos{\'{e}} Camacho{-}Collados, and Steven Schockaert. 2020.
\newblock Inducing relational knowledge from {BERT}.
\newblock In \emph{Proceedings of AAAI}, pages 7456--7463.

\bibitem[{Brown(1958)}]{brown1958some}
John Brown. 1958.
\newblock Some tests of the decay theory of immediate memory.
\newblock \emph{Quarterly journal of experimental psychology}, 10(1):12--21.

\bibitem[{Brown et~al.(2020)Brown, Mann, Ryder, Subbiah, Kaplan, Dhariwal,
  Neelakantan, Shyam, Sastry, Askell, Agarwal, Herbert-Voss, Krueger, Henighan,
  Child, Ramesh, Ziegler, Wu, Winter, Hesse, Chen, Sigler, Litwin, Gray, Chess,
  Clark, Berner, McCandlish, Radford, Sutskever, and
  Amodei}]{brown2020language}
Tom Brown, Benjamin Mann, Nick Ryder, Melanie Subbiah, Jared~D Kaplan, Prafulla
  Dhariwal, Arvind Neelakantan, Pranav Shyam, Girish Sastry, Amanda Askell,
  Sandhini Agarwal, Ariel Herbert-Voss, Gretchen Krueger, Tom Henighan, Rewon
  Child, Aditya Ramesh, Daniel Ziegler, Jeffrey Wu, Clemens Winter, Chris
  Hesse, Mark Chen, Eric Sigler, Mateusz Litwin, Scott Gray, Benjamin Chess,
  Jack Clark, Christopher Berner, Sam McCandlish, Alec Radford, Ilya Sutskever,
  and Dario Amodei. 2020.
\newblock Language models are few-shot learners.
\newblock In \emph{Proceedings of NeurIPS}, pages 1877--1901.

\bibitem[{Carion et~al.(2020)Carion, Massa, Synnaeve, Usunier, Kirillov, and
  Zagoruyko}]{carion2020end}
Nicolas Carion, Francisco Massa, Gabriel Synnaeve, Nicolas Usunier, Alexander
  Kirillov, and Sergey Zagoruyko. 2020.
\newblock End-to-end object detection with transformers.
\newblock In \emph{Proceedings of ECCV}, pages 213--229.

\bibitem[{Chen et~al.(2020{\natexlab{a}})Chen, Sun, and Liu}]{Chen:ArXiv2020}
Gang Chen, Maosong Sun, and Yang Liu. 2020{\natexlab{a}}.
\newblock Towards a universal continuous knowledge base.
\newblock \emph{arXiv preprint arXiv:2012.13568}.

\bibitem[{Chen et~al.(2020{\natexlab{b}})Chen, Zhang, He, Ke, Wang, and
  Liu}]{chen2020variance}
Liang Chen, Tianyuan Zhang, Di~He, Guolin Ke, Liwei Wang, and Tie-Yan Liu.
  2020{\natexlab{b}}.
\newblock Variance-reduced language pretraining via a mask proposal network.
\newblock \emph{arXiv preprint arXiv:2008.05333}.

\bibitem[{Chen et~al.(2020{\natexlab{c}})Chen, Gan, Cheng, Li, Carin, and
  Liu}]{chen2020got}
Liqun Chen, Zhe Gan, Yu~Cheng, Linjie Li, Lawrence Carin, and Jingjing Liu.
  2020{\natexlab{c}}.
\newblock Graph optimal transport for cross-domain alignment.
\newblock In \emph{Proceedings of ICML}, pages 1542--1553. PMLR.

\bibitem[{Chen et~al.(2020{\natexlab{d}})Chen, Kornblith, Norouzi, and
  Hinton}]{chen2020simple}
Ting Chen, Simon Kornblith, Mohammad Norouzi, and Geoffrey Hinton.
  2020{\natexlab{d}}.
\newblock A simple framework for contrastive learning of visual
  representations.
\newblock In \emph{Proceedings of ICML}, pages 1597--1607.

\bibitem[{Chen et~al.(2020{\natexlab{e}})Chen, Su, Yan, and
  Wang}]{chen2020kgpt}
Wenhu Chen, Yu~Su, Xifeng Yan, and William~Yang Wang. 2020{\natexlab{e}}.
\newblock Kgpt: Knowledge-grounded pre-training for data-to-text generation.
\newblock In \emph{Proceedings of EMNLP}, pages 8635--8648.

\bibitem[{Chen et~al.(2015)Chen, Fang, Lin, Vedantam, Gupta, Doll{\'a}r, and
  Zitnick}]{chen2015microsoft}
Xinlei Chen, Hao Fang, Tsung-Yi Lin, Ramakrishna Vedantam, Saurabh Gupta, Piotr
  Doll{\'a}r, and C~Lawrence Zitnick. 2015.
\newblock Microsoft coco captions: Data collection and evaluation server.
\newblock \emph{arXiv preprint arXiv:1504.00325}.

\bibitem[{Chen and He(2020)}]{chen2020exploring}
Xinlei Chen and Kaiming He. 2020.
\newblock Exploring simple siamese representation learning.
\newblock \emph{arXiv preprint arXiv:2011.10566}.

\bibitem[{Chen et~al.(2020{\natexlab{f}})Chen, Li, Yu, El~Kholy, Ahmed, Gan,
  Cheng, and Liu}]{chen2020uniter}
Yen-Chun Chen, Linjie Li, Licheng Yu, Ahmed El~Kholy, Faisal Ahmed, Zhe Gan,
  Yu~Cheng, and Jingjing Liu. 2020{\natexlab{f}}.
\newblock Uniter: Universal image-text representation learning.
\newblock In \emph{Proceedings of ECCV}, pages 104--120.

\bibitem[{Chi et~al.(2020{\natexlab{a}})Chi, Dong, Wei, Wang, Mao, and
  Huang}]{xnlg}
Zewen Chi, Li~Dong, Furu Wei, Wenhui Wang, Xian-Ling Mao, and Heyan Huang.
  2020{\natexlab{a}}.
\newblock Cross-lingual natural language generation via pre-training.
\newblock In \emph{Proceedings of AAAI}, pages 7570--7577.

\bibitem[{Chi et~al.(2020{\natexlab{b}})Chi, Dong, Wei, Yang, Singhal, Wang,
  Song, Mao, Huang, and Zhou}]{chi2020infoxlm}
Zewen Chi, Li~Dong, Furu Wei, Nan Yang, Saksham Singhal, Wenhui Wang, Xia Song,
  Xian-Ling Mao, Heyan Huang, and Ming Zhou. 2020{\natexlab{b}}.
\newblock Infoxlm: An information-theoretic framework for cross-lingual
  language model pre-training.
\newblock \emph{arXiv preprint arXiv:2007.07834}.

\bibitem[{Child et~al.(2019)Child, Gray, Radford, and
  Sutskever}]{child2019generating}
Rewon Child, Scott Gray, Alec Radford, and Ilya Sutskever. 2019.
\newblock Generating long sequences with sparse transformers.
\newblock \emph{arXiv preprint arXiv:1904.10509}.

\bibitem[{Choromanski et~al.(2021)Choromanski, Likhosherstov, Dohan, Song,
  Gane, Sarlos, Hawkins, Davis, Mohiuddin, Kaiser
  et~al.}]{choromanski2020rethinking}
Krzysztof Choromanski, Valerii Likhosherstov, David Dohan, Xingyou Song,
  Andreea Gane, Tamas Sarlos, Peter Hawkins, Jared Davis, Afroz Mohiuddin,
  Lukasz Kaiser, et~al. 2021.
\newblock Rethinking attention with performers.
\newblock In \emph{Proceedings of ICLR}.

\bibitem[{Chuang et~al.(2019)Chuang, Liu, Lee, and Lee}]{chuang2019speechbert}
Yung-Sung Chuang, Chi-Liang Liu, Hung-Yi Lee, and Lin-shan Lee. 2019.
\newblock Speechbert: An audio-and-text jointly learned language model for
  end-to-end spoken question answering.
\newblock \emph{arXiv preprint arXiv:1910.11559}.

\bibitem[{Clark et~al.(2019)Clark, Khandelwal, Levy, and
  Manning}]{DBLP:journals/corr/abs-1906-04341}
Kevin Clark, Urvashi Khandelwal, Omer Levy, and Christopher~D Manning. 2019.
\newblock What does bert look at? an analysis of bert’s attention.
\newblock In \emph{Proceedings of BlackboxNLP}, pages 276--286.

\bibitem[{Clark et~al.(2020)Clark, Luong, Le, and Manning}]{clark2020electra}
Kevin Clark, Minh-Thang Luong, Quoc~V Le, and Christopher~D Manning. 2020.
\newblock Electra: Pre-training text encoders as discriminators rather than
  generators.
\newblock In \emph{Proceedings of ICLR}.

\bibitem[{Collobert and Weston(2008)}]{collobert2008unified}
Ronan Collobert and Jason Weston. 2008.
\newblock A unified architecture for natural language processing: Deep neural
  networks with multitask learning.
\newblock In \emph{Proceedings of ICML}, pages 160--167.

\bibitem[{Conneau et~al.(2020)Conneau, Khandelwal, Goyal, Chaudhary, Wenzek,
  Guzm{\'a}n, Grave, Ott, Zettlemoyer, and Stoyanov}]{conneau2020unsupervised}
Alexis Conneau, Kartikay Khandelwal, Naman Goyal, Vishrav Chaudhary, Guillaume
  Wenzek, Francisco Guzm{\'a}n, {\'E}douard Grave, Myle Ott, Luke Zettlemoyer,
  and Veselin Stoyanov. 2020.
\newblock Unsupervised cross-lingual representation learning at scale.
\newblock In \emph{Proceedings of ACL}, pages 8440--8451.

\bibitem[{Conneau et~al.(2018{\natexlab{a}})Conneau, Kruszewski, Lample,
  Barrault, and Baroni}]{DBLP:conf/acl/BaroniBLKC18}
Alexis Conneau, Germ{\'{a}}n Kruszewski, Guillaume Lample, Lo{\"{\i}}c
  Barrault, and Marco Baroni. 2018{\natexlab{a}}.
\newblock What you can cram into a single
  {\textbackslash}{\textdollar}{\&}!{\#}* vector: Probing sentence embeddings
  for linguistic properties.
\newblock In \emph{Proceedings of ACL}, pages 2126--2136.

\bibitem[{Conneau et~al.(2018{\natexlab{b}})Conneau, Rinott, Lample, Williams,
  Bowman, Schwenk, and Stoyanov}]{conneau2018xnli}
Alexis Conneau, Ruty Rinott, Guillaume Lample, Adina Williams, Samuel Bowman,
  Holger Schwenk, and Veselin Stoyanov. 2018{\natexlab{b}}.
\newblock Xnli: Evaluating cross-lingual sentence representations.
\newblock In \emph{Proceedings of EMNLP}, pages 2475--2485.

\bibitem[{Cordts et~al.(2016)Cordts, Omran, Ramos, Rehfeld, Enzweiler,
  Benenson, Franke, Roth, and Schiele}]{cordts2016cityscapes}
Marius Cordts, Mohamed Omran, Sebastian Ramos, Timo Rehfeld, Markus Enzweiler,
  Rodrigo Benenson, Uwe Franke, Stefan Roth, and Bernt Schiele. 2016.
\newblock The cityscapes dataset for semantic urban scene understanding.
\newblock In \emph{Proceedings of CVPR}, pages 3213--3223.

\bibitem[{Cui et~al.(2019)Cui, Che, Liu, Qin, Yang, Wang, and Hu}]{cui2019pre}
Yiming Cui, Wanxiang Che, Ting Liu, Bing Qin, Ziqing Yang, Shijin Wang, and
  Guoping Hu. 2019.
\newblock Pre-training with whole word masking for chinese bert.
\newblock \emph{arXiv preprint arXiv:1906.08101}.

\bibitem[{Da and Kasai(2019)}]{da-kasai-2019-cracking}
Jeff Da and Jungo Kasai. 2019.
\newblock Cracking the contextual commonsense code: Understanding commonsense
  reasoning aptitude of deep contextual representations.
\newblock In \emph{Proceedings of EMNLP Workshop}.

\bibitem[{Dai et~al.(2007)Dai, Xue, Yang, and Yu}]{dai2007co}
Wenyuan Dai, Gui-Rong Xue, Qiang Yang, and Yong Yu. 2007.
\newblock Co-clustering based classification for out-of-domain documents.
\newblock In \emph{Proceedings of KDD}, pages 210--219.

\bibitem[{Dai et~al.(2008)Dai, Yang, Xue, and Yu}]{dai2008self}
Wenyuan Dai, Qiang Yang, Gui-Rong Xue, and Yong Yu. 2008.
\newblock Self-taught clustering.
\newblock In \emph{Proceedings of ICML}, pages 200--207.

\bibitem[{Dai et~al.(2019)Dai, Yang, Yang, Carbonell, Le, and
  Salakhutdinov}]{dai2019transformer}
Zihang Dai, Zhilin Yang, Yiming Yang, Jaime Carbonell, Quoc~V Le, and Ruslan
  Salakhutdinov. 2019.
\newblock Transformer-xl: Attentive language models beyond a fixed-length
  context.
\newblock In \emph{Proceedings of ACL}, pages 2978--2988.

\bibitem[{Daume~III and Marcu(2006)}]{daume2006domain}
Hal Daume~III and Daniel Marcu. 2006.
\newblock Domain adaptation for statistical classifiers.
\newblock \emph{JAIR}, 26:101--126.

\bibitem[{Davison et~al.(2019)Davison, Feldman, and
  er~M.~Rush}]{DBLP:conf/emnlp/DavisonFR19}
Joe Davison, Joshua Feldman, and Alexand er~M.~Rush. 2019.
\newblock Commonsense knowledge mining from pretrained models.
\newblock In \emph{Proceedings of EMNLP-IJCNLP}, pages 1173--1178.

\bibitem[{Deng et~al.(2009)Deng, Dong, Socher, Li, Li, and
  Fei-Fei}]{deng2009imagenet}
Jia Deng, Wei Dong, Richard Socher, Li-Jia Li, Kai Li, and Li~Fei-Fei. 2009.
\newblock Imagenet: A large-scale hierarchical image database.
\newblock In \emph{Proceedings of CVPR}, pages 248--255.

\bibitem[{Der~Kiureghian and Ditlevsen(2009)}]{der2009aleatory}
Armen Der~Kiureghian and Ove Ditlevsen. 2009.
\newblock Aleatory or epistemic? does it matter?
\newblock \emph{Structural safety}, 31(2):105--112.

\bibitem[{Devlin et~al.(2019)Devlin, Chang, Lee, and
  Toutanova}]{devlin2019bert}
Jacob Devlin, Ming-Wei Chang, Kenton Lee, and Kristina Toutanova. 2019.
\newblock {BERT}: Pre-training of deep bidirectional transformers for language
  understanding.
\newblock In \emph{Proceedings of NAACL-HLT}, pages 4171--4186.

\bibitem[{Dhingra et~al.(2020)Dhingra, Zaheer, Balachandran, Neubig,
  Salakhutdinov, and Cohen}]{dhingra2020differentiable}
Bhuwan Dhingra, Manzil Zaheer, Vidhisha Balachandran, Graham Neubig, Ruslan
  Salakhutdinov, and William~W Cohen. 2020.
\newblock Differentiable reasoning over a virtual knowledge base.
\newblock In \emph{Proceedings of ICLR}.

\bibitem[{Ding et~al.(2021{\natexlab{a}})Ding, Yang, Hong, Zheng, Zhou, Yin,
  Lin, Zou, Shao, Yang et~al.}]{ding2021cogview}
Ming Ding, Zhuoyi Yang, Wenyi Hong, Wendi Zheng, Chang Zhou, Da~Yin, Junyang
  Lin, Xu~Zou, Zhou Shao, Hongxia Yang, et~al. 2021{\natexlab{a}}.
\newblock Cogview: Mastering text-to-image generation via transformers.
\newblock \emph{arXiv preprint arXiv:2105.13290}.

\bibitem[{Ding et~al.(2019)Ding, Zhou, Chen, Yang, and
  Tang}]{ding2019cognitive}
Ming Ding, Chang Zhou, Qibin Chen, Hongxia Yang, and Jie Tang. 2019.
\newblock Cognitive graph for multi-hop reading comprehension at scale.
\newblock In \emph{Proceedings of ACL}, pages 2694--2703.

\bibitem[{Ding et~al.(2020)Ding, Zhou, Yang, and Tang}]{ding2020cogltx}
Ming Ding, Chang Zhou, Hongxia Yang, and Jie Tang. 2020.
\newblock Cogltx: Applying bert to long texts.
\newblock In \emph{Proceedings of NeurIPS}, volume~33, pages 12792--12804.

\bibitem[{Ding et~al.(2021{\natexlab{b}})Ding, Wang, Fu, Xu, Wang, Xie, Shen,
  Huang, Zheng, and Zhang}]{ding2021prototypical}
Ning Ding, Xiaobin Wang, Yao Fu, Guangwei Xu, Rui Wang, Pengjun Xie, Ying Shen,
  Fei Huang, Hai-Tao Zheng, and Rui Zhang. 2021{\natexlab{b}}.
\newblock Prototypical representation learning for relation extraction.
\newblock In \emph{Proceedings of ICLR}.

\bibitem[{Donahue et~al.(2015)Donahue, Anne~Hendricks, Guadarrama, Rohrbach,
  Venugopalan, Saenko, and Darrell}]{donahue2015long}
Jeffrey Donahue, Lisa Anne~Hendricks, Sergio Guadarrama, Marcus Rohrbach,
  Subhashini Venugopalan, Kate Saenko, and Trevor Darrell. 2015.
\newblock Long-term recurrent convolutional networks for visual recognition and
  description.
\newblock In \emph{Proceedings of CVPR}, pages 2625--2634.

\bibitem[{Dong et~al.(2019)Dong, Yang, Wang, Wei, Liu, Wang, Gao, Zhou, and
  Hon}]{dong2019unified}
Li~Dong, Nan Yang, Wenhui Wang, Furu Wei, Xiaodong Liu, Yu~Wang, Jianfeng Gao,
  Ming Zhou, and Hsiao-Wuen Hon. 2019.
\newblock Unified language model pre-training for natural language
  understanding and generation.
\newblock In \emph{Proceedings of NeurIPS}.

\bibitem[{Du et~al.(2021)Du, Qian, Liu, Ding, Qiu, Yang, and Tang}]{du2021all}
Zhengxiao Du, Yujie Qian, Xiao Liu, Ming Ding, Jiezhong Qiu, Zhilin Yang, and
  Jie Tang. 2021.
\newblock All nlp tasks are generation tasks: A general pretraining framework.
\newblock \emph{arXiv preprint arXiv:2103.10360}.

\bibitem[{Erhan et~al.(2010)Erhan, Courville, Bengio, and
  Vincent}]{DBLP:journals/jmlr/ErhanBCMVB10}
Dumitru Erhan, Aaron Courville, Yoshua Bengio, and Pascal Vincent. 2010.
\newblock Why does unsupervised pre-training help deep learning?
\newblock In \emph{Proceedings of AISTATS}, pages 201--208.

\bibitem[{Ethayarajh(2019)}]{Ethayarajh:EMNLP2019}
Kawin Ethayarajh. 2019.
\newblock How contextual are contextualized word representations? comparing the
  geometry of bert, elmo, and gpt-2 embeddings.
\newblock In \emph{Proceedings of EMNLP-IJCNLP}, pages 55--65.

\bibitem[{Ettinger(2020)}]{DBLP:journals/tacl/Ettinger20}
Allyson Ettinger. 2020.
\newblock What {BERT} is not: Lessons from a new suite of psycholinguistic
  diagnostics for language models.
\newblock \emph{TACL}, 8:34--48.

\bibitem[{Ettinger et~al.(2016)Ettinger, Elgohary, and
  Resnik}]{DBLP:conf/repeval/EttingerER16}
Allyson Ettinger, Ahmed Elgohary, and Philip Resnik. 2016.
\newblock Probing for semantic evidence of composition by means of simple
  classification tasks.
\newblock In \emph{Proceedings of RepEval}, pages 134--139.

\bibitem[{Evgeniou and Pontil(2007)}]{evgeniou2007multi}
An~Evgeniou and Massimiliano Pontil. 2007.
\newblock Multi-task feature learning.
\newblock In \emph{Proceedings of NeurIPS}.

\bibitem[{Evgeniou and Pontil(2004)}]{evgeniou2004regularized}
Theodoros Evgeniou and Massimiliano Pontil. 2004.
\newblock Regularized multi--task learning.
\newblock In \emph{Proceedings of KDD}, pages 109--117.

\bibitem[{Fan et~al.(2019)Fan, Grave, and Joulin}]{fan2019reducing}
Angela Fan, Edouard Grave, and Armand Joulin. 2019.
\newblock Reducing transformer depth on demand with structured dropout.
\newblock In \emph{Proceedings of ICLR}.

\bibitem[{Fedus et~al.(2021)Fedus, Zoph, and Shazeer}]{fedus2021switch}
William Fedus, Barret Zoph, and Noam Shazeer. 2021.
\newblock Switch transformers: Scaling to trillion parameter models with simple
  and efficient sparsity.
\newblock \emph{arXiv preprint arXiv:2101.03961}.

\bibitem[{F{\'e}vry et~al.(2020)F{\'e}vry, Soares, FitzGerald, Choi, and
  Kwiatkowski}]{fevry2020entities}
Thibault F{\'e}vry, Livio~Baldini Soares, Nicholas FitzGerald, Eunsol Choi, and
  Tom Kwiatkowski. 2020.
\newblock Entities as experts: Sparse memory access with entity supervision.
\newblock In \emph{Proceedings of EMNLP}, pages 4937--4951.

\bibitem[{Forbes et~al.(2019)Forbes, Holtzman, and
  Choi}]{DBLP:conf/cogsci/ForbesHC19}
Maxwell Forbes, Ari Holtzman, and Yejin Choi. 2019.
\newblock Do neural language representations learn physical commonsense?
\newblock In \emph{Proceedings of CogSci}, pages 1753--1759.

\bibitem[{Gao et~al.(2015)Gao, Mao, Zhou, Huang, Wang, and Xu}]{gao2015you}
Haoyuan Gao, Junhua Mao, Jie Zhou, Zhiheng Huang, Lei Wang, and Wei Xu. 2015.
\newblock Are you talking to a machine? dataset and methods for multilingual
  image question answering.
\newblock In \emph{Proceedings of NeurIPS}, pages 2296--2304.

\bibitem[{Gao et~al.(2008)Gao, Fan, Jiang, and Han}]{gao2008knowledge}
Jing Gao, Wei Fan, Jing Jiang, and Jiawei Han. 2008.
\newblock Knowledge transfer via multiple model local structure mapping.
\newblock In \emph{Proceedings of KDD}, pages 283--291.

\bibitem[{Gao et~al.(2021)Gao, Fisch, and Chen}]{gao2021making}
Tianyu Gao, Adam Fisch, and Danqi Chen. 2021.
\newblock Making pre-trained language models better few-shot learners.
\newblock In \emph{Proceedings of ACL}.

\bibitem[{Gidaris and Komodakis(2015)}]{gidaris2015object}
Spyros Gidaris and Nikos Komodakis. 2015.
\newblock Object detection via a multi-region and semantic segmentation-aware
  cnn model.
\newblock In \emph{Proceedings of ICCV}, pages 1134--1142.

\bibitem[{Glava{\v{s}} and Vuli{\'c}(2021)}]{DBLP:journals/corr/abs-2008-06788}
Goran Glava{\v{s}} and Ivan Vuli{\'c}. 2021.
\newblock Is supervised syntactic parsing beneficial for language understanding
  tasks? an empirical investigation.
\newblock In \emph{Proceedings of EACL}, pages 3090--3104.

\bibitem[{Goldberg(2019)}]{DBLP:journals/corr/abs-1901-05287}
Yoav Goldberg. 2019.
\newblock Assessing bert's syntactic abilities.
\newblock \emph{arXiv preprint arXiv:1901.05287}.

\bibitem[{Gong et~al.(2019)Gong, He, Li, Qin, Wang, and Liu}]{pmlr-v97-gong19a}
Linyuan Gong, Di~He, Zhuohan Li, Tao Qin, Liwei Wang, and Tieyan Liu. 2019.
\newblock Efficient training of {BERT} by progressively stacking.
\newblock In \emph{Proceedings of ICML}, pages 2337--2346.

\bibitem[{Gordon et~al.(2020)Gordon, Duh, and
  Andrews}]{DBLP:conf/rep4nlp/GordonDA20}
Mitchell~A. Gordon, Kevin Duh, and Nicholas Andrews. 2020.
\newblock Compressing {BERT:} studying the effects of weight pruning on
  transfer learning.
\newblock In \emph{Proceedings of RepL4NLP}, pages 143--155.

\bibitem[{Goyal et~al.(2017)Goyal, Doll{\'a}r, Girshick, Noordhuis, Wesolowski,
  Kyrola, Tulloch, Jia, and He}]{goyal2017accurate}
Priya Goyal, Piotr Doll{\'a}r, Ross Girshick, Pieter Noordhuis, Lukasz
  Wesolowski, Aapo Kyrola, Andrew Tulloch, Yangqing Jia, and Kaiming He. 2017.
\newblock Accurate, large minibatch sgd: Training imagenet in 1 hour.
\newblock \emph{arXiv preprint arXiv:1706.02677}.

\bibitem[{Gu et~al.(2020)Gu, Zhang, Wang, Liu, and Sun}]{gu2020train}
Yuxian Gu, Zhengyan Zhang, Xiaozhi Wang, Zhiyuan Liu, and Maosong Sun. 2020.
\newblock Train no evil: Selective masking for task-guided pre-training.
\newblock In \emph{Proceedings of EMNLP}, pages 6966--6974.

\bibitem[{Guan et~al.(2020)Guan, Huang, Zhao, Zhu, and
  Huang}]{guan2020knowledge}
Jian Guan, Fei Huang, Zhihao Zhao, Xiaoyan Zhu, and Minlie Huang. 2020.
\newblock A knowledge-enhanced pretraining model for commonsense story
  generation.
\newblock \emph{TACL}, 8:93--108.

\bibitem[{Guo et~al.(2019)Guo, Qiu, Liu, Shao, Xue, and Zhang}]{guo2019star}
Qipeng Guo, Xipeng Qiu, Pengfei Liu, Yunfan Shao, Xiangyang Xue, and Zheng
  Zhang. 2019.
\newblock Star-transformer.
\newblock In \emph{Proceedings of HLT-NAACL}, pages 1315--1325.

\bibitem[{Gupta et~al.(2015)Gupta, Agrawal, Gopalakrishnan, and
  Narayanan}]{pmlr-v37-gupta15}
Suyog Gupta, Ankur Agrawal, Kailash Gopalakrishnan, and Pritish Narayanan.
  2015.
\newblock Deep learning with limited numerical precision.
\newblock In \emph{Proceedings of ICML}, pages 1737--1746.

\bibitem[{Gururangan et~al.(2020)Gururangan, Marasovi{\'c}, Swayamdipta, Lo,
  Beltagy, Downey, and Smith}]{gururangan-etal-2020-dont}
Suchin Gururangan, Ana Marasovi{\'c}, Swabha Swayamdipta, Kyle Lo, Iz~Beltagy,
  Doug Downey, and Noah~A. Smith. 2020.
\newblock Don{'}t stop pretraining: Adapt language models to domains and tasks.
\newblock In \emph{Proceedings of ACL}.

\bibitem[{Guu et~al.(2020)Guu, Lee, Tung, Pasupat, and Chang}]{guu2020realm}
Kelvin Guu, Kenton Lee, Zora Tung, Panupong Pasupat, and Ming-Wei Chang. 2020.
\newblock Realm: Retrieval-augmented language model pre-training.
\newblock \emph{arXiv preprint arXiv:2002.08909}.

\bibitem[{Han et~al.(2021)Han, Zhao, Ding, Liu, and Sun}]{han2021ptr}
Xu~Han, Weilin Zhao, Ning Ding, Zhiyuan Liu, and Maosong Sun. 2021.
\newblock Ptr: Prompt tuning with rules for text classification.
\newblock \emph{arXiv preprint arXiv:2105.11259}.

\bibitem[{Hashemi et~al.(2019)Hashemi, Jyothi, and Campbell}]{hashemi19tictac}
Sayed~Hadi Hashemi, Sangeetha~Abdu Jyothi, and Roy~H Campbell. 2019.
\newblock Tictac: Accelerating distributed deep learning with communication
  scheduling.
\newblock In \emph{Proceedings of MLSys}.

\bibitem[{He et~al.(2021)He, Qiu, Zeng, Yang, Zhai, and Tang}]{he2021fastmoe}
Jiaao He, Jiezhong Qiu, Aohan Zeng, Zhilin Yang, Jidong Zhai, and Jie Tang.
  2021.
\newblock Fastmoe: A fast mixture-of-expert training system.
\newblock \emph{arXiv preprint arXiv:2103.13262}.

\bibitem[{He et~al.(2020)He, Fan, Wu, Xie, and Girshick}]{he2020momentum}
Kaiming He, Haoqi Fan, Yuxin Wu, Saining Xie, and Ross Girshick. 2020.
\newblock Momentum contrast for unsupervised visual representation learning.
\newblock In \emph{Proceedings of CVPR}, pages 9729--9738.

\bibitem[{He et~al.(2019)He, Girshick, and Doll{\'a}r}]{he2019rethinking}
Kaiming He, Ross Girshick, and Piotr Doll{\'a}r. 2019.
\newblock Rethinking imagenet pre-training.
\newblock In \emph{Proceedings of ICCV}, pages 4918--4927.

\bibitem[{He et~al.(2016)He, Zhang, Ren, and Sun}]{he2016deep}
Kaiming He, Xiangyu Zhang, Shaoqing Ren, and Jian Sun. 2016.
\newblock Deep residual learning for image recognition.
\newblock In \emph{Proceedings of CVPR}, pages 770--778.

\bibitem[{Heusel et~al.(2017)Heusel, Ramsauer, Unterthiner, Nessler, and
  Hochreiter}]{heusel2017gans}
Martin Heusel, Hubert Ramsauer, Thomas Unterthiner, Bernhard Nessler, and Sepp
  Hochreiter. 2017.
\newblock Gans trained by a two time-scale update rule converge to a local nash
  equilibrium.
\newblock \emph{Advances in neural information processing systems}, 30.

\bibitem[{Hewitt and Manning(2019)}]{DBLP:conf/naacl/HewittM19}
John Hewitt and Christopher~D. Manning. 2019.
\newblock A structural probe for finding syntax in word representations.
\newblock In \emph{Proceedings of NAACL-HLT}, pages 4129--4138.

\bibitem[{Hinton et~al.(2014)Hinton, Vinyals, and Dean}]{Hinton:NeurIPS2015}
Geoffrey Hinton, Oriol Vinyals, and Jeff Dean. 2014.
\newblock Distilling the knowledge in a neural network.
\newblock In \emph{Proceedings of NeurIPS}.

\bibitem[{Hinton et~al.(2006)Hinton, Osindero, and
  Teh}]{DBLP:journals/neco/HintonOT06}
Geoffrey~E Hinton, Simon Osindero, and Yee-Whye Teh. 2006.
\newblock A fast learning algorithm for deep belief nets.
\newblock \emph{Neural Computation}, 18(7):1527--1554.

\bibitem[{Howard and Ruder(2018)}]{howard2018universal}
Jeremy Howard and Sebastian Ruder. 2018.
\newblock Universal language model fine-tuning for text classification.
\newblock In \emph{Proceedings of ACL}, pages 328--339.

\bibitem[{Htut et~al.(2019)Htut, Phang, Bordia, and
  Bowman}]{DBLP:journals/corr/abs-1911-12246}
Phu~Mon Htut, Jason Phang, Shikha Bordia, and Samuel~R Bowman. 2019.
\newblock Do attention heads in bert track syntactic dependencies?
\newblock \emph{arXiv preprint arXiv:1911.12246}.

\bibitem[{Hu et~al.(2021)Hu, Ding, Wang, Liu, Li, and
  Sun}]{hu2021knowledgeable}
Shengding Hu, Ning Ding, Huadong Wang, Zhiyuan Liu, Juanzi Li, and Maosong Sun.
  2021.
\newblock Knowledgeable prompt-tuning: Incorporating knowledge into prompt
  verbalizer for text classification.
\newblock \emph{arXiv preprint arXiv:2108.02035}.

\bibitem[{Huang et~al.(2020{\natexlab{a}})Huang, Jin, and
  Li}]{Huang2020SwapAdvisorPD}
Chien-Chin Huang, Gu~Jin, and Jinyang Li. 2020{\natexlab{a}}.
\newblock Swapadvisor: Pushing deep learning beyond the gpu memory limit via
  smart swapping.
\newblock In \emph{Proceedings of ASPLOS}, page 1341–1355.

\bibitem[{Huang et~al.(2019{\natexlab{a}})Huang, Liang, Duan, Gong, Shou,
  Jiang, and Zhou}]{huang-etal-2019-unicoder}
Haoyang Huang, Yaobo Liang, Nan Duan, Ming Gong, Linjun Shou, Daxin Jiang, and
  Ming Zhou. 2019{\natexlab{a}}.
\newblock {U}nicoder: A universal language encoder by pre-training with
  multiple cross-lingual tasks.
\newblock In \emph{Proceedings of EMNLP-IJCNLP}, pages 2485--2494.

\bibitem[{Huang et~al.(2020{\natexlab{b}})Huang, Su, Qi, Duan, Cui, Bharti,
  Zhang, Wang, Gao, Liu et~al.}]{huang2020m3p}
Haoyang Huang, Lin Su, Di~Qi, Nan Duan, Edward Cui, Taroon Bharti, Lei Zhang,
  Lijuan Wang, Jianfeng Gao, Bei Liu, et~al. 2020{\natexlab{b}}.
\newblock M3p: Learning universal representations via multitask multilingual
  multimodal pre-training.
\newblock \emph{arXiv preprint arXiv:2006.02635}.

\bibitem[{Huang et~al.(2019{\natexlab{b}})Huang, Cheng, Bapna, Firat, Chen,
  Chen, Lee, Ngiam, Le, Wu et~al.}]{huang2018gpipe}
Yanping Huang, Youlong Cheng, Ankur Bapna, Orhan Firat, Mia~Xu Chen, Dehao
  Chen, HyoukJoong Lee, Jiquan Ngiam, Quoc~V Le, Yonghui Wu, et~al.
  2019{\natexlab{b}}.
\newblock Gpipe: Efficient training of giant neural networks using pipeline
  parallelism.
\newblock In \emph{Proceedings of NeurIPS}, pages 103--112.

\bibitem[{Hudson and Manning(2019)}]{hudson2019gqa}
Drew~A Hudson and Christopher~D Manning. 2019.
\newblock Gqa: A new dataset for real-world visual reasoning and compositional
  question answering.
\newblock In \emph{Proceedings of CVPR}, pages 6700--6709.

\bibitem[{Huo et~al.(2021)Huo, Zhang, Liu, Lu, Gao, Yang, Wen, Zhang, Xu, Zheng
  et~al.}]{huo2021wenlan}
Yuqi Huo, Manli Zhang, Guangzhen Liu, Haoyu Lu, Yizhao Gao, Guoxing Yang,
  Jingyuan Wen, Heng Zhang, Baogui Xu, Weihao Zheng, et~al. 2021.
\newblock Wenlan: Bridging vision and language by large-scale multi-modal
  pre-training.
\newblock \emph{arXiv preprint arXiv:2103.06561}.

\bibitem[{Ioffe and Szegedy(2015)}]{ioffe2015batch}
Sergey Ioffe and Christian Szegedy. 2015.
\newblock Batch normalization: Accelerating deep network training by reducing
  internal covariate shift.
\newblock In \emph{Proceedings of ICML}, pages 448--456.

\bibitem[{Jacobs et~al.(1991)Jacobs, Jordan, Nowlan, and Hinton}]{Jacobs:1991}
Robert~A Jacobs, Michael~I Jordan, Steven~J Nowlan, and Geoffrey~E Hinton.
  1991.
\newblock Adaptive mixtures of local experts.
\newblock \emph{Neural Computation}, 3:79--87.

\bibitem[{Jaderberg et~al.(2015)Jaderberg, Simonyan, Zisserman, and
  Kavukcuoglu}]{jaderberg2015spatial}
Max Jaderberg, Karen Simonyan, Andrew Zisserman, and Koray Kavukcuoglu. 2015.
\newblock Spatial transformer networks.
\newblock In \emph{Proceedings of NeurIPS}, pages 2017--2025.

\bibitem[{Jawahar et~al.(2019{\natexlab{a}})Jawahar, Sagot, and
  Seddah}]{DBLP:conf/acl/JawaharSS19}
Ganesh Jawahar, Beno{\^{\i}}t Sagot, and Djam{\'{e}} Seddah.
  2019{\natexlab{a}}.
\newblock What does {BERT} learn about the structure of language?
\newblock In \emph{Proceedings of ACL}, pages 3651--3657.

\bibitem[{Jawahar et~al.(2019{\natexlab{b}})Jawahar, Sagot, and
  Seddah}]{Jawahar:ACL2019}
Ganesh Jawahar, Beno{\^\i}t Sagot, and Djam{\'e} Seddah. 2019{\natexlab{b}}.
\newblock What does bert learn about the structure of language?
\newblock In \emph{Proceedings of ACL}, pages 3651--3657.

\bibitem[{Jia et~al.(2019)Jia, Zaharia, and Aiken}]{jia-2019-sysml}
Zhihao Jia, Matei Zaharia, and Alex Aiken. 2019.
\newblock Beyond data and model parallelism for deep neural networks.
\newblock In \emph{Proceedings of MLSys}.

\bibitem[{Jiang et~al.(2020{\natexlab{a}})Jiang, Zhu, Lan, Yi, Cui, and
  Guo}]{jiang-osdi-2020}
Yimin Jiang, Yibo Zhu, Chang Lan, Bairen Yi, Yong Cui, and Chuanxiong Guo.
  2020{\natexlab{a}}.
\newblock A unified architecture for accelerating distributed {DNN} training in
  heterogeneous gpu/cpu clusters.
\newblock In \emph{Proceedings of OSDI}, pages 463--479.

\bibitem[{Jiang et~al.(2020{\natexlab{b}})Jiang, Xu, Araki, and
  Neubig}]{DBLP:journals/tacl/JiangXAN20}
Zhengbao Jiang, Frank~F. Xu, Jun Araki, and Graham Neubig. 2020{\natexlab{b}}.
\newblock How can we know what language models know.
\newblock \emph{TACL}, 8:423--438.

\bibitem[{Jiao et~al.(2019)Jiao, Yin, Shang, Jiang, Chen, Li, Wang, and
  Liu}]{jiao2019tinybert}
Xiaoqi Jiao, Yichun Yin, Lifeng Shang, Xin Jiang, Xiao Chen, Linlin Li, Fang
  Wang, and Qun Liu. 2019.
\newblock Tinybert: Distilling bert for natural language understanding.
\newblock In \emph{Proceedings of EMNLP}, pages 4163--4174.

\bibitem[{Jin et~al.(2020)Jin, Jin, Zhou, and Szolovits}]{Jin2019textfooler}
Di~Jin, Zhijing Jin, Joey~Tianyi Zhou, and Peter Szolovits. 2020.
\newblock Is bert really robust? a strong baseline for natural language attack
  on text classification and entailment.
\newblock In \emph{Proceedings of AAAI}, pages 8018--8025.

\bibitem[{Johnson et~al.(2016)Johnson, Karpathy, and
  Fei-Fei}]{johnson2016densecap}
Justin Johnson, Andrej Karpathy, and Li~Fei-Fei. 2016.
\newblock Densecap: Fully convolutional localization networks for dense
  captioning.
\newblock In \emph{Proceedings of CVPR}, pages 4565--4574.

\bibitem[{Johnson and Zhang(2005)}]{johnson2005high}
Rie Johnson and Tong Zhang. 2005.
\newblock A high-performance semi-supervised learning method for text chunking.
\newblock In \emph{Proceedings of ACL}, pages 1--9.

\bibitem[{Joshi et~al.(2020)Joshi, Chen, Liu, Weld, Zettlemoyer, and
  Levy}]{joshi2020spanbert}
Mandar Joshi, Danqi Chen, Yinhan Liu, Daniel~S Weld, Luke Zettlemoyer, and Omer
  Levy. 2020.
\newblock Spanbert: Improving pre-training by representing and predicting
  spans.
\newblock \emph{TACL}, 8:64--77.

\bibitem[{Kalchbrenner et~al.(2014)Kalchbrenner, Grefenstette, and
  Blunsom}]{kalchbrenner2014convolutional}
Nal Kalchbrenner, Edward Grefenstette, and Phil Blunsom. 2014.
\newblock A convolutional neural network for modelling sentences.
\newblock In \emph{Proceedings of ACL}, pages 655--665.

\bibitem[{Kao et~al.(2020)Kao, Wu, Chi, Hsieh, and Lee}]{kao2021berts}
Wei-Tsung Kao, Tsung-Han Wu, Po-Han Chi, Chun-Cheng Hsieh, and Hung-Yi Lee.
  2020.
\newblock Further boosting bert-based models by duplicating existing layers:
  Some intriguing phenomena inside bert.
\newblock \emph{arXiv preprint arXiv:2001.09309}.

\bibitem[{Kaplan et~al.(2020)Kaplan, McCandlish, Henighan, Brown, Chess, Child,
  Gray, Radford, Wu, and Amodei}]{kaplan-scaling-2020}
Jared Kaplan, Sam McCandlish, Tom Henighan, Tom~B. Brown, Benjamin Chess, Rewon
  Child, Scott Gray, Alec Radford, Jeffrey Wu, and Dario Amodei. 2020.
\newblock Scaling laws for neural language models.
\newblock \emph{arXiv preprint arXiv:2001.08361}.

\bibitem[{Katharopoulos et~al.(2020)Katharopoulos, Vyas, Pappas, and
  Fleuret}]{katharopoulos2020transformers}
Angelos Katharopoulos, Apoorv Vyas, Nikolaos Pappas, and Fran{\c{c}}ois
  Fleuret. 2020.
\newblock Transformers are rnns: Fast autoregressive transformers with linear
  attention.
\newblock In \emph{Proceedings of ICML}, pages 5156--5165.

\bibitem[{Ke et~al.(2020)Ke, Ji, Liu, Zhu, and Huang}]{ke2020sentilare}
Pei Ke, Haozhe Ji, Siyang Liu, Xiaoyan Zhu, and Minlie Huang. 2020.
\newblock Sentilare: Linguistic knowledge enhanced language representation for
  sentiment analysis.
\newblock In \emph{Proceedings of EMNLP}, pages 6975--6988.

\bibitem[{Kim et~al.(2020)Kim, Choi, Edmiston, and
  Lee}]{DBLP:conf/iclr/KimCEL20}
Taeuk Kim, Jihun Choi, Daniel Edmiston, and Sang{-}goo Lee. 2020.
\newblock Are pre-trained language models aware of phrases? simple but strong
  baselines for grammar induction.
\newblock In \emph{Proceedings of ICLR}.

\bibitem[{Kim(2014)}]{kim2014convolutional}
Yoon Kim. 2014.
\newblock Convolutional neural networks for sentence classification.
\newblock In \emph{Proceedings of EMNLP}, pages 1746--1751.

\bibitem[{Kipf and Welling(2016)}]{kipf2016semi}
Thomas~N Kipf and Max Welling. 2016.
\newblock Semi-supervised classification with graph convolutional networks.
\newblock In \emph{Proceedings of ICLR}.

\bibitem[{Kitaev et~al.(2020)Kitaev, Kaiser, and Levskaya}]{kitaev2020reformer}
Nikita Kitaev, {\L}ukasz Kaiser, and Anselm Levskaya. 2020.
\newblock Reformer: The efficient transformer.
\newblock In \emph{Proceedings of ICLR}.

\bibitem[{K{\"{o}}hn(2015)}]{DBLP:conf/emnlp/Kohn15}
Arne K{\"{o}}hn. 2015.
\newblock What's in an embedding? analyzing word embeddings through
  multilingual evaluation.
\newblock In \emph{Proceedings of EMNLP}, pages 2067--2073.

\bibitem[{Kong et~al.(2020)Kong, de~Masson~d'Autume, Yu, Ling, Dai, and
  Yogatama}]{DBLP:conf/iclr/KongdYLDY20}
Lingpeng Kong, Cyprien de~Masson~d'Autume, Lei Yu, Wang Ling, Zihang Dai, and
  Dani Yogatama. 2020.
\newblock A mutual information maximization perspective of language
  representation learning.
\newblock In \emph{Proceedings of ICLR}.

\bibitem[{Kovaleva et~al.(2019)Kovaleva, Romanov, Rogers, and
  Rumshisky}]{DBLP:conf/emnlp/KovalevaRRR19}
Olga Kovaleva, Alexey Romanov, Anna Rogers, and Anna Rumshisky. 2019.
\newblock Revealing the dark secrets of {BERT}.
\newblock In \emph{Proceedings of EMNLP-IJCNLP}, pages 4364--4373.

\bibitem[{Krishna et~al.(2017)Krishna, Zhu, Groth, Johnson, Hata, Kravitz,
  Chen, Kalantidis, Li, Shamma et~al.}]{krishna2017visual}
Ranjay Krishna, Yuke Zhu, Oliver Groth, Justin Johnson, Kenji Hata, Joshua
  Kravitz, Stephanie Chen, Yannis Kalantidis, Li-Jia Li, David~A Shamma, et~al.
  2017.
\newblock Visual genome: Connecting language and vision using crowdsourced
  dense image annotations.
\newblock \emph{IJCV}, 123:32--73.

\bibitem[{Krizhevsky et~al.(2012)Krizhevsky, Sutskever, and
  Hinton}]{krizhevsky2012imagenet}
Alex Krizhevsky, Ilya Sutskever, and Geoffrey~E Hinton. 2012.
\newblock {ImageNet} classification with deep convolutional neural networks.
\newblock In \emph{Proceedings of NeurIPS}, pages 1097--1105.

\bibitem[{Lample and Conneau(2019)}]{lample2019cross}
Guillaume Lample and Alexis Conneau. 2019.
\newblock Cross-lingual language model pretraining.
\newblock \emph{Proceedings of NeurIPS}.

\bibitem[{Lample et~al.(2019)Lample, Sablayrolles, Ranzato, Denoyer, and
  Jégou}]{lample2019large}
Guillaume Lample, Alexandre Sablayrolles, Marc'Aurelio Ranzato, Ludovic
  Denoyer, and Hervé Jégou. 2019.
\newblock Large memory layers with product keys.
\newblock In \emph{Proceedings of NeurIPS}, pages 8546--8557.

\bibitem[{Lan et~al.(2019)Lan, Chen, Goodman, Gimpel, Sharma, and
  Soricut}]{lan2019albert}
Zhenzhong Lan, Mingda Chen, Sebastian Goodman, Kevin Gimpel, Piyush Sharma, and
  Radu Soricut. 2019.
\newblock Albert: A lite bert for self-supervised learning of language
  representations.
\newblock In \emph{Proceedings of ICLR}.

\bibitem[{Lawrence and Platt(2004)}]{lawrence2004learning}
Neil~D Lawrence and John~C Platt. 2004.
\newblock Learning to learn with the informative vector machine.
\newblock In \emph{Proceedings of ICML}.

\bibitem[{LeCun et~al.(2012)LeCun, Bottou, Orr, and
  M{\"u}ller}]{lecun2012efficient}
Yann~A LeCun, L{\'e}on Bottou, Genevieve~B Orr, and Klaus-Robert M{\"u}ller.
  2012.
\newblock Efficient backprop.
\newblock In \emph{Neural networks: Tricks of the trade}, pages 9--48.
  Springer.

\bibitem[{Lee et~al.(2015)Lee, Xie, Gallagher, Zhang, and Tu}]{lee2015deeply}
Chen-Yu Lee, Saining Xie, Patrick Gallagher, Zhengyou Zhang, and Zhuowen Tu.
  2015.
\newblock Deeply-supervised nets.
\newblock In \emph{Proceedings of AISTATS}, pages 562--570.

\bibitem[{Lee et~al.(2020)Lee, Yoon, Kim, Kim, Kim, So, and
  Kang}]{lee2020biobert}
Jinhyuk Lee, Wonjin Yoon, Sungdong Kim, Donghyeon Kim, Sunkyu Kim, Chan~Ho So,
  and Jaewoo Kang. 2020.
\newblock Biobert: a pre-trained biomedical language representation model for
  biomedical text mining.
\newblock \emph{Bioinformatics}, 36(4):1234--1240.

\bibitem[{Lee et~al.(2019)Lee, Lee, Kim, Kosiorek, Choi, and Teh}]{lee2019set}
Juho Lee, Yoonho Lee, Jungtaek Kim, Adam Kosiorek, Seungjin Choi, and Yee~Whye
  Teh. 2019.
\newblock Set transformer: A framework for attention-based
  permutation-invariant neural networks.
\newblock In \emph{Proceedings of ICML}, pages 3744--3753.

\bibitem[{Lepikhin et~al.(2021)Lepikhin, Lee, Xu, Chen, Firat, Huang, Krikun,
  Shazeer, and Chen}]{lepikhin2020gshard}
Dmitry Lepikhin, HyoukJoong Lee, Yuanzhong Xu, Dehao Chen, Orhan Firat, Yanping
  Huang, Maxim Krikun, Noam Shazeer, and Zhifeng Chen. 2021.
\newblock Gshard: Scaling giant models with conditional computation and
  automatic sharding.
\newblock In \emph{Proceedings of ICLR}.

\bibitem[{Lester et~al.(2021)Lester, Al-Rfou, and Constant}]{lester2021power}
Brian Lester, Rami Al-Rfou, and Noah Constant. 2021.
\newblock The power of scale for parameter-efficient prompt tuning.
\newblock \emph{arXiv preprint arXiv:2104.08691}.

\bibitem[{Lewis et~al.(2020{\natexlab{a}})Lewis, Liu, Goyal, Ghazvininejad,
  Mohamed, Levy, Stoyanov, and Zettlemoyer}]{lewis2020bart}
Mike Lewis, Yinhan Liu, Naman Goyal, Marjan Ghazvininejad, Abdelrahman Mohamed,
  Omer Levy, Veselin Stoyanov, and Luke Zettlemoyer. 2020{\natexlab{a}}.
\newblock {BART}: Denoising sequence-to-sequence pre-training for natural
  language generation, translation, and comprehension.
\newblock In \emph{Proceedings of ACL}, pages 7871--7880.

\bibitem[{Lewis et~al.(2020{\natexlab{b}})Lewis, Perez, Piktus, Petroni,
  Karpukhin, Goyal, K{\"u}ttler, Lewis, Yih, Rockt{\"a}schel
  et~al.}]{lewis2020retrieval}
Patrick Lewis, Ethan Perez, Aleksandara Piktus, Fabio Petroni, Vladimir
  Karpukhin, Naman Goyal, Heinrich K{\"u}ttler, Mike Lewis, Wen-tau Yih, Tim
  Rockt{\"a}schel, et~al. 2020{\natexlab{b}}.
\newblock Retrieval-augmented generation for knowledge-intensive nlp tasks.
\newblock In \emph{Proceedings of NeurIPS}, pages 9459--9474.

\bibitem[{Li et~al.(2020{\natexlab{a}})Li, Duan, Fang, Gong, and
  Jiang}]{li2020unicoder}
Gen Li, Nan Duan, Yuejian Fang, Ming Gong, and Daxin Jiang. 2020{\natexlab{a}}.
\newblock Unicoder-vl: A universal encoder for vision and language by
  cross-modal pre-training.
\newblock In \emph{Proceedings of AAAI}, pages 11336--11344.

\bibitem[{Li et~al.(2020{\natexlab{b}})Li, Ma, Guo, Xue, and Qiu}]{li2020bert}
Linyang Li, Ruotian Ma, Qipeng Guo, Xiangyang Xue, and Xipeng Qiu.
  2020{\natexlab{b}}.
\newblock {BERT-ATTACK}: Adversarial attack against bert using bert.
\newblock In \emph{Proceedings of EMNLP}, pages 6193--6202.

\bibitem[{Li and Qiu(2021)}]{li2021token}
Linyang Li and Xipeng Qiu. 2021.
\newblock Token-aware virtual adversarial training in natural language
  understanding.
\newblock In \emph{Proceedings of AAAI}, pages 8410--8418.

\bibitem[{Li et~al.(2020{\natexlab{c}})Li, Shao, Song, Qiu, and
  Huang}]{li2020generating}
Linyang Li, Yunfan Shao, Demin Song, Xipeng Qiu, and Xuanjing Huang.
  2020{\natexlab{c}}.
\newblock Generating adversarial examples in chinese texts using
  sentence-pieces.
\newblock \emph{arXiv preprint arXiv:2012.14769}.

\bibitem[{Li et~al.(2019)Li, Yatskar, Yin, Hsieh, and Chang}]{li2019visualbert}
Liunian~Harold Li, Mark Yatskar, Da~Yin, Cho-Jui Hsieh, and Kai-Wei Chang.
  2019.
\newblock {VisualBERT}: A simple and performant baseline for vision and
  language.
\newblock \emph{arXiv preprint arXiv:1908.03557}.

\bibitem[{Li et~al.(2020{\natexlab{d}})Li, Zhao, Varma, Salpekar, Noordhuis,
  Li, Paszke, Smith, Vaughan, Damania, and Chintala}]{li2020pytorch}
Shen Li, Yanli Zhao, Rohan Varma, Omkar Salpekar, Pieter Noordhuis, Teng Li,
  Adam Paszke, Jeff Smith, Brian Vaughan, Pritam Damania, and Soumith Chintala.
  2020{\natexlab{d}}.
\newblock Pytorch distributed: Experiences on accelerating data parallel
  training.
\newblock In \emph{Proceedings of PVLDB}, page 3005–3018.

\bibitem[{Li et~al.(2020{\natexlab{e}})Li, Yin, Li, Zhang, Hu, Zhang, Wang, Hu,
  Dong, Wei et~al.}]{li2020oscar}
Xiujun Li, Xi~Yin, Chunyuan Li, Pengchuan Zhang, Xiaowei Hu, Lei Zhang, Lijuan
  Wang, Houdong Hu, Li~Dong, Furu Wei, et~al. 2020{\natexlab{e}}.
\newblock Oscar: Object-semantics aligned pre-training for vision-language
  tasks.
\newblock In \emph{Proceedings of ECCV}, pages 121--137.

\bibitem[{Li et~al.(2021)Li, Zhuang, Guo, Zhuo, Zhang, Song, and
  Stoica}]{li2021terapipe}
Zhuohan Li, Siyuan Zhuang, Shiyuan Guo, Danyang Zhuo, Hao Zhang, Dawn Song, and
  Ion Stoica. 2021.
\newblock Terapipe: Token-level pipeline parallelism for training large-scale
  language models.
\newblock \emph{arXiv preprint arXiv:2102.07988}.

\bibitem[{Lin et~al.(2021)Lin, Wang, Liu, and Qiu}]{lin2021surveytransformers}
Tianyang Lin, Yuxin Wang, Xiangyang Liu, and Xipeng Qiu. 2021.
\newblock A survey of transformers.
\newblock \emph{arXiv preprint arXiv:2106.04554}.

\bibitem[{Lin et~al.(2014)Lin, Maire, Belongie, Hays, Perona, Ramanan,
  Doll{\'a}r, and Zitnick}]{lin2014microsoft}
Tsung-Yi Lin, Michael Maire, Serge Belongie, James Hays, Pietro Perona, Deva
  Ramanan, Piotr Doll{\'a}r, and C~Lawrence Zitnick. 2014.
\newblock Microsoft coco: Common objects in context.
\newblock In \emph{Proceedings of ECCV}, pages 740--755.

\bibitem[{Lin et~al.(2019)Lin, Tan, and
  Frank}]{DBLP:journals/corr/abs-1906-01698}
Yongjie Lin, Yi~Chern Tan, and Robert Frank. 2019.
\newblock Open sesame: Getting inside bert’s linguistic knowledge.
\newblock In \emph{Proceedings of BlackboxNLP}, pages 241--253.

\bibitem[{Liu et~al.(2019)Liu, Gardner, Belinkov, Peters, and
  Smith}]{DBLP:conf/naacl/Liu0BPS19}
Nelson~F. Liu, Matt Gardner, Yonatan Belinkov, Matthew~E. Peters, and Noah~A.
  Smith. 2019.
\newblock Linguistic knowledge and transferability of contextual
  representations.
\newblock In \emph{Proceedings of NAACL-HLT}, pages 1073--1094.

\bibitem[{Liu et~al.(2016)Liu, Qiu, and Huang}]{liu2016recurrent}
Pengfei Liu, Xipeng Qiu, and Xuanjing Huang. 2016.
\newblock Recurrent neural network for text classification with multi-task
  learning.
\newblock In \emph{Proceedings of IJCAI}, pages 2873--2879.

\bibitem[{Liu et~al.(2020{\natexlab{a}})Liu, Zhou, Zhao, Wang, Ju, Deng, and
  Wang}]{liu2020k}
Weijie Liu, Peng Zhou, Zhe Zhao, Zhiruo Wang, Qi~Ju, Haotang Deng, and Ping
  Wang. 2020{\natexlab{a}}.
\newblock K-bert: Enabling language representation with knowledge graph.
\newblock In \emph{Proceedings of AAAI}, pages 2901--2908.

\bibitem[{Liu et~al.(2021{\natexlab{a}})Liu, Yin, Zhang, Su, Wu, Yang, and
  Tang}]{liu2021oag}
Xiao Liu, Da~Yin, Xingjian Zhang, Kai Su, Kan Wu, Hongxia Yang, and Jie Tang.
  2021{\natexlab{a}}.
\newblock Oag-bert: Pre-train heterogeneous entity-augmented academic language
  model.
\newblock \emph{arXiv preprint arXiv:2103.02410}.

\bibitem[{Liu et~al.(2020{\natexlab{b}})Liu, Zhang, Hou, Wang, Mian, Zhang, and
  Tang}]{liu2020self}
Xiao Liu, Fanjin Zhang, Zhenyu Hou, Zhaoyu Wang, Li~Mian, Jing Zhang, and Jie
  Tang. 2020{\natexlab{b}}.
\newblock Self-supervised learning: Generative or contrastive.
\newblock \emph{arXiv preprint arXiv:2006.08218}.

\bibitem[{Liu et~al.(2021{\natexlab{b}})Liu, Zheng, Du, Ding, Qian, Yang, and
  Tang}]{liu2021gpt}
Xiao Liu, Yanan Zheng, Zhengxiao Du, Ming Ding, Yujie Qian, Zhilin Yang, and
  Jie Tang. 2021{\natexlab{b}}.
\newblock Gpt understands, too.
\newblock \emph{arXiv preprint arXiv:2103.10385}.

\bibitem[{Liu et~al.(2020{\natexlab{c}})Liu, Gu, Goyal, Li, Edunov,
  Ghazvininejad, Lewis, and Zettlemoyer}]{10.1162/tacl_a_00343}
Yinhan Liu, Jiatao Gu, Naman Goyal, Xian Li, Sergey Edunov, Marjan
  Ghazvininejad, Mike Lewis, and Luke Zettlemoyer. 2020{\natexlab{c}}.
\newblock {Multilingual Denoising Pre-training for Neural Machine Translation}.
\newblock \emph{TACL}, 8:726--742.

\bibitem[{Liu et~al.(2020{\natexlab{d}})Liu, Ott, Goyal, Du, Joshi, Chen, Levy,
  Lewis, Zettlemoyer, and Stoyanov}]{liu2019roberta}
Yinhan Liu, Myle Ott, Naman Goyal, Jingfei Du, Mandar Joshi, Danqi Chen, Omer
  Levy, Mike Lewis, Luke Zettlemoyer, and Veselin Stoyanov. 2020{\natexlab{d}}.
\newblock Roberta: A robustly optimized bert pretraining approach.
\newblock In \emph{Proceedings of ICLR}.

\bibitem[{Liu et~al.(2021{\natexlab{c}})Liu, Lin, Cao, Hu, Wei, Zhang, Lin, and
  Guo}]{liu2021swin}
Ze~Liu, Yutong Lin, Yue Cao, Han Hu, Yixuan Wei, Zheng Zhang, Stephen Lin, and
  Baining Guo. 2021{\natexlab{c}}.
\newblock Swin transformer: Hierarchical vision transformer using shifted
  windows.
\newblock \emph{arXiv preprint arXiv:2103.14030}.

\bibitem[{Long et~al.(2015)Long, Shelhamer, and Darrell}]{long2015fully}
Jonathan Long, Evan Shelhamer, and Trevor Darrell. 2015.
\newblock Fully convolutional networks for semantic segmentation.
\newblock In \emph{Proceedings of CVPR}, pages 3431--3440.

\bibitem[{Lu et~al.(2019)Lu, Batra, Parikh, and Lee}]{lu2019vilbert}
Jiasen Lu, Dhruv Batra, Devi Parikh, and Stefan Lee. 2019.
\newblock Vilbert: Pretraining task-agnostic visiolinguistic representations
  for vision-and-language tasks.
\newblock In \emph{Proceedings of NeurIPS Reproducibility Challenge}.

\bibitem[{Lu et~al.(2020)Lu, Goswami, Rohrbach, Parikh, and Lee}]{lu202012}
Jiasen Lu, Vedanuj Goswami, Marcus Rohrbach, Devi Parikh, and Stefan Lee. 2020.
\newblock 12-in-1: Multi-task vision and language representation learning.
\newblock In \emph{Proceedings of CVPR}, pages 10437--10446.

\bibitem[{Manning et~al.(2020)Manning, Clark, Hewitt, Khandelwal, and
  Levy}]{DBLP:journals/pnas/ManningCHKL20}
Christopher~D Manning, Kevin Clark, John Hewitt, Urvashi Khandelwal, and Omer
  Levy. 2020.
\newblock Emergent linguistic structure in artificial neural networks trained
  by self-supervision.
\newblock \emph{PNAS}, 117(48):30046--30054.

\bibitem[{McCann et~al.(2017)McCann, Bradbury, Xiong, and
  Socher}]{DBLP:conf/nips/McCannBXS17}
Bryan McCann, James Bradbury, Caiming Xiong, and Richard Socher. 2017.
\newblock Learned in translation: Contextualized word vectors.
\newblock In \emph{Proceedings of NeurIPS}, pages 6294--6305.

\bibitem[{Melamud et~al.(2016)Melamud, Goldberger, and
  Dagan}]{melamud2016context2vec}
Oren Melamud, Jacob Goldberger, and Ido Dagan. 2016.
\newblock context2vec: Learning generic context embedding with bidirectional
  lstm.
\newblock In \emph{Proceedings of CoNLL}, pages 51--61.

\bibitem[{Miaschi and Dell'Orletta(2020)}]{DBLP:conf/rep4nlp/MiaschiD20}
Alessio Miaschi and Felice Dell'Orletta. 2020.
\newblock Contextual and non-contextual word embeddings: an in-depth linguistic
  investigation.
\newblock In \emph{Proceedings of RepL4NLP}, pages 110--119.

\bibitem[{Michel et~al.(2019)Michel, Levy, and
  Neubig}]{DBLP:conf/nips/MichelLN19}
Paul Michel, Omer Levy, and Graham Neubig. 2019.
\newblock Are sixteen heads really better than one?
\newblock In \emph{Proceedings of NeurIPS}, pages 14014--14024.

\bibitem[{Micikevicius et~al.(2018)Micikevicius, Narang, Alben, Diamos, Elsen,
  Garcia, Ginsburg, Houston, Kuchaiev, Venkatesh
  et~al.}]{micikevicius2018mixed}
Paulius Micikevicius, Sharan Narang, Jonah Alben, Gregory Diamos, Erich Elsen,
  David Garcia, Boris Ginsburg, Michael Houston, Oleksii Kuchaiev, Ganesh
  Venkatesh, et~al. 2018.
\newblock Mixed precision training.
\newblock In \emph{Proceedings of ICLR}.

\bibitem[{Mihalkova et~al.(2007)Mihalkova, Huynh, and
  Mooney}]{mihalkova2007mapping}
Lilyana Mihalkova, Tuyen Huynh, and Raymond~J Mooney. 2007.
\newblock Mapping and revising markov logic networks for transfer learning.
\newblock In \emph{Proceedings of AAAI}, pages 608--614.

\bibitem[{Mikolov et~al.(2013{\natexlab{a}})Mikolov, Chen, Corrado, and
  Dean}]{mikolov2013efficient}
Tomas Mikolov, Kai Chen, Greg Corrado, and Jeffrey Dean. 2013{\natexlab{a}}.
\newblock Efficient estimation of word representations in vector space.
\newblock In \emph{Proceedings of ICLR Workshop}.

\bibitem[{Mikolov et~al.(2013{\natexlab{b}})Mikolov, Sutskever, Chen, Corrado,
  and Dean}]{mikolov2013distributed}
Tomas Mikolov, Ilya Sutskever, Kai Chen, Greg Corrado, and Jeffrey Dean.
  2013{\natexlab{b}}.
\newblock Distributed representations of words and phrases and their
  compositionality.
\newblock In \emph{Proceedings of NeurIPS}.

\bibitem[{Mikolov et~al.(2013{\natexlab{c}})Mikolov, Yih, and
  Zweig}]{mikolov2013linguistic}
Tom{\'a}{\v{s}} Mikolov, Wen-tau Yih, and Geoffrey Zweig. 2013{\natexlab{c}}.
\newblock Linguistic regularities in continuous space word representations.
\newblock In \emph{Proceedings of NAACL-HLT}, pages 746--751.

\bibitem[{Narayanan et~al.(2019)Narayanan, Harlap, Phanishayee, Seshadri,
  Devanur, Ganger, Gibbons, and Zaharia}]{narayanan-2019-sosp}
Deepak Narayanan, Aaron Harlap, Amar Phanishayee, Vivek Seshadri, Nikhil~R.
  Devanur, Gregory~R. Ganger, Phillip~B. Gibbons, and Matei Zaharia. 2019.
\newblock Pipedream: Generalized pipeline parallelism for dnn training.
\newblock In \emph{Proceedings of SOSP}.

\bibitem[{Narayanan et~al.(2021)Narayanan, Shoeybi, Casper, LeGresley, Patwary,
  Korthikanti, Vainbrand, Kashinkunti, Bernauer, Catanzaro, Phanishayee, and
  Zaharia}]{narayanan-megatron-2021}
Deepak Narayanan, Mohammad Shoeybi, Jared Casper, Patrick LeGresley, Mostofa
  Patwary, Vijay Korthikanti, Dmitri Vainbrand, Prethvi Kashinkunti, Julie
  Bernauer, Bryan Catanzaro, Amar Phanishayee, and Matei Zaharia. 2021.
\newblock Efficient large-scale language model training on gpu clusters.
\newblock \emph{arXiv preprint arXiv:2104.04473}.

\bibitem[{Nie et~al.(2020)Nie, Williams, Dinan, Bansal, Weston, and
  Kiela}]{nie2020adversarial}
Yixin Nie, Adina Williams, Emily Dinan, Mohit Bansal, Jason Weston, and Douwe
  Kiela. 2020.
\newblock Adversarial nli: A new benchmark for natural language understanding.
\newblock In \emph{Proceedings of ACL}, pages 4885--4901.

\bibitem[{Niven and Kao(2019)}]{niven2019probing}
Timothy Niven and Hung-Yu Kao. 2019.
\newblock Probing neural network comprehension of natural language arguments.
\newblock In \emph{Proceedings of ACL}, pages 4658--4664.

\bibitem[{Oldridge et~al.(2020)Oldridge, Perez, Frederickson, Koumchatzky, Lee,
  Wang, Wu, Yu, Zamora, Yılmaz, Gunny, Nguyen, and Lee}]{Oldridge2020MerlinAG}
Even Oldridge, J.~Perez, Ben Frederickson, Nicolas Koumchatzky, M.~Lee, Z.-H.
  Wang, Lei Wu, F.~Yu, Rick Zamora, O.~Yılmaz, Alec~M. Gunny, Vinh~Phu Nguyen,
  and S.~Lee. 2020.
\newblock Merlin: A gpu accelerated recommendation framework.
\newblock In \emph{Proceedings of IRS}.

\bibitem[{Ordonez et~al.(2011)Ordonez, Kulkarni, and Berg}]{ordonez2011sbu}
Vicente Ordonez, Girish Kulkarni, and Tamara Berg. 2011.
\newblock Im2text: Describing images using 1 million captioned photographs.
\newblock \emph{Advances in neural information processing systems},
  24:1143--1151.

\bibitem[{Ouyang et~al.(2020)Ouyang, Wang, Pang, Sun, Tian, Wu, and
  Wang}]{2020arXiv201215674O}
Xuan Ouyang, Shuohuan Wang, Chao Pang, Yu~Sun, Hao Tian, Hua Wu, and Haifeng
  Wang. 2020.
\newblock {ERNIE-M}: Enhanced multilingual representation by aligning
  cross-lingual semantics with monolingual corpora.
\newblock \emph{arXiv preprint arXiv:2012.15674}.

\bibitem[{Pan and Yang(2009)}]{pan2009survey}
Sinno~Jialin Pan and Qiang Yang. 2009.
\newblock A survey on transfer learning.
\newblock \emph{IEEE TKDE}, 22(10):1345--1359.

\bibitem[{Pang et~al.(2020)Pang, Xu, Dong, Du, Chen, and
  Zhu}]{pang2019rethinking}
Tianyu Pang, Kun Xu, Yinpeng Dong, Chao Du, Ning Chen, and Jun Zhu. 2020.
\newblock Rethinking softmax cross-entropy loss for adversarial robustness.
\newblock In \emph{Proceedings of ICLR}.

\bibitem[{Paszke et~al.(2019)Paszke, Gross, Massa, Lerer, Bradbury, Chanan,
  Killeen, Lin, Gimelshein, Antiga, Desmaison, Kopf, Yang, DeVito, Raison,
  Tejani, Chilamkurthy, Steiner, Fang, Bai, and Chintala}]{paszke-2019-pytorch}
Adam Paszke, Sam Gross, Francisco Massa, Adam Lerer, James Bradbury, Gregory
  Chanan, Trevor Killeen, Zeming Lin, Natalia Gimelshein, Luca Antiga, Alban
  Desmaison, Andreas Kopf, Edward Yang, Zachary DeVito, Martin Raison, Alykhan
  Tejani, Sasank Chilamkurthy, Benoit Steiner, Lu~Fang, Junjie Bai, and Soumith
  Chintala. 2019.
\newblock Pytorch: An imperative style, high-performance deep learning library.
\newblock In \emph{Proceedings of NeurIPS}.

\bibitem[{Peng et~al.(2021)Peng, Pappas, Yogatama, Schwartz, Smith, and
  Kong}]{peng2021random}
Hao Peng, Nikolaos Pappas, Dani Yogatama, Roy Schwartz, Noah~A Smith, and
  Lingpeng Kong. 2021.
\newblock Random feature attention.
\newblock In \emph{Proceedings of ICLR}.

\bibitem[{Peng et~al.(2019)Peng, Zhu, Chen, Bao, Yi, Lan, Wu, and
  Guo}]{peng-2019-sosp}
Yanghua Peng, Yibo Zhu, Yangrui Chen, Yixin Bao, Bairen Yi, Chang Lan, Chuan
  Wu, and Chuanxiong Guo. 2019.
\newblock A generic communication scheduler for distributed dnn training
  acceleration.
\newblock In \emph{Proceedings of SOSP}, pages 16--29.

\bibitem[{Pennington et~al.(2014)Pennington, Socher, and
  Manning}]{pennington2014glove}
Jeffrey Pennington, Richard Socher, and Christopher~D Manning. 2014.
\newblock Glove: Global vectors for word representation.
\newblock In \emph{Proceedings of EMNLP}, pages 1532--1543.

\bibitem[{Peters et~al.(2018)Peters, Neumann, Iyyer, Gardner, Clark, Lee, and
  Zettlemoyer}]{peters2018deep}
Matthew Peters, Mark Neumann, Mohit Iyyer, Matt Gardner, Christopher Clark,
  Kenton Lee, and Luke Zettlemoyer. 2018.
\newblock Deep contextualized word representations.
\newblock In \emph{Proceedings of NAACL-HLT}, pages 2227--2237.

\bibitem[{Peters et~al.(2019)Peters, Neumann, Logan~IV, Schwartz, Joshi, Singh,
  and Smith}]{peters2019knowledge}
Matthew~E Peters, Mark Neumann, Robert~L Logan~IV, Roy Schwartz, Vidur Joshi,
  Sameer Singh, and Noah~A Smith. 2019.
\newblock Knowledge enhanced contextual word representations.
\newblock In \emph{Proceedings of EMNLP-IJCNLP}, pages 43--54.

\bibitem[{Petroni et~al.(2019)Petroni, Rockt{\"a}schel, Riedel, Lewis, Bakhtin,
  Wu, and Miller}]{Petroni:EMNLP2019}
Fabio Petroni, Tim Rockt{\"a}schel, Sebastian Riedel, Patrick Lewis, Anton
  Bakhtin, Yuxiang Wu, and Alexander Miller. 2019.
\newblock Language models as knowledge bases?
\newblock In \emph{Proceedings of EMNLP-IJCNLP}, pages 2463--2473.

\bibitem[{Pires et~al.(2019)Pires, Schlinger, and
  Garrette}]{pires-etal-2019-multilingual}
Telmo Pires, Eva Schlinger, and Dan Garrette. 2019.
\newblock How multilingual is multilingual {BERT}?
\newblock In \emph{Proceedings of ACL}, pages 4996--5001.

\bibitem[{Plummer et~al.(2015)Plummer, Wang, Cervantes, Caicedo, Hockenmaier,
  and Lazebnik}]{plummer2015flickr30k}
Bryan~A Plummer, Liwei Wang, Chris~M Cervantes, Juan~C Caicedo, Julia
  Hockenmaier, and Svetlana Lazebnik. 2015.
\newblock Flickr30k entities: Collecting region-to-phrase correspondences for
  richer image-to-sentence models.
\newblock In \emph{Proceedings of ICCV}, pages 2641--2649.

\bibitem[{Polino et~al.(2018)Polino, Pascanu, and Alistarh}]{polino2018model}
Antonio Polino, Razvan Pascanu, and Dan Alistarh. 2018.
\newblock Model compression via distillation and quantization.
\newblock In \emph{Proceedings of ICLR}.

\bibitem[{P{\"{o}}rner et~al.(2020)P{\"{o}}rner, Waltinger, and
  Sch{\"{u}}tze}]{DBLP:conf/emnlp/PornerWS20}
Nina P{\"{o}}rner, Ulli Waltinger, and Hinrich Sch{\"{u}}tze. 2020.
\newblock {E-BERT:} efficient-yet-effective entity embeddings for {BERT}.
\newblock In \emph{Proceedings of EMNLP}, pages 803--818.

\bibitem[{Prasanna et~al.(2020)Prasanna, Rogers, and
  Rumshisky}]{DBLP:conf/emnlp/PrasannaRR20}
Sai Prasanna, Anna Rogers, and Anna Rumshisky. 2020.
\newblock When {BERT} plays the lottery, all tickets are winning.
\newblock In \emph{Proceedings of EMNLP}, pages 3208--3229.

\bibitem[{Qi et~al.(2020)Qi, Su, Song, Cui, Bharti, and
  Sacheti}]{qi2020imagebert}
Di~Qi, Lin Su, Jia Song, Edward Cui, Taroon Bharti, and Arun Sacheti. 2020.
\newblock Imagebert: Cross-modal pre-training with large-scale weak-supervised
  image-text data.
\newblock \emph{arXiv preprint arXiv:2001.07966}.

\bibitem[{Qin et~al.(2021)Qin, Lin, Takanobu, Liu, Li, Ji, Huang, Sun, and
  Zhou}]{qin2020erica}
Yujia Qin, Yankai Lin, Ryuichi Takanobu, Zhiyuan Liu, Peng Li, Heng Ji, Minlie
  Huang, Maosong Sun, and Jie Zhou. 2021.
\newblock Erica: Improving entity and relation understanding for pre-trained
  language models via contrastive learning.
\newblock In \emph{Proceedings of ACL}.

\bibitem[{Qiu et~al.(2020)Qiu, Sun, Xu, Shao, Dai, and Huang}]{qiu2020pre}
Xipeng Qiu, Tianxiang Sun, Yige Xu, Yunfan Shao, Ning Dai, and Xuanjing Huang.
  2020.
\newblock Pre-trained models for natural language processing: A survey.
\newblock \emph{Science China Technological Sciences}, 63:1872–--1897.

\bibitem[{Radford et~al.(2021)Radford, Kim, Hallacy, Ramesh, Goh, Agarwal,
  Sastry, Askell, Mishkin, Clark et~al.}]{radford2021learning}
Alec Radford, Jong~Wook Kim, Chris Hallacy, Aditya Ramesh, Gabriel Goh,
  Sand~hini Agarwal, Girish Sastry, Amand~a Askell, Pamela Mishkin, Jack Clark,
  et~al. 2021.
\newblock Learning transferable visual models from natural language
  supervision.
\newblock \emph{OpenAI Blog}.

\bibitem[{Radford and Narasimhan(2018)}]{radfordimproving}
Alec Radford and Karthik Narasimhan. 2018.
\newblock Improving language understanding by generative pre-training.
\newblock \emph{OpenAI Blog}.

\bibitem[{Radford et~al.(2019)Radford, Wu, Child, Luan, Amodei, and
  Sutskever}]{radford2019language}
Alec Radford, Jeff Wu, Rewon Child, David Luan, Dario Amodei, and Ilya
  Sutskever. 2019.
\newblock Language models are unsupervised multitask learners.
\newblock \emph{OpenAI Blog}.

\bibitem[{Raffel et~al.(2020)Raffel, Shazeer, Roberts, Lee, Narang, Matena,
  Zhou, Li, and Liu}]{raffel2020exploring}
Colin Raffel, Noam Shazeer, Adam Roberts, Katherine Lee, Sharan Narang, Michael
  Matena, Yanqi Zhou, Wei Li, and Peter~J Liu. 2020.
\newblock Exploring the limits of transfer learning with a unified text-to-text
  transformer.
\newblock \emph{JMLR}, 21:1--67.

\bibitem[{Raina et~al.(2007)Raina, Battle, Lee, Packer, and Ng}]{raina2007self}
Rajat Raina, Alexis Battle, Honglak Lee, Benjamin Packer, and Andrew~Y Ng.
  2007.
\newblock Self-taught learning: transfer learning from unlabeled data.
\newblock In \emph{Proceedings of ICML}, pages 759--766.

\bibitem[{Rajbhandari et~al.(2020)Rajbhandari, Rasley, Ruwase, and
  He}]{rajbhandari2020zero}
Samyam Rajbhandari, Jeff Rasley, Olatunji Ruwase, and Yuxiong He. 2020.
\newblock Zero: Memory optimizations toward training trillion parameter models.
\newblock In \emph{Proceedings of SC}.

\bibitem[{Rajbhandari et~al.(2021)Rajbhandari, Ruwase, Rasley, Smith, and
  He}]{rajbhandari-zeroinfinity-2021}
Samyam Rajbhandari, Olatunji Ruwase, Jeff Rasley, Shaden Smith, and Yuxiong He.
  2021.
\newblock Zero-infinity: Breaking the gpu memory wall for extreme scale deep
  learning.
\newblock \emph{arXiv preprint arXiv:2104.07857}.

\bibitem[{Ramesh et~al.(2021)Ramesh, Pavlov, Goh, Gray, Voss, Radford, Chen,
  and Sutskever}]{ramesh2021zero}
Aditya Ramesh, Mikhail Pavlov, Gabriel Goh, Scott Gray, Chelsea Voss, Alec
  Radford, Mark Chen, and Ilya Sutskever. 2021.
\newblock Zero-shot text-to-image generation.
\newblock \emph{arXiv preprint arXiv:2102.12092}.

\bibitem[{Rasley et~al.(2020)Rasley, Rajbhandari, Ruwase, and
  He}]{rasley2020deepspeed}
Jeff Rasley, Samyam Rajbhandari, Olatunji Ruwase, and Yuxiong He. 2020.
\newblock Deepspeed: System optimizations enable training deep learning models
  with over 100 billion parameters.
\newblock In \emph{Proceedings of KDD}, pages 3505--3506.

\bibitem[{Ren et~al.(2021)Ren, Rajbhandari, Aminabadi, Ruwase, Yang, Zhang, Li,
  and He}]{ren2021zerooffload}
Jie Ren, Samyam Rajbhandari, Reza~Yazdani Aminabadi, Olatunji Ruwase, Shuangyan
  Yang, Minjia Zhang, Dong Li, and Yuxiong He. 2021.
\newblock Zero-offload: Democratizing billion-scale model training.
\newblock \emph{arxiv preprint arXiv:2101.06840}.

\bibitem[{Ren et~al.(2016)Ren, He, Girshick, and Sun}]{ren2015faster}
Shaoqing Ren, Kaiming He, Ross Girshick, and Jian Sun. 2016.
\newblock Faster r-cnn: towards real-time object detection with region proposal
  networks.
\newblock \emph{IEEE PAMI}, 39(6):1137--1149.

\bibitem[{Roberts et~al.(2020)Roberts, Raffel, and
  Shazeer}]{DBLP:conf/emnlp/RobertsRS20}
Adam Roberts, Colin Raffel, and Noam Shazeer. 2020.
\newblock How much knowledge can you pack into the parameters of a language
  model?
\newblock In \emph{Proceedings of EMNLP}, pages 5418--5426.

\bibitem[{Roller et~al.(2021)Roller, Dinan, Goyal, Ju, Williamson, Liu, Xu,
  Ott, Smith, Boureau, and Weston}]{roller-etal-2021-recipes}
Stephen Roller, Emily Dinan, Naman Goyal, Da~Ju, Mary Williamson, Yinhan Liu,
  Jing Xu, Myle Ott, Eric~Michael Smith, Y-Lan Boureau, and Jason Weston. 2021.
\newblock Recipes for building an open-domain chatbot.
\newblock In \emph{Proceedings of EACL}.

\bibitem[{Rosa and Mare{\v{c}}ek(2019)}]{DBLP:journals/corr/abs-1906-11511}
Rudolf Rosa and David Mare{\v{c}}ek. 2019.
\newblock Inducing syntactic trees from bert representations.
\newblock \emph{arXiv preprint arXiv:1906.11511}.

\bibitem[{Rosset et~al.(2020)Rosset, Xiong, Phan, Song, Bennett, and
  Tiwary}]{rosset2020knowledge}
Corby Rosset, Chenyan Xiong, Minh Phan, Xia Song, Paul Bennett, and Saurabh
  Tiwary. 2020.
\newblock Knowledge-aware language model pretraining.
\newblock \emph{arXiv preprint arXiv:2007.00655}.

\bibitem[{Roy et~al.(2021)Roy, Saffar, Vaswani, and
  Grangier}]{roy2021efficient}
Aurko Roy, Mohammad Saffar, Ashish Vaswani, and David Grangier. 2021.
\newblock Efficient content-based sparse attention with routing transformers.
\newblock \emph{TACL}, 9:53--68.

\bibitem[{Russakovsky et~al.(2015)Russakovsky, Deng, Su, Krause, Satheesh, Ma,
  Huang, Karpathy, Khosla, Bernstein et~al.}]{russakovsky2015imagenet}
Olga Russakovsky, Jia Deng, Hao Su, Jonathan Krause, Sanjeev Satheesh, Sean Ma,
  Zhiheng Huang, Andrej Karpathy, Aditya Khosla, Michael Bernstein, et~al.
  2015.
\newblock Imagenet large scale visual recognition challenge.
\newblock \emph{IJCV}, 115(3):211--252.

\bibitem[{Sanh et~al.(2019)Sanh, Debut, Chaumond, and
  Wolf}]{sanh2019distilbert}
Victor Sanh, Lysandre Debut, Julien Chaumond, and Thomas Wolf. 2019.
\newblock Distilbert, a distilled version of bert: smaller, faster, cheaper and
  lighter.
\newblock In \emph{Proceedings of NeurIPS}.

\bibitem[{Saunshi et~al.(2019)Saunshi, Plevrakis, Arora, Khodak, and
  eparkar}]{DBLP:conf/icml/SaunshiPAKK19}
Nikunj Saunshi, Orestis Plevrakis, Sanjeev Arora, Mikhail Khodak, and
  Hrishikesh~Khand eparkar. 2019.
\newblock A theoretical analysis of contrastive unsupervised representation
  learning.
\newblock In \emph{Proceedings of ICML}, pages 5628--5637.

\bibitem[{Saxe et~al.(2013)Saxe, McClelland, and Ganguli}]{saxe2013exact}
Andrew~M Saxe, James~L McClelland, and Surya Ganguli. 2013.
\newblock Exact solutions to the nonlinear dynamics of learning in deep linear
  neural networks.
\newblock \emph{arXiv preprint arXiv:1312.6120}.

\bibitem[{Schick and Sch{\"u}tze(2020)}]{schick2020s}
Timo Schick and Hinrich Sch{\"u}tze. 2020.
\newblock It's not just size that matters: Small language models are also
  few-shot learners.
\newblock \emph{arXiv preprint arXiv:2009.07118}.

\bibitem[{Schijndel et~al.(2019)Schijndel, Mueller, and
  Linzen}]{DBLP:conf/emnlp/SchijndelML19}
Marten~Van Schijndel, Aaron Mueller, and Tal Linzen. 2019.
\newblock Quantity doesn't buy quality syntax with neural language models.
\newblock In \emph{Proceedings of EMNLP-IJCNLP}, pages 5830--5836.

\bibitem[{Schlichtkrull et~al.(2018)Schlichtkrull, Kipf, Bloem, van~den Berg,
  Titov, and Welling}]{schlichtkrull2018modeling}
Michael~Sejr Schlichtkrull, Thomas~N Kipf, Peter Bloem, Rianne van~den Berg,
  Ivan Titov, and Max Welling. 2018.
\newblock Modeling relational data with graph convolutional networks.
\newblock In \emph{Proceedings of ESWC}, pages 593--607.

\bibitem[{Schmidt et~al.(2018)Schmidt, Santurkar, Tsipras, Talwar, and
  Madry}]{Schmidt2018samplecomplexity}
Ludwig Schmidt, Shibani Santurkar, Dimitris Tsipras, Kunal Talwar, and
  Aleksander Madry. 2018.
\newblock Adversarially robust generalization requires more data.
\newblock In \emph{Proceedings of NeurIPS}.

\bibitem[{Sermanet et~al.(2014)Sermanet, Eigen, Zhang, Mathieu, Fergus, and
  LeCun}]{sermanet2013overfeat}
Pierre Sermanet, David Eigen, Xiang Zhang, Micha{\"e}l Mathieu, Rob Fergus, and
  Yann LeCun. 2014.
\newblock Overfeat: Integrated recognition, localization and detection using
  convolutional networks.
\newblock In \emph{Proceedings of ICLR}.

\bibitem[{Sharma et~al.(2018)Sharma, Ding, Goodman, and
  Soricut}]{sharma2018conceptual}
Piyush Sharma, Nan Ding, Sebastian Goodman, and Radu Soricut. 2018.
\newblock Conceptual captions: A cleaned, hypernymed, image alt-text dataset
  for automatic image captioning.
\newblock In \emph{Proceedings of ACL)}, pages 2556--2565.

\bibitem[{Shazeer et~al.(2018)Shazeer, Cheng, Parmar, Tran, Vaswani,
  Koanantakool, Hawkins, Lee, Hong, Young et~al.}]{shazeer-2018-nips}
Noam Shazeer, Youlong Cheng, Niki Parmar, Dustin Tran, Ashish Vaswani, Penporn
  Koanantakool, Peter Hawkins, HyoukJoong Lee, Mingsheng Hong, Cliff Young,
  et~al. 2018.
\newblock Mesh-tensorflow: Deep learning for supercomputers.
\newblock In \emph{Proceedings of NeurIPS}.

\bibitem[{Shazeer et~al.(2017)Shazeer, Mirhoseini, Maziarz, Davis, Le, Hinton,
  and Dean}]{shazeer2017outrageously}
Noam Shazeer, Azalia Mirhoseini, Krzysztof Maziarz, Andy Davis, Quoc Le,
  Geoffrey Hinton, and Jeff Dean. 2017.
\newblock Outrageously large neural networks: The sparsely-gated
  mixture-of-experts layer.
\newblock In \emph{Proceedings of ICLR}.

\bibitem[{Shen et~al.(2020{\natexlab{a}})Shen, Dong, Ye, Ma, Yao, Gholami,
  Mahoney, and Keutzer}]{shen2020q}
Sheng Shen, Zhen Dong, Jiayu Ye, Linjian Ma, Zhewei Yao, Amir Gholami,
  Michael~W Mahoney, and Kurt Keutzer. 2020{\natexlab{a}}.
\newblock Q-bert: Hessian based ultra low precision quantization of bert.
\newblock In \emph{Proceedings of AAAI}, pages 8815--8821.

\bibitem[{Shen et~al.(2020{\natexlab{b}})Shen, Quach, Barzilay, and
  Jaakkola}]{shen2020blank}
Tianxiao Shen, Victor Quach, Regina Barzilay, and Tommi Jaakkola.
  2020{\natexlab{b}}.
\newblock Blank language models.
\newblock In \emph{Proceedings of EMNLP}, pages 5186--5198.

\bibitem[{Shi et~al.(2017)Shi, Chen, Zhu, Sun, Luo, Gu, and Zhou}]{zhusuan2017}
Jiaxin Shi, Jianfei. Chen, Jun Zhu, Shengyang Sun, Yucen Luo, Yihong Gu, and
  Yuhao Zhou. 2017.
\newblock Zhu{S}uan: A library for {B}ayesian deep learning.
\newblock \emph{arXiv preprint arXiv:1709.05870}.

\bibitem[{Shi et~al.(2016)Shi, Padhi, and Knight}]{DBLP:conf/emnlp/ShiPK16}
Xing Shi, Inkit Padhi, and Kevin Knight. 2016.
\newblock Does string-based neural {MT} learn source syntax?
\newblock In \emph{Proceedings of EMNLP}, pages 1526--1534.

\bibitem[{Shimodaira(2000)}]{shimodaira2000improving}
Hidetoshi Shimodaira. 2000.
\newblock Improving predictive inference under covariate shift by weighting the
  log-likelihood function.
\newblock \emph{Journal of statistical planning and inference}, 90(2):227--244.

\bibitem[{Shin et~al.(2020)Shin, Razeghi, Logan~IV, Wallace, and
  Singh}]{shin2020autoprompt}
Taylor Shin, Yasaman Razeghi, Robert~L Logan~IV, Eric Wallace, and Sameer
  Singh. 2020.
\newblock Autoprompt: Eliciting knowledge from language models with
  automatically generated prompts.
\newblock In \emph{Proceedings of EMNLP}, pages 4222--4235.

\bibitem[{Shoeybi et~al.(2019)Shoeybi, Patwary, Puri, LeGresley, Casper, and
  Catanzaro}]{shoeybi2019megatron}
Mohammad Shoeybi, Mostofa Patwary, Raul Puri, Patrick LeGresley, Jared Casper,
  and Bryan Catanzaro. 2019.
\newblock Megatron-lm: Training multi-billion parameter language models using
  model parallelism.
\newblock \emph{arXiv preprint arXiv:1909.08053}.

\bibitem[{Si et~al.(2020)Si, Zhang, Qi, Liu, Wang, Liu, and Sun}]{si2020better}
Chenglei Si, Zhengyan Zhang, Fanchao Qi, Zhiyuan Liu, Yasheng Wang, Qun Liu,
  and Maosong Sun. 2020.
\newblock Better robustness by more coverage: Adversarial training with mixup
  augmentation for robust fine-tuning.
\newblock \emph{arXiv preprint arXiv:2012.15699}.

\bibitem[{Simonyan and Zisserman(2015)}]{simonyan2014very}
Karen Simonyan and Andrew Zisserman. 2015.
\newblock Very deep convolutional networks for large-scale image recognition.
\newblock In \emph{Proceedings of ICLR}.

\bibitem[{Soares et~al.(2019)Soares, FitzGerald, Ling, and
  Kwiatkowski}]{soares2019matching}
Livio~Baldini Soares, Nicholas FitzGerald, Jeffrey Ling, and Tom Kwiatkowski.
  2019.
\newblock Matching the blanks: Distributional similarity for relation learning.
\newblock In \emph{Proceedings of ACL}.

\bibitem[{Song et~al.(2019)Song, Tan, Qin, Lu, and Liu}]{song2019mass}
Kaitao Song, Xu~Tan, Tao Qin, Jianfeng Lu, and Tie-Yan Liu. 2019.
\newblock Mass: Masked sequence to sequence pre-training for language
  generation.
\newblock In \emph{Proceedings of ICML}, pages 5926--5936.

\bibitem[{Song et~al.(2020)Song, Tan, Qin, Lu, and Liu}]{song2020mpnet}
Kaitao Song, Xu~Tan, Tao Qin, Jianfeng Lu, and Tie-Yan Liu. 2020.
\newblock Mpnet: Masked and permuted pre-training for language understanding.
\newblock In \emph{Proceedings of NeurIPS}, pages 16857--16867.

\bibitem[{Stock et~al.(2020)Stock, Joulin, Gribonval, Graham, and
  J{\'e}gou}]{stock2019and}
Pierre Stock, Armand Joulin, R{\'e}mi Gribonval, Benjamin Graham, and Herv{\'e}
  J{\'e}gou. 2020.
\newblock And the bit goes down: Revisiting the quantization of neural
  networks.
\newblock In \emph{Proceedings of ICLR}.

\bibitem[{Su et~al.(2020)Su, Zhu, Cao, Li, Lu, Wei, and Dai}]{su2019vl}
Weijie Su, Xizhou Zhu, Yue Cao, Bin Li, Lewei Lu, Furu Wei, and Jifeng Dai.
  2020.
\newblock Vl-bert: Pre-training of generic visual-linguistic representations.
\newblock In \emph{Proceedings of ICLR}.

\bibitem[{Sun et~al.(2019{\natexlab{a}})Sun, Myers, Vondrick, Murphy, and
  Schmid}]{sun2019videobert}
Chen Sun, Austin Myers, Carl Vondrick, Kevin Murphy, and Cordelia Schmid.
  2019{\natexlab{a}}.
\newblock Videobert: A joint model for video and language representation
  learning.
\newblock In \emph{Proceedings of ICCV}, pages 7464--7473.

\bibitem[{Sun et~al.(2021)Sun, Verga, Dhingra, Salakhutdinov, and
  Cohen}]{sun2021reasoning}
Haitian Sun, Pat Verga, Bhuwan Dhingra, Ruslan Salakhutdinov, and William~W
  Cohen. 2021.
\newblock Reasoning over virtual knowledge bases with open predicate relations.
\newblock \emph{arXiv preprint arXiv:2102.07043}.

\bibitem[{Sun et~al.(2019{\natexlab{b}})Sun, Cheng, Gan, and
  Liu}]{sun2019patient}
Siqi Sun, Yu~Cheng, Zhe Gan, and Jingjing Liu. 2019{\natexlab{b}}.
\newblock Patient knowledge distillation for bert model compression.
\newblock In \emph{Proceedings of EMNLP-IJCNLP}, pages 4323--4332.

\bibitem[{Sun et~al.(2020)Sun, Shao, Qiu, Guo, Hu, Huang, and
  Zhang}]{sun2020colake}
Tianxiang Sun, Yunfan Shao, Xipeng Qiu, Qipeng Guo, Yaru Hu, Xuanjing Huang,
  and Zheng Zhang. 2020.
\newblock Colake: Contextualized language and knowledge embedding.
\newblock In \emph{Proceedings of COLING}, pages 3660--3670.

\bibitem[{Sun et~al.(2019{\natexlab{c}})Sun, Wang, Li, Feng, Chen, Zhang, Tian,
  Zhu, Tian, and Wu}]{sun2019ernie}
Yu~Sun, Shuohuan Wang, Yukun Li, Shikun Feng, Xuyi Chen, Han Zhang, Xin Tian,
  Danxiang Zhu, Hao Tian, and Hua Wu. 2019{\natexlab{c}}.
\newblock Ernie: Enhanced representation through knowledge integration.
\newblock In \emph{Proceedings of ACL}, pages 1441--1451.

\bibitem[{Sun et~al.(2019{\natexlab{d}})Sun, Wang, Li, Feng, Tian, Wu, and
  Wang}]{sun2019ernie20}
Yu~Sun, Shuohuan Wang, Yukun Li, Shikun Feng, Hao Tian, Hua Wu, and Haifeng
  Wang. 2019{\natexlab{d}}.
\newblock Ernie 2.0: A continual pre-training framework for language
  understanding.
\newblock \emph{arXiv preprint arXiv:1907.12412}.

\bibitem[{Sutskever et~al.(2014)Sutskever, Vinyals, and
  Le}]{sutskever2014sequence}
Ilya Sutskever, Oriol Vinyals, and Quoc~V Le. 2014.
\newblock Sequence to sequence learning with neural networks.
\newblock In \emph{Proceedings of NeurIPS}, pages 3104--3112.

\bibitem[{Szegedy et~al.(2015)Szegedy, Liu, Jia, Sermanet, Reed, Anguelov,
  Erhan, Vanhoucke, and Rabinovich}]{szegedy2015going}
Christian Szegedy, Wei Liu, Yangqing Jia, Pierre Sermanet, Scott Reed, Dragomir
  Anguelov, Dumitru Erhan, Vincent Vanhoucke, and Andrew Rabinovich. 2015.
\newblock Going deeper with convolutions.
\newblock In \emph{Proceedings of CVPR}, pages 1--9.

\bibitem[{Tan and Bansal(2019)}]{tan2019lxmert}
Hao Tan and Mohit Bansal. 2019.
\newblock {LXMERT}: Learning cross-modality encoder representations from
  transformers.
\newblock In \emph{Proceedings of EMNLP-IJCNLP}, pages 5103--5114.

\bibitem[{Tay et~al.(2020)Tay, Dehghani, Bahri, and Metzler}]{tay2020efficient}
Yi~Tay, Mostafa Dehghani, Dara Bahri, and Donald Metzler. 2020.
\newblock Efficient transformers: A survey.
\newblock \emph{arXiv preprint arXiv:2009.06732}.

\bibitem[{Taylor(1953)}]{taylor1953cloze}
Wilson~L Taylor. 1953.
\newblock Cloze procedure: A new tool for measuring readability.
\newblock \emph{Journalism quarterly}, 30(4):415--433.

\bibitem[{Tenney et~al.(2019{\natexlab{a}})Tenney, Das, and
  Pavlick}]{DBLP:conf/acl/TenneyDP19}
Ian Tenney, Dipanjan Das, and Ellie Pavlick. 2019{\natexlab{a}}.
\newblock {BERT} rediscovers the classical {NLP} pipeline.
\newblock In \emph{Proceedings of ACL}, pages 4593--4601.

\bibitem[{Tenney et~al.(2019{\natexlab{b}})Tenney, Xia, Chen, Wang, Poliak,
  McCoy, Kim, Durme, Bowman, Das, and Pavlick}]{tenney2018what}
Ian Tenney, Patrick Xia, Berlin Chen, Alex Wang, Adam Poliak, R~Thomas McCoy,
  Najoung Kim, Benjamin~Van Durme, Sam Bowman, Dipanjan Das, and Ellie Pavlick.
  2019{\natexlab{b}}.
\newblock What do you learn from context? probing for sentence structure in
  contextualized word representations.
\newblock In \emph{Proceedings of ICLR}.

\bibitem[{Thrun and Pratt(1998)}]{thrun1998learning}
Sebastian Thrun and Lorien Pratt. 1998.
\newblock \emph{Learning to learn: Introduction and overview}.
\newblock Springer Science \& Business Media.

\bibitem[{Turian et~al.(2010)Turian, Ratinov, and Bengio}]{turian2010word}
Joseph Turian, Lev Ratinov, and Yoshua Bengio. 2010.
\newblock Word representations: a simple and general method for semi-supervised
  learning.
\newblock In \emph{Proceedings of ACL}, pages 384--394.

\bibitem[{Vaswani et~al.(2017)Vaswani, Shazeer, Parmar, Uszkoreit, Jones,
  Gomez, Kaiser, and Polosukhin}]{vaswani2017attention}
Ashish Vaswani, Noam Shazeer, Niki Parmar, Jakob Uszkoreit, Llion Jones,
  Aidan~N Gomez, {\L}ukasz Kaiser, and Illia Polosukhin. 2017.
\newblock Attention is all you need.
\newblock In \emph{Proceedings of NeurIPS}, pages 5998--6008.

\bibitem[{Veli{\v{c}}kovi{\'c} et~al.(2018)Veli{\v{c}}kovi{\'c}, Cucurull,
  Casanova, Romero, Lio, and Bengio}]{velivckovic2017graph}
Petar Veli{\v{c}}kovi{\'c}, Guillem Cucurull, Arantxa Casanova, Adriana Romero,
  Pietro Lio, and Yoshua Bengio. 2018.
\newblock Graph attention networks.
\newblock In \emph{Proceedings of ICLR}.

\bibitem[{Verga et~al.(2020)Verga, Sun, Soares, and Cohen}]{verga2020facts}
Pat Verga, Haitian Sun, Livio~Baldini Soares, and William~W Cohen. 2020.
\newblock Facts as experts: Adaptable and interpretable neural memory over
  symbolic knowledge.
\newblock \emph{arXiv preprint arXiv:2007.00849}.

\bibitem[{Vilares et~al.(2020)Vilares, Strzyz, S{\o}gaard, and
  G{\'{o}}mez{-}Rodr{\'{\i}}guez}]{DBLP:conf/aaai/VilaresSSG20}
David Vilares, Michalina Strzyz, Anders S{\o}gaard, and Carlos
  G{\'{o}}mez{-}Rodr{\'{\i}}guez. 2020.
\newblock Parsing as pretraining.
\newblock In \emph{Proceedings of AAAI}, pages 9114--9121.

\bibitem[{Vinyals et~al.(2015)Vinyals, Toshev, Bengio, and
  Erhan}]{vinyals2015show}
Oriol Vinyals, Alexander Toshev, Samy Bengio, and Dumitru Erhan. 2015.
\newblock Show and tell: A neural image caption generator.
\newblock In \emph{Proceedings of CVPR}, pages 3156--3164.

\bibitem[{Voita et~al.(2019)Voita, Talbot, Moiseev, Sennrich, and
  Titov}]{voita2019analyzing}
Elena Voita, David Talbot, Fedor Moiseev, Rico Sennrich, and Ivan Titov. 2019.
\newblock Analyzing multi-head self-attention: Specialized heads do the heavy
  lifting, the rest can be pruned.
\newblock In \emph{Proceedings of ACL}, pages 5797--5808.

\bibitem[{Wallace et~al.(2019{\natexlab{a}})Wallace, Feng, Kand~pal, Gardner,
  and Singh}]{wallace2019trigger}
Eric Wallace, Shi Feng, Nikhil Kand~pal, Matt Gardner, and Sameer Singh.
  2019{\natexlab{a}}.
\newblock Universal adversarial triggers for attacking and analyzing nlp.
\newblock In \emph{Proceedings of EMNLP-IJCNLP}, pages 2153--2162.

\bibitem[{Wallace et~al.(2019{\natexlab{b}})Wallace, Rodriguez, Feng, Yamada,
  and Boyd-Graber}]{wallace2019trick}
Eric Wallace, Pedro Rodriguez, Shi Feng, Ikuya Yamada, and Jordan Boyd-Graber.
  2019{\natexlab{b}}.
\newblock Trick me if you can: Human-in-the-loop generation of adversarial
  examples for question answering.
\newblock \emph{TACL}, 7:387--401.

\bibitem[{Wallace et~al.(2019{\natexlab{c}})Wallace, Wang, Li, Singh, and
  Gardner}]{DBLP:conf/emnlp/WallaceWLSG19}
Eric Wallace, Yizhong Wang, Sujian Li, Sameer Singh, and Matt Gardner.
  2019{\natexlab{c}}.
\newblock Do {NLP} models know numbers? probing numeracy in embeddings.
\newblock In \emph{Proceedings of EMNLP-IJCNLP}, pages 5306--5314.

\bibitem[{Wang et~al.(2020{\natexlab{a}})Wang, Liu, and
  Song}]{wang2020language}
Chenguang Wang, Xiao Liu, and Dawn Song. 2020{\natexlab{a}}.
\newblock Language models are open knowledge graphs.
\newblock \emph{arXiv preprint arXiv:2010.11967}.

\bibitem[{Wang et~al.(2021{\natexlab{a}})Wang, Ding, Li, and
  Zheng}]{wang2021cline}
Dong Wang, Ning Ding, Piji Li, and Hai-Tao Zheng. 2021{\natexlab{a}}.
\newblock Cline: Contrastive learning with semantic negative examples for
  natural language understanding.
\newblock In \emph{Proceedings of ACL}.

\bibitem[{Wang et~al.(2017)Wang, Jiang, Qian, Yang, Li, Zhang, Wang, and
  Tang}]{wang2017residual}
Fei Wang, Mengqing Jiang, Chen Qian, Shuo Yang, Cheng Li, Honggang Zhang,
  Xiaogang Wang, and Xiaoou Tang. 2017.
\newblock Residual attention network for image classification.
\newblock In \emph{Proceedings of the IEEE conference on computer vision and
  pattern recognition}, pages 3156--3164.

\bibitem[{Wang et~al.(2019)Wang, Huang, and Li}]{wang-2019-eurosys}
Minjie Wang, Chien-chin Huang, and Jinyang Li. 2019.
\newblock Supporting very large models using automatic dataflow graph
  partitioning.
\newblock In \emph{Proceedings of EuroSys}.

\bibitem[{Wang et~al.(2020{\natexlab{b}})Wang, Tang, Duan, Wei, Huang, Cao,
  Jiang, Zhou et~al.}]{wang2020k}
Ruize Wang, Duyu Tang, Nan Duan, Zhongyu Wei, Xuanjing Huang, Cuihong Cao,
  Daxin Jiang, Ming Zhou, et~al. 2020{\natexlab{b}}.
\newblock K-adapter: Infusing knowledge into pre-trained models with adapters.
\newblock \emph{arXiv preprint arXiv:2002.01808}.

\bibitem[{Wang et~al.(2020{\natexlab{c}})Wang, Li, Khabsa, Fang, and
  Ma}]{wang2020linformer}
Sinong Wang, Belinda Li, Madian Khabsa, Han Fang, and Hao Ma.
  2020{\natexlab{c}}.
\newblock Linformer: Self-attention with linear complexity.
\newblock \emph{arXiv preprint arXiv:2006.04768}.

\bibitem[{Wang et~al.(2020{\natexlab{d}})Wang, Wei, Dong, Bao, Yang, and
  Zhou}]{wang2020minilm}
Wenhui Wang, Furu Wei, Li~Dong, Hangbo Bao, Nan Yang, and Ming Zhou.
  2020{\natexlab{d}}.
\newblock Minilm: Deep self-attention distillation for task-agnostic
  compression of pre-trained transformers.
\newblock In \emph{Proceedings of NeurIPS}.

\bibitem[{Wang et~al.(2021{\natexlab{b}})Wang, Gao, Zhu, Zhang, Liu, Li, and
  Tang}]{wang2021kepler}
Xiaozhi Wang, Tianyu Gao, Zhaocheng Zhu, Zhengyan Zhang, Zhiyuan Liu, Juanzi
  Li, and Jian Tang. 2021{\natexlab{b}}.
\newblock Kepler: A unified model for knowledge embedding and pre-trained
  language representation.
\newblock \emph{TACL}, 9:176--194.

\bibitem[{Wang et~al.(2020{\natexlab{e}})Wang, Ke, Zheng, Huang, Jiang, Zhu,
  and Huang}]{wang2020chinese}
Yida Wang, Pei Ke, Yinhe Zheng, Kaili Huang, Yong Jiang, Xiaoyan Zhu, and
  Minlie Huang. 2020{\natexlab{e}}.
\newblock A large-scale chinese short-text conversation dataset.
\newblock In \emph{NLPCC}.

\bibitem[{Wang et~al.(2008)Wang, Song, and Zhang}]{wang2008transferred}
Zheng Wang, Yangqiu Song, and Changshui Zhang. 2008.
\newblock Transferred dimensionality reduction.
\newblock In \emph{Proceedings of ECML-PKDD}, pages 550--565.

\bibitem[{Wang et~al.(2020{\natexlab{f}})Wang, Dai, Wipf, and
  Zhu}]{wang2020ood}
Ziyu Wang, Bin Dai, David Wipf, and Jun Zhu. 2020{\natexlab{f}}.
\newblock Further analysis of outlier detection with deep generative models.
\newblock In \emph{Proceedings of NeurIPS}.

\bibitem[{Warstadt and Bowman(2020)}]{DBLP:conf/cogsci/WarstadtB20}
Alex Warstadt and Samuel~R. Bowman. 2020.
\newblock Can neural networks acquire a structural bias from raw linguistic
  data?
\newblock In \emph{Proceedings of CogSci}.

\bibitem[{Wei et~al.(2019)Wei, Ren, Li, Huang, Liao, Wang, Lin, Jiang, Chen,
  and Liu}]{wei2019nezha}
Junqiu Wei, Xiaozhe Ren, Xiaoguang Li, Wenyong Huang, Yi~Liao, Yasheng Wang,
  Jiashu Lin, Xin Jiang, Xiao Chen, and Qun Liu. 2019.
\newblock Nezha: Neural contextualized representation for chinese language
  understanding.
\newblock \emph{arXiv preprint arXiv:1909.00204}.

\bibitem[{{Wei} et~al.(2021){Wei}, {Hu}, {Weng}, {Xing}, {Yu}, and
  {Luo}}]{wei2021on}
Xiangpeng {Wei}, Yue {Hu}, Rongxiang {Weng}, Luxi {Xing}, Heng {Yu}, and Weihua
  {Luo}. 2021.
\newblock On learning universal representations across languages.
\newblock In \emph{Proceedings of ICLR}.

\bibitem[{Wharton et~al.(1994)Wharton, Holyoak, Downing, Lange, Wickens, and
  Melz}]{wharton1994below}
Charles~M Wharton, Keith~J Holyoak, Paul~E Downing, Trent~E Lange, Thomas~D
  Wickens, and Eric~R Melz. 1994.
\newblock Below the surface: Analogical similarity and retrieval competition in
  reminding.
\newblock \emph{Cognitive Psychology}, 26:64--101.

\bibitem[{Williams et~al.(2007)Williams, Bonilla, and Chai}]{williams2007multi}
Chris Williams, Edwin~V Bonilla, and Kian~M Chai. 2007.
\newblock Multi-task gaussian process prediction.
\newblock In \emph{Proceedings of NeurIPS}, pages 153--160.

\bibitem[{Wu et~al.(2016)Wu, Schuster, Chen, Le, Norouzi, Macherey, Krikun,
  Cao, Gao, Macherey et~al.}]{wu2016google}
Yonghui Wu, Mike Schuster, Zhifeng Chen, Quoc~V Le, Mohammad Norouzi, Wolfgang
  Macherey, Maxim Krikun, Yuan Cao, Qin Gao, Klaus Macherey, et~al. 2016.
\newblock Google's neural machine translation system: Bridging the gap between
  human and machine translation.
\newblock \emph{arXiv preprint arXiv:1609.08144}.

\bibitem[{Wu et~al.(2018)Wu, Xiong, Yu, and Lin}]{wu2018unsupervised}
Zhirong Wu, Yuanjun Xiong, Stella~X Yu, and Dahua Lin. 2018.
\newblock Unsupervised feature learning via non-parametric instance
  discrimination.
\newblock In \emph{Proceedings of CVPR}, pages 3733--3742.

\bibitem[{Wu et~al.(2020)Wu, Chen, Kao, and Liu}]{DBLP:conf/acl/WuCKL20}
Zhiyong Wu, Yun Chen, Ben Kao, and Qun Liu. 2020.
\newblock Perturbed masking: Parameter-free probing for analyzing and
  interpreting {BERT}.
\newblock In \emph{Proceedings of ACL}, pages 4166--4176.

\bibitem[{Xia et~al.(2020)Xia, Huang, Duan, Zhang, Ji, Sui, Cui, Bharti, Liu,
  and Zhou}]{xia2020xgpt}
Qiaolin Xia, Haoyang Huang, Nan Duan, Dongdong Zhang, Lei Ji, Zhifang Sui,
  Edward Cui, Taroon Bharti, Xin Liu, and Ming Zhou. 2020.
\newblock Xgpt: Cross-modal generative pre-training for image captioning.
\newblock \emph{arXiv preprint arXiv:2003.01473}.

\bibitem[{Xiong et~al.(2016)Xiong, Merity, and Socher}]{xiong2016dynamic}
Caiming Xiong, Stephen Merity, and Richard Socher. 2016.
\newblock Dynamic memory networks for visual and textual question answering.
\newblock In \emph{Proceedings of ICML}, pages 2397--2406.

\bibitem[{Xiong et~al.(2019)Xiong, Du, Wang, and
  Stoyanov}]{xiong2019pretrained}
Wenhan Xiong, Jingfei Du, William~Yang Wang, and Veselin Stoyanov. 2019.
\newblock Pretrained encyclopedia: Weakly supervised knowledge-pretrained
  language model.
\newblock In \emph{Proceedings of ICLR}.

\bibitem[{Xu et~al.(2021)Xu, Li, Zhang, Du, Kawarabayashi, and
  Jegelka}]{xu2020neural}
Keyulu Xu, Jingling Li, Mozhi Zhang, Simon~S Du, Ken-ichi Kawarabayashi, and
  Stefanie Jegelka. 2021.
\newblock How neural networks extrapolate: From feedforward to graph neural
  networks.
\newblock In \emph{Proceedings of ICLR}.

\bibitem[{Yang et~al.(2020)Yang, Ma, Zhang, Wu, jun Li, and
  Zhou}]{Yang2020AlternatingLM}
Jian Yang, Shuming Ma, D.~Zhang, Shuangzhi Wu, Zhou jun Li, and M.~Zhou. 2020.
\newblock Alternating language modeling for cross-lingual pre-training.
\newblock In \emph{Proceedings of AAAI}, pages 9386--9393.

\bibitem[{Yang et~al.(2019)Yang, Dai, Yang, Carbonell, Salakhutdinov, and
  Le}]{yang2019xlnet}
Zhilin Yang, Zihang Dai, Yiming Yang, Jaime Carbonell, Ruslan Salakhutdinov,
  and Quoc~V Le. 2019.
\newblock Xlnet: Generalized autoregressive pretraining for language
  understanding.
\newblock In \emph{Proceedings of NeurIPS}.

\bibitem[{Yao et~al.(2021)Yao, Zhong, Zhang, Han, Wang, Xiao, Zeng, Liu, and
  Sun}]{yao2019adversarial}
Yuan Yao, Haoxi Zhong, Zhengyan Zhang, Xu~Han, Xiaozhi Wang, Chaojun Xiao,
  Guoyang Zeng, Zhiyuan Liu, and Maosong Sun. 2021.
\newblock Adversarial language games for advanced natural language
  intelligence.
\newblock In \emph{Proceedings of AAAI}.

\bibitem[{You et~al.(2017)You, Gitman, and Ginsburg}]{you2017scaling}
Yang You, Igor Gitman, and Boris Ginsburg. 2017.
\newblock Scaling sgd batch size to 32k for imagenet training.
\newblock \emph{arXiv preprint arXiv:1708.03888}.

\bibitem[{You et~al.(2020)You, Li, Reddi, Hseu, Kumar, Bhojanapalli, Song,
  Demmel, Keutzer, and Hsieh}]{You2020Large}
Yang You, Jing Li, Sashank Reddi, Jonathan Hseu, Sanjiv Kumar, Srinadh
  Bhojanapalli, Xiaodan Song, James Demmel, Kurt Keutzer, and Cho-Jui Hsieh.
  2020.
\newblock Large batch optimization for deep learning: Training bert in 76
  minutes.
\newblock In \emph{Proceedings of ICLR}.

\bibitem[{Zadrozny(2004)}]{zadrozny2004learning}
Bianca Zadrozny. 2004.
\newblock Learning and evaluating classifiers under sample selection bias.
\newblock In \emph{Proceedings of ICML}.

\bibitem[{Zafrir et~al.(2019)Zafrir, Boudoukh, Izsak, and
  Wasserblat}]{zafrir2019q8bert}
Ofir Zafrir, Guy Boudoukh, Peter Izsak, and Moshe Wasserblat. 2019.
\newblock Q8bert: Quantized 8bit bert.
\newblock In \emph{Proceedings of NeurIPS}.

\bibitem[{Zaheer et~al.(2020)Zaheer, Guruganesh, Dubey, Ainslie, Alberti,
  Ontanon, Pham, Ravula, Wang, Yang et~al.}]{zaheer2020big}
Manzil Zaheer, Guru Guruganesh, Avinava Dubey, Joshua Ainslie, Chris Alberti,
  Santiago Ontanon, Philip Pham, Anirudh Ravula, Qifan Wang, Li~Yang, et~al.
  2020.
\newblock Big bird: Transformers for longer sequences.
\newblock In \emph{Proceedings of NeurIPS}, pages 17283--17297.

\bibitem[{Zang et~al.(2020)Zang, Qi, Yang, Liu, Zhang, Liu, and
  Sun}]{zang2020word}
Yuan Zang, Fanchao Qi, Chenghao Yang, Zhiyuan Liu, Meng Zhang, Qun Liu, and
  Maosong Sun. 2020.
\newblock Word-level textual adversarial attacking as combinatorial
  optimization.
\newblock In \emph{Proceedings of ACL}, pages 6066--6080.

\bibitem[{Zeng et~al.(2021)Zeng, Ren, Su, Wang, Liao, Wang, Jiang, Yang, Wang,
  Zhang et~al.}]{zeng2021pangu}
Wei Zeng, Xiaozhe Ren, Teng Su, Hui Wang, Yi~Liao, Zhiwei Wang, Xin Jiang,
  ZhenZhang Yang, Kaisheng Wang, Xiaoda Zhang, et~al. 2021.
\newblock Pangu-alpha: Large-scale autoregressive pretrained chinese language
  models with auto-parallel computation.
\newblock \emph{arXiv preprint arXiv:2104.12369}.

\bibitem[{Zhang et~al.(2017)Zhang, Bengio, Hardt, Recht, and
  Vinyals}]{zhang2017understanding}
Chiyuan Zhang, Samy Bengio, Moritz Hardt, Benjamin Recht, and Oriol Vinyals.
  2017.
\newblock Understanding deep learning requires rethinking generalization.
\newblock In \emph{Proceedings of ICLR}.

\bibitem[{Zhang et~al.(2019{\natexlab{a}})Zhang, Liu, Tang, Dong, Yao, Zhang,
  Gu, Wang, Shao, Li et~al.}]{zhang2019oag}
Fanjin Zhang, Xiao Liu, Jie Tang, Yuxiao Dong, Peiran Yao, Jie Zhang, Xiaotao
  Gu, Yan Wang, Bin Shao, Rui Li, et~al. 2019{\natexlab{a}}.
\newblock Oag: Toward linking large-scale heterogeneous entity graphs.
\newblock In \emph{Proceedings of KDD}, pages 2585--2595.

\bibitem[{Zhang et~al.(2020{\natexlab{a}})Zhang, Zhao, Saleh, and
  Liu}]{zhang2020pegasus}
Jingqing Zhang, Yao Zhao, Mohammad Saleh, and Peter Liu. 2020{\natexlab{a}}.
\newblock Pegasus: Pre-training with extracted gap-sentences for abstractive
  summarization.
\newblock In \emph{Proceedings of ICML}, pages 11328--11339.

\bibitem[{Zhang and He(2020)}]{NEURIPS2020_a1140a3d}
Minjia Zhang and Yuxiong He. 2020.
\newblock Accelerating training of transformer-based language models with
  progressive layer dropping.
\newblock In \emph{Proceedings of NeurIPS}, pages 14011--14023.

\bibitem[{Zhang et~al.(2020{\natexlab{b}})Zhang, Hou, Yin, Shang, Chen, Jiang,
  and Liu}]{zhang2020ternarybert}
Wei Zhang, Lu~Hou, Yichun Yin, Lifeng Shang, Xiao Chen, Xin Jiang, and Qun Liu.
  2020{\natexlab{b}}.
\newblock Ternarybert: Distillation-aware ultra-low bit bert.
\newblock In \emph{Proceedings of EMNLP}, pages 509--521.

\bibitem[{Zhang et~al.(2021{\natexlab{a}})Zhang, Gu, Han, Chen, Xiao, Sun, Yao,
  Qi, Guan, Ke, Cai, Zeng, Tan, Liu, Huang, Han, Liu, Zhu, and Sun}]{cpm-v2}
Zhengyan Zhang, Yuxian Gu, Xu~Han, Shengqi Chen, Chaojun Xiao, Zhenbo Sun, Yuan
  Yao, Fanchao Qi, Jian Guan, Pei Ke, Yanzheng Cai, Guoyang Zeng, Zhixing Tan,
  Zhiyuan Liu, Minlie Huang, Wentao Han, Yang Liu, Xiaoyan Zhu, and Maosong
  Sun. 2021{\natexlab{a}}.
\newblock Cpm-2: Large-scale cost-efficient pre-trained language models.

\bibitem[{Zhang et~al.(2019{\natexlab{b}})Zhang, Han, Liu, Jiang, Sun, and
  Liu}]{zhang2019ernie}
Zhengyan Zhang, Xu~Han, Zhiyuan Liu, Xin Jiang, Maosong Sun, and Qun Liu.
  2019{\natexlab{b}}.
\newblock Ernie: Enhanced language representation with informative entities.
\newblock In \emph{Proceedings of ACL}, pages 1441--1451.

\bibitem[{Zhang et~al.(2020{\natexlab{c}})Zhang, Han, Zhou, Ke, Gu, Ye, Qin,
  Su, Ji, Guan et~al.}]{zhang2020cpm}
Zhengyan Zhang, Xu~Han, Hao Zhou, Pei Ke, Yuxian Gu, Deming Ye, Yujia Qin,
  Yusheng Su, Haozhe Ji, Jian Guan, et~al. 2020{\natexlab{c}}.
\newblock Cpm: A large-scale generative chinese pre-trained language model.
\newblock \emph{arXiv preprint arXiv:2012.00413}.

\bibitem[{Zhang et~al.(2021{\natexlab{b}})Zhang, Qi, Liu, Liu, and
  Sun}]{zhang_know_2021}
Zhengyan Zhang, Fanchao Qi, Zhiyuan Liu, Qun Liu, and Maosong Sun.
  2021{\natexlab{b}}.
\newblock Know what you don't need: {Single}-{Shot} {Meta}-{Pruning} for
  attention heads.
\newblock \emph{AI Open}, 2:36--42.

\bibitem[{Zhang et~al.(2021{\natexlab{c}})Zhang, Xiao, Li, Lv, Qi, Liu, Wang,
  Jiang, and Sun}]{zhang2021red}
Zhengyan Zhang, Guangxuan Xiao, Yongwei Li, Tian Lv, Fanchao Qi, Zhiyuan Liu,
  Yasheng Wang, Xin Jiang, and Maosong Sun. 2021{\natexlab{c}}.
\newblock Red alarm for pre-trained models: Universal vulnerabilities by
  neuron-level backdoor attacks.
\newblock \emph{arXiv preprint arXiv:2101.06969}.

\bibitem[{Zheng et~al.(2015)Zheng, Jayasumana, Romera-Paredes, Vineet, Su, Du,
  Huang, and Torr}]{zheng2015conditional}
Shuai Zheng, Sadeep Jayasumana, Bernardino Romera-Paredes, Vibhav Vineet,
  Zhizhong Su, Dalong Du, Chang Huang, and Philip~HS Torr. 2015.
\newblock Conditional random fields as recurrent neural networks.
\newblock In \emph{Proceedings of ICCV}, pages 1529--1537.

\bibitem[{Zhou et~al.(2020{\natexlab{a}})Zhou, Palangi, Zhang, Hu, Corso, and
  Gao}]{zhou2020unified}
Luowei Zhou, Hamid Palangi, Lei Zhang, Houdong Hu, Jason Corso, and Jianfeng
  Gao. 2020{\natexlab{a}}.
\newblock Unified vision-language pre-training for image captioning and vqa.
\newblock In \emph{Proceedings of AAAI}, pages 13041--13049.

\bibitem[{Zhou et~al.(2020{\natexlab{b}})Zhou, Zhang, Cui, and
  an~Huang}]{DBLP:conf/aaai/ZhouZCH20}
Xuhui Zhou, Yue Zhang, Leyang Cui, and Dand an~Huang. 2020{\natexlab{b}}.
\newblock Evaluating commonsense in pre-trained language models.
\newblock In \emph{Proceedings of AAAI}, pages 9733--9740.

\bibitem[{Zhu et~al.(2015)Zhu, Kiros, Zemel, Salakhutdinov, Urtasun, Torralba,
  and Fidler}]{zhu2015book}
Yukun Zhu, Ryan Kiros, Rich Zemel, Ruslan Salakhutdinov, Raquel Urtasun,
  Antonio Torralba, and Sanja Fidler. 2015.
\newblock Aligning books and movies: Towards story-like visual explanations by
  watching movies and reading books.
\newblock In \emph{Proceedings of ICCV}, pages 19--27.

\bibitem[{Zoph et~al.(2020)Zoph, Ghiasi, Lin, Cui, Liu, Cubuk, and
  Le}]{zoph2020rethinking}
Barret Zoph, Golnaz Ghiasi, Tsung-Yi Lin, Yin Cui, Hanxiao Liu, Ekin~Dogus
  Cubuk, and Quoc Le. 2020.
\newblock Rethinking pre-training and self-training.
\newblock \emph{Proceedings of NeurIPS}, 33.

\bibitem[{Zou et~al.(2021)Zou, Yin, Zhong, Yang, Yang, and
  Tang}]{zou2021controllable}
Xu~Zou, Da~Yin, Qingyang Zhong, Hongxia Yang, Zhilin Yang, and Jie Tang. 2021.
\newblock Controllable generation from pre-trained language models via inverse
  prompting.
\newblock \emph{arXiv preprint arXiv:2103.10685}.

\end{thebibliography}

\end{document}